\documentclass{article}

\usepackage{microtype}
\usepackage{graphicx}
\usepackage{booktabs} %
\usepackage[export]{adjustbox}
\usepackage{subcaption}
\usepackage{tabularx}
\usepackage{multirow}
\usepackage{cellspace}
\usepackage{makecell}
\usepackage{boldline}
\usepackage{setspace}
\usepackage{enumitem}
\usepackage{physics}
\usepackage{numprint}
\usepackage{placeins}
\npthousandsep{,\,}
\setlist[itemize]{itemsep=0.5pt, topsep=0pt}
\setlist[enumerate]{itemsep=0.5pt, topsep=0pt}
\usepackage[font=small,skip=0pt]{caption}

\usepackage{everypage}
\usepackage{environ}
\newcounter{abspage}%

\DeclareMathOperator{\atan2}{atan2}
\DeclareMathOperator{\Adam-atan2}{Adam-atan2}
\DeclareMathOperator{\Adamoptim}{Adam}

\makeatletter
\newcommand{\newSFPage}[1]%
  {\global\expandafter\let\csname SFPage@#1\endcsname\null}

\NewEnviron{SidewaysFigure}{\begin{figure}[p]
\protected@write\@auxout{\let\theabspage=\relax}%
  {\string\newSFPage{\theabspage}}%
\ifdim\textwidth=\textheight
  \rotatebox{270}{\parbox[c][\textwidth][c]{\linewidth}{\BODY}}%
\else
  \rotatebox{270}{\parbox[c][\textwidth][c]{\textheight}{\BODY}}%
\fi
\end{figure}}

\AddEverypageHook{%
  \ifdim\textwidth=\textheight
    \stepcounter{abspage}%
  \else
    \@ifundefined{SFPage@\theabspage}{}{\global\pdfpageattr{/Rotate 0}}%
    \stepcounter{abspage}%
    \@ifundefined{SFPage@\theabspage}{}{\global\pdfpageattr{/Rotate 270}}%
  \fi}
\makeatother

\newcommand{\sref}[1]{\S\ref{#1}}
\newcounter{repeattablecounter}
\newcounter{stashtablecounter}

\usepackage{hyperref}

\DeclareCaptionLabelFormat{repeattable}{#1 #2 (repeated)}

\newenvironment{tightcenter}{%
  \setlength\topsep{2pt}
  \setlength\parskip{0pt}
  \begin{center}
}{%
  \end{center}
}

\usepackage[accepted]{icml2024}

\usepackage{amsmath}
\usepackage{amssymb}
\usepackage{mathtools}
\usepackage{amsthm}

\usepackage[capitalize,noabbrev]{cleveref}

\usepackage{fancyvrb}

\theoremstyle{plain}
\newtheorem{theorem}{Theorem}[section]

\theoremstyle{definition}
\newtheorem{definition}[theorem]{Definition}

\newtheorem{appendixdef}{Definition}[section]

\theoremstyle{remark}

\makeatletter
\def\convertto#1#2{\strip@pt\dimexpr #2*65536/\number\dimexpr 1#1}
\makeatother

\usepackage[textsize=tiny]{todonotes}

\newcommand{\R}{\mathbb{R}}
\newcommand{\loss}{\mathcal{L}}
\newcommand{\hquad}{\hspace{0.5em}}
\newcommand{\partialloss}[1]{\frac{\partial \mathcal{L}}{\partial #1}}
\newcommand{\partialfrac}[2]{\frac{\partial #1}{\partial #2}}

\icmltitlerunning{Scaling Exponents Across Parameterizations and Optimizers}

\begin{document}

\definecolor{tabcyan}{HTML}{17becf}
\definecolor{tabblue}{HTML}{1f77b4}

\twocolumn[
\icmltitle{Scaling Exponents Across Parameterizations and Optimizers}

\icmlsetsymbol{equal}{*}

\begin{icmlauthorlist}
\icmlauthor{Katie Everett}{GDM,MIT}
\icmlauthor{Lechao Xiao}{GDM}
\icmlauthor{Mitchell Wortsman}{doneatgdm}
\icmlauthor{Alexander A. Alemi}{GDM}
\icmlauthor{Roman Novak}{doneatgdm}
\icmlauthor{Peter J. Liu}{GDM}
\icmlauthor{Izzeddin Gur}{GDM}
\icmlauthor{Jascha Sohl-Dickstein}{doneatgdm}
\icmlauthor{Leslie Pack Kaelbling}{MIT}
\icmlauthor{Jaehoon Lee}{GDM}
\icmlauthor{Jeffrey Pennington}{GDM}
\end{icmlauthorlist}

\icmlaffiliation{GDM}{Google DeepMind}
\icmlaffiliation{MIT}{MIT}
\icmlaffiliation{doneatgdm}{Work done at Google DeepMind}
\icmlcorrespondingauthor{Katie Everett}{everettk@google.com}

\icmlkeywords{Machine Learning, ICML, parameterization, neural tangent kernel, NTK, muP, maximal update parameterization, mean field parameterization, scaling law, hyperparameter transfer, epsilon, alignment}

\vskip 0.3in
]

\printAffiliationsAndNotice{}  %

\begin{abstract}
Robust and effective scaling of models from small to large width typically requires the precise adjustment of many algorithmic and architectural details, such as parameterization and optimizer choices. In this work, we propose a new perspective on parameterization by investigating a key assumption in prior work about the alignment between parameters and data and derive new theoretical results under weaker assumptions and a broader set of optimizers. Our extensive empirical investigation includes \emph{tens of thousands} of models trained with \emph{all combinations of} three optimizers, four parameterizations, several alignment assumptions, more than a dozen learning rates, and fourteen model sizes up to 26.8B parameters. We find that the best learning rate scaling prescription would often have been excluded by the assumptions in prior work. Our results show that all parameterizations, not just maximal update parameterization (muP), can achieve hyperparameter transfer; moreover, our novel per-layer learning rate prescription for standard parameterization outperforms muP. Finally, we demonstrate that an overlooked aspect of parameterization, the epsilon parameter in Adam, must be scaled correctly to avoid gradient underflow and propose \emph{Adam-atan2}, a new numerically stable, scale-invariant version of Adam that eliminates the epsilon hyperparameter entirely.
\end{abstract}

\vspace{-16pt}
\section{Introduction} 
\label{introduction}

A neural network parameterization is a prescription for scaling a set of important quantities with respect to a set of scaling dimensions. Most often, the parameterized quantities include the initialization scale, parameter multipliers and learning rate, and scaling dimensions may include model width, model depth, context length, batch size and training horizon. Parameterizations with well-understood scaling behavior can prescribe the exponents for these quantities to ensure that the training dynamics behave in a stable and predictable manner as the model increases in scale.

When exponents are not carefully selected, models can have scaling mismatches that are not obvious from experiments at small scale. One such mismatch can occur in the learning rate exponents between different layers in a neural network: while most models currently train with a global learning rate, where all parameters in the model are updated with the same learning rate, this constrains all layers to use the same exponent when differing exponents may be required. For example, we will see this phenomenon in standard parameterization models trained with stochastic gradient descent (SGD). Under common assumptions, the hidden layer learning rate would ideally scale like $O(1/\sqrt{n})$ whereas the readout layer learning rate would scale like $O(1/n)$ where $n$ is the model width. With a single global learning rate, we are forced to make one of two bad choices: either we choose a learning rate that scales like $O(1/\sqrt{n})$, which makes nontrivial updates to the hidden layer but causes the readout layer to explode with scale, or we use a learning rate that scales like $O(1/n)$, which preserves the readout layer stability but causes ``vanishing" updates to the earlier activations in the sense that the updates approach zero as the width goes to infinity. This mismatch may persist silently until the model is scaled past a threshold where the difference in exponents dominates over other factors: a risky situation in the current ``train once" era of very large models. This strongly motivates a principled understanding of parameterization and well-defined scaling limits rather than relying solely on extrapolation of empirical results.

In addition, parameterizations that are implemented with a particular functional form can enable hyperparameter transfer across scales, where relatively cheap hyperparameter search using small models can be used to select hyperparameters for larger, more expensive models~\citep{yang2022tensorv}. This functional form specifies each parameterized quantity using a constant multiplicative factor that is typically determined empirically along with a scaling exponent that is motivated theoretically. This recipe for parameterized quantities, where
\begin{align*}
\begin{footnotesize}
    \parbox{55pt}{\centering{parameterized\newline quantity}}\hquad=\hquad \parbox{41pt}{\centering{empirical\newline constant}} \cdot \biggl(\,\parbox{36pt}{\centering{scaling\newline dimension}}\,\biggr)^{\parbox{30pt}{\scriptsize \centering{theoretical\newline exponent}}}
\end{footnotesize}
\end{align*}
allows for the reuse of the constant factors: when the scaling behavior is fully encapsulated by the theoretical exponent, the optimal empirical constant is the same across scales. We can therefore determine the optimal constant factors via hyperparameter search on small models and reuse them on large models where hyperparameter search would be expensive or impossible.

Among the scaling dimensions, the existing literature for \emph{width scaling} has the most extensive theoretical results but important open questions remain. \citet{yang2021tensoriv} and \citet{yang2023tensorivb} define a space of width-scaling parameterizations for SGD and Adam respectively. Based on the goals of ensuring the activations remain at constant scale and the logits do not exceed constant scale with respect to width, they derive constraints for stability and non-triviality of a parameterization that predict how the learning rate should scale with width. The derivations of these constraints make a key assumption: the updates to the parameters are assumed to be sufficiently correlated with the data distribution to impact the scaling exponent of the activations. We refer to this correlation as ``alignment" because, when correlated, the parameter and data vectors point in similar directions. However, the literature is lacking in extensive empirical measurements of when, where, and how much alignment accumulates between the parameters and activations during training and its impact on the scaling exponents of the learning rate.

\setcounter{repeattablecounter}{\value{table}}
\begin{table*}
  \vspace{2pt}
  \centering
  \begin{footnotesize}
    \setlength\extrarowheight{2.5pt}
    \adjustbox{scale=0.8}{
    \begin{tabular}
        {r | l | p{1.45cm}| p{1.3cm} | p{1.1cm} || p{1.3cm} 
        | p{1.3cm} 
        | p{1.8cm}  || p{1.3cm}
        | p{1.3cm}
        | p{1.8cm} |}
        \hline
        \multicolumn{2}{ V{1} r V{1}}{}
        & Initialization Variance & Parameter Multiplier & Gradient & SGD LR, Full Align & Adam LR, Full Align & Adafactor LR, Full Align & SGD LR, No Align & Adam LR, No Align & Adafactor LR, No Align
        \\ 
        \hline
        \multicolumn{1}{ V{1} r V{1}}{\multirow{3}{*}{Standard}} & Embedding & 
        $1\hphantom{/n}$    & $1\hphantom{/\sqrt{n}}$                  & $1/\sqrt{n}$                  & $\sqrt{n}\hphantom{/\sqrt{n}}$         & $1\hphantom{/\sqrt{n}}$ & 1  & $\sqrt{n}\hphantom{/\sqrt{n}}$         & $1\hphantom{/\sqrt{n}}$ & 1              
        \\ 
        \multicolumn{1}{ V{1} r V{1}}{}                                   & Hidden    & 
        $1/n$               & $1\hphantom{/\sqrt{n}}$                  & $1/\sqrt{n}$                  & $1/\sqrt{n}$                           & $1/n\hphantom{n}$               & $1/\sqrt{n}$ & $1\hphantom{/\sqrt{n}}$  & $1/\sqrt{n}$ & 1 
        \\ 
        \multicolumn{1}{ V{1} r V{1}}{}                                   & Readout   & 
        $1/n$               & $1\hphantom{/\sqrt{n}}$                  & $1\hphantom{/\sqrt{n}}$       & $1/n\hphantom{n}$                      & $1/n\hphantom{n}$                & $1/\sqrt{n}$ & $1/\sqrt{n}$ & $1/\sqrt{n}$ & 1 
        \\ 
        \hline
        \multicolumn{1}{ V{1} r V{1}}{\multirow{3}{*}{NTK}}      & Embedding & 
        $1\hphantom{/n}$    & $1\hphantom{/\sqrt{n}}$                  & $1/\sqrt{n}$                  & $\sqrt{n}\hphantom{/\sqrt{n}}$         & $1\hphantom{/\sqrt{n}}$  & 1 & $\sqrt{n}\hphantom{/\sqrt{n}}$         & $1\hphantom{/\sqrt{n}}$  & 1               
        \\ 
        \multicolumn{1}{ V{1} r V{1}}{}                                   & Hidden    & 
        $1\hphantom{/n}$    & $1/\sqrt{n}$                             & $1/n\hphantom{n}$             & $\sqrt{n}\hphantom{/\sqrt{n}}$         & $1/\sqrt{n}$      & $1/\sqrt{n}$ & $n\hphantom{/\sqrt{n}}$ & $1\hphantom{/\sqrt{n}}$ & 1   
        \\ 
        \multicolumn{1}{ V{1} r V{1}}{}                                   & Readout   & 
        $1\hphantom{/n}$    & $1/\sqrt{n}$                             & $1/\sqrt{n}$                  & $1\hphantom{/\sqrt{n}}$                & $1/\sqrt{n}$        & $1/\sqrt{n}$ & $\sqrt{n}\hphantom{/\sqrt{n}}$ & $1\hphantom{/\sqrt{n}}$ & 1 
        \\ 
        \hline
        \multicolumn{1}{ V{1} r V{1}}{\multirow{3}{*}{muP}}      & Embedding & 
        $1/n$               & $\sqrt{n}\hphantom{/\sqrt{n}}$           & $1/\sqrt{n}$                  & $1\hphantom{/\sqrt{n}}$                & $1/\sqrt{n}$     & 1    & $1\hphantom{/\sqrt{n}}$                & $1/\sqrt{n}$     & 1 
        \\ 
        \multicolumn{1}{ V{1} r V{1}}{}                                   & Hidden    & 
        $1/n$               & $1\hphantom{/\sqrt{n}}$                  & $1/n\hphantom{n}$             & $1\hphantom{/\sqrt{n}}$                & $1/n\hphantom{n}$       & $1/\sqrt{n}$    & $\sqrt{n}\hphantom{/\sqrt{n}}$ &  $1/\sqrt{n}$ & 1      
        \\ 
        \multicolumn{1}{ V{1} r V{1}}{}                                   & Readout   & 
        $1/n$               & $1/\sqrt{n}$                             & $1/\sqrt{n}$                  & $1\hphantom{/\sqrt{n}}$                & $1/\sqrt{n}$         & $1$ & 1 & $1\hphantom{/\sqrt{n}}$ & 1 
        \\ 
        \hline
        \multicolumn{1}{ V{1} r V{1}}{\multirow{3}{*}{Mean Field}}      & Embedding & 
        $1\hphantom{/n}$    & $1\hphantom{/\sqrt{n}}$                  & $1/n\hphantom{n}$             & $n\hphantom{/\sqrt{n}}$                & $1\hphantom{/\sqrt{n}}$     & 1       & $n\hphantom{/\sqrt{n}}$                & $1\hphantom{/\sqrt{n}}$     & 1       
        \\ 
        \multicolumn{1}{ V{1} r V{1}}{}                                   & Hidden    & 
        $1\hphantom{/n}$    & $1/\sqrt{n}$                             & $1/n^{1.5}\hspace{-2pt}$      & $n\hphantom{/\sqrt{n}}$                & $1/\sqrt{n}$         & $1/\sqrt{n}$ & $n^{1.5}\hphantom{/\sqrt{n}}$  & $1\hphantom{/\sqrt{n}}$ & 1 
        \\ 
        \multicolumn{1}{ V{1} r V{1}}{}                                   & Readout   & 
        $1\hphantom{/n}$    & $1/n\hphantom{n}$                        & $1/n\hphantom{n}$             & $n\hphantom{/\sqrt{n}}$                & $1\hphantom{/\sqrt{n}}$       & $1$          & $n\hphantom{/\sqrt{n}}$  & $\sqrt{n}\hphantom{/\sqrt{n}}$ & 1 
        \\ 
        \hline
    \end{tabular}
    }
    \caption{\textbf{Summary of parameterizations, their gradients and the learning rates derived in \cref{sec:theory}.} Left: Parameterizations and gradients at initialization for width $n$. Middle: Max stable per-layer learning rate scaling for each optimizer assuming ``full alignment" where $\alpha_l = u_l = 1, \omega_l = 1/2$ for all layers $l$. Right: Max stable learning rates assuming ``no alignment" where $\alpha_l = \omega_l = u_l = 1/2$ for all layers.}
    \label{tab:common_parameterizations}
    \end{footnotesize}
\end{table*}

In this paper, we take a broad perspective on the theory and practice of width scaling across parameterizations and optimizers. Our theoretical contributions consider a more general space of parameterizations which explicitly quantifies the contribution of several distinct alignment terms; under specific alignment assumptions we recover prior work as a special case. We propose a metric for alignment that we use in our empirical investigation. In addition, we develop new parameterization theory for a family of adaptive optimizers with parameter scaling, including Adafactor.

In our experiments, we measure alignment throughout training across optimizers, parameterizations and model sizes. Our measurements suggest that existing theory may be overly conservative, thereby excluding interesting parameterizations. We show that for all parameterizations and optimizers, there are theoretically motivated per-layer learning rate exponents that improve performance over global learning rate baselines, and that in numerous settings the best performing exponents would have been excluded by the alignment assumptions of prior work.

In particular, a novel per-layer learning rate prescription for standard parameterization is shown to outperform muP: we scale the learning rate for the embedding layer as $O(1)$ and the learning rate for hidden and readout layers as $O(1/n)$ for standard parameterization, and this outperforms all other combinations of parameterizations and learning rate prescriptions for Adam. In addition, while prior work emphasizes the hyperparameter transfer properties of muP specifically~\citep{yang2022tensorv}, we show that all parameterizations can perform hyperparameter transfer. We introduce constant multiplicative factors to the per-layer learning rate prescriptions and show that tuning these factors is both essential and practical: the constants can be tuned at relatively small scale and successfully reused across model sizes up to $26.8$ billion parameters, inducing substantial performance gains.

Finally, we consider the epsilon hyperparameter in Adam and similar adaptive optimizers. Our theoretical prediction that the mean-field parameterization will be most sensitive to epsilon underflow is validated in our experiments. We see significant performance improvements from three strategies to mitigate epsilon underflow, including our proposal $\emph{Adam-atan2}$ that eliminates epsilon entirely. After addressing epsilon underflow, parameterizations that are theoretically equivalent give very similar performance, illustrating that finite precision plays a practical role in the study of parameterization.

\section{Background}
\label{sec:background}

\subsection{Parameterizations and Optimizers}
We define a width-scaling parameterization as in ~\citet{yang2021tensoriv} as the prescription of the scaling exponents for three quantities on each layer: (1) the initialization variance for the parameters, (2) a parameter multiplier\footnote{The parameter multiplier is a constant that multiplies the output of the matrix multiplication in the layer during the forward pass. The trainable parameters are updated during training but the parameter multiplier is not. However, the backpropagated gradients for the parameters will include this multiplier as a term.} by which the trainable parameter weights are multiplied during the forward pass, and (3) the learning rate. It is typical for different layer types (embedding, hidden, and readout) to use different parameterizations within the same network.

We will consider all combinations of four common parameterizations and three common optimizer families. Our parameterizations, shown in \cref{tab:common_parameterizations}, include standard parameterization~\cite{neal1996priors,glorot2010understanding,he2015delving}, Neural Tangent Kernel (NTK) parameterization~\cite{jacot2018neural}, Maximal Update parameterization (muP)~\cite{yang2021tensoriv}, and Mean-Field parameterization (MFP)~\cite{mei2018mean,bordelon2022self}. Following convention, the \emph{names} of parameterizations will refer to the initialization scale and parameter multipliers, although formally speaking, the learning rate prescription is an essential element of a parameterization.

We select optimizers that represent three distinct width-scaling regimes: Stochastic Gradient Descent (SGD), Adam~\cite{kingma2014adam}, and Adafactor~\cite{shazeer2018adafactor} due to their varying relationships between the parameter, gradient, and update scales. Our theoretical perspective focuses on the width-scaling relationships between these elements and will omit more specific optimizer features like momentum and gradient or update clipping. SGD represents optimizers where the scale of the update matches the scale of the learning rate times the scale of the gradients. Adam represents adaptive optimizers where, due to the normalization by the gradient scale, the scale of the update matches the scale of the learning rate regardless of the gradient. Finally, Adafactor represents adaptive optimizers with parameter scaling that normalize the gradient similarly to Adam but then multiply by the parameter scale. This results in an update scale that matches the learning rate scale $\times$ the parameter scale; under constant learning rates, the Adafactor updates match the RMS (or Frobenius) norm of the parameters. Note that parameter scaling is the key distinction between the last two regimes: Adam plus parameter scaling falls into the Adafactor family; Adafactor with parameter scaling removed falls into the Adam family.

\subsection{Stability, nontriviality and feature learning}
\citet{yang2021tensoriv} and \citet{yang2023tensorivb} derive a system of linear constraints on the exponents of a parameterization from the two following concepts: \emph{stability}, where the activations are exactly constant scale and the logits are no more than constant scale, and \emph{nontriviality}, where the change in logits after the initialization is at least constant scale. Note these constraints are defined solely in terms of the activations and logits, so that only the forward pass is directly constrained and any constraints on backward pass quantities like gradients or updates are indirect consequences. 

In addition, they define \emph{feature learning}, where the latent representations change and adapt to the data during training, as a constant scale change after initialization in the activations directly before the readout layer. These activations are the \emph{features} or latent representations learned by the model, which, for example, could be reused with another classifier if we were to remove the readout layer. When the change in these activations is exactly constant scale, the change is meaningful in the infinite-width limit rather than becoming infinitesimally small as the width becomes large. Finally, they fully characterize the infinite-width limits of the space of width-scaling parameterizations as a dichotomy between a feature learning regime and a kernel regime.

\subsection{Alignment}\label{alignment}
As we will see in \cref{sec:theory}, the conditions for stability differ between the first and subsequent forward passes because of the learning process itself. While the initial random parameters are independent from the data distribution, correlations can develop over time because the updates carry information about the data. Such correlations cause ``alignment" between the parameters and activations, in the sense that the vectors may point in similar directions. As such, the norm of the activations after a given layer depends on three quantities: the scale of the input to the layer, the scale of the parameters in the layer, and the alignment between the parameters and the input or ``data". In a matrix multiplication, when we sum over the interior dimension $n$, this alignment contributes a scaling term that is $O(\sqrt{n})$ when there is no alignment and $O(n)$ when there is significant alignment.

The intuition for this calculation can be seen by considering the simpler case of the scaling of the inner product of two random vectors, because the entries in the product of a matrix multiplication are each vector-vector inner products themselves. As the length of the two random vectors becomes large, by straightforward application of the Central Limit Theorem the inner product is a sample from a normal distribution, so its norm has two terms coming from the mean and the variance of this distribution. The takeaway is that the mean term contributes an $O(n)$ term to the norm of the inner product and the variance term contributes $O(\sqrt{n})$. When the mean term is zero because the vectors are zero-mean and independent, then the variance term dominates and the inner product scales like $O(\sqrt{n})$. However, when the vectors are \emph{correlated} or in the worst case are identical, then the inner product scales like $O(n)$ because the coefficient to the mean term is a constant. \citet{yang2021tensoriv} refers to this idea as ``Central Limit Scaling'' versus ``Law of Large Numbers'' scaling owing to the idea that the Law of Large Numbers governs how the mean converges and the Central Limit Theorem governs how the variance converges.

This scaling affects the norm of the outgoing activations from a layer that multiplies its parameter matrix by the input to that layer. On the first forward pass, due to the random initialization the parameters cannot be aligned with the data, so the $O(\sqrt{n})$ scaling holds. However, during the first backward pass, the updates to the parameters are a function of the first batch of data, so the parameters may become correlated to the data distribution. As a result, during subsequent forward passes, we can no longer assume perfect independence between the parameters and data and instead the activations might scale with up to an $O(n)$ term.

Similar to this alignment between parameter updates and the data distribution, it is also possible to develop alignment between the parameters in two different layers in the network. For example, the backpropagated gradients used to update an earlier layer are a function of the parameters in later layers, which can introduce correlation between the earlier layer updates and the later layer parameters. The consequence of either type of alignment is that the learning rate needs to \emph{counteract} the $O(\sqrt{n})$ or $O(n)$ term, so the maximal stable learning rates can be \emph{smaller} by a factor of $O(\sqrt{n})$ when there is significant alignment than when there is none.

\section{Theoretical Contributions}
\label{sec:theory}

\begin{figure*}[ht!]
    \centering
    \includegraphics[width=0.8\textwidth]{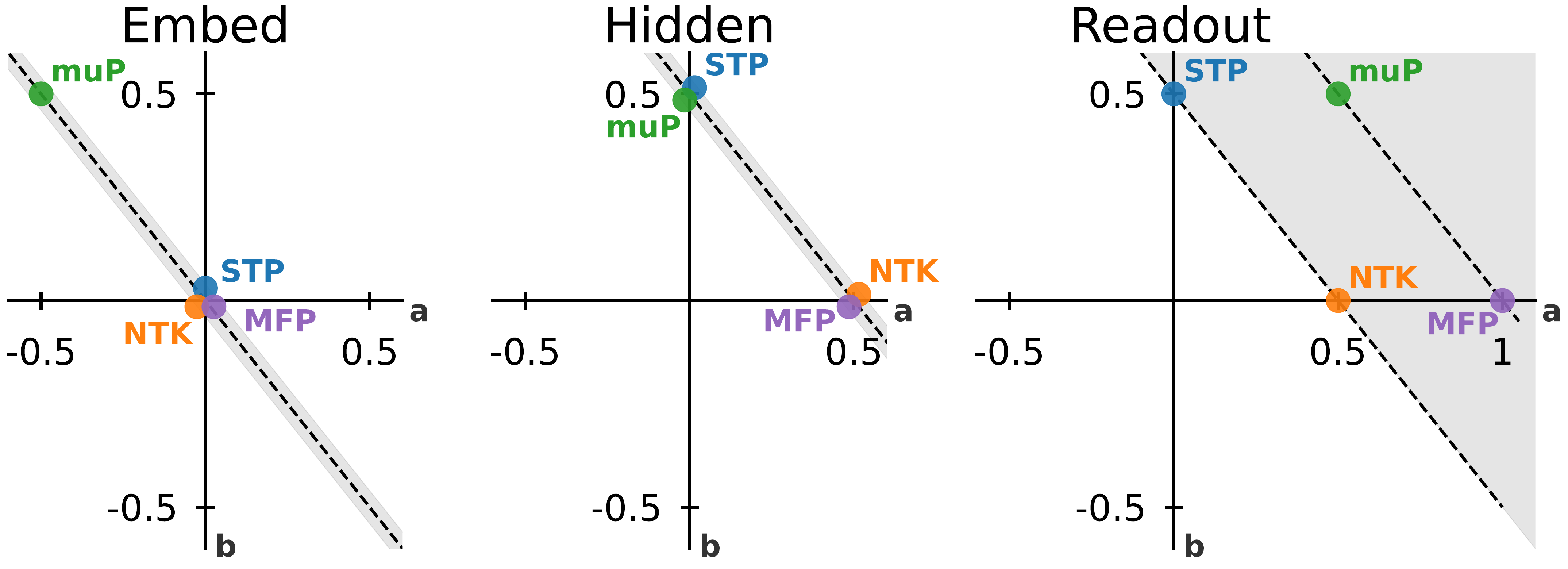}
    \caption{\textbf{The four parameterizations occupy two equivalence classes at initialization, which differ only in the readout layer}. Each parameterization is plotted for each layer type at $(a_l, b_l)$ where $a_l$ is the negative parameter multiplier exponent and $b_l$ is the negative initialization standard deviation exponent. The black dashed lines span the equivalence classes for each layer. The region where parameterizations are \emph{stable} is highlighted in gray: this is the line $a_1 + b_1 = 0$ for the embedding layer, the line $a_l + b_l = 1/2$ for hidden layers, and the region $a_{L+1} + b_{L+1} \geq 1/2$ for the readout layer. For equivalence during training, the learning rates must also obey the optimizer-specific equivalence relations.}
    \label{fig:equivalence_classes}
\end{figure*}

In this section we make four theoretical contributions. First, we define a general space of width-scaling parameterizations that explicitly quantifies the contribution of three alignment terms. Rather than making specific assumptions about alignment and then deriving which parameterizations are stable and nontrivial under those assumptions, as in \citet{yang2021tensoriv} and \citet{yang2023tensorivb}, we propose general stability and nontriviality constraints as a function of those alignment variables. Second, we propose theory for Adafactor or other adaptive optimizers using parameter scaling. Third, for all parameterizations $\times$ optimizers, we find the maximum stable learning rate for each layer type as a function of the alignment terms, and compute the learning rate exponents under two specific alignment assumptions, which we refer to as ``full alignment'' and ``no alignment''. While the alignment assumptions in \citet{yang2021tensoriv} prevent standard and NTK parameterizations from feature learning regardless of the per-layer learning rate prescription, under our assumptions of either full alignment or no alignment, all parameterizations have per-layer learning rates in the feature learning limit. Fourth, we propose the \emph{alignment ratio} metric that we will use for empirical investigation, which measures the contribution of alignment to the activations during training.

\subsection{Model and Notation}
Following a similar model and notation as \citet{yang2021tensoriv}, we consider a multilayer perceptron with $L$ hidden layers, input and output dimensionality $d$, hidden layer dimensionality $n$, and nonlinearity $\phi: \R \rightarrow \R$. The weight matrices are denoted:
\begin{itemize}
    \item $W_1 \in \R^{n \times d}$ for the embedding layer
    \item $W_2, \ldots W_L \in \R ^ {n \times n}$ for the hidden layers, and
    \item $W_{L+1} \in \R^{d \times n}$ for the readout layer.
\end{itemize}

The parameterization for each layer $l$ is specified by three values $\{a_l, b_l, c_l\}$, where:
\begin{itemize}
    \item the parameter multiplier is $n^{-a_l}$,
    \item the parameter initialization is $W_l \sim \mathcal{N}(0, n^{-2b_l})$, and
    \item  the learning rate $\eta_l \propto n^{-c_l}$ with width-independent constant of proportionality that we omit here.
\end{itemize}

For an input $x \in \R^d$, the model has activations $z_1, \ldots z_L$ and outputs logits $z_{L+1}$:
\begingroup
\begin{align*}
    z_1 &= \phi(n^{-a_1} W_1 \cdot x)\\
    z_l &= \phi(n^{-a_l} W_l \cdot z_{l-1}), \quad l \in [2, L]\\
    z_{L+1} &= n^{-a_{L+1}} W_{L+1} \cdot z_L
\end{align*}
\endgroup

In addition, we define $\Delta W_l^t$ and $\Delta z_l^t$ to be the change in parameters and activations, respectively, in layer $l$ between initialization and step $t$. We omit the time superscript throughout this section when it is clear from context or the statement holds for any $O(1)$ value of $t$.

We are interested in the scaling behavior of various quantities as we increase the \emph{width} or hidden layer dimensionality $n$, while other dimensions are held constant. In particular, we assume input and output dimensionality $d$, the depth $L$, and the number of training steps $T$ are fixed and constants with respect to width. We omit batch size dimensions, assuming the batch size is one. We assume the nonlinearity $\phi$ has bounded derivative so its contribution is negligible in the large-width limit, and as such treat it as the identity function in our derivations. Throughout this section, we refer to the ``scale" of quantities, meaning their exponents with respect to width in the large-width limit. For formal definitions, additional assumptions, and a full derivation see \cref{app:theory}.

\subsection{Equivalence classes}
\label{sec:equivalence_relations}
These parameterizations occupy equivalence classes because in any layer we can ``factor out'' a constant term from the parameter initialization into the parameter multiplier, which exactly preserves the output of the forward pass while multiplying the gradients by this constant. This change in the gradients can then be ``corrected for'' by modifying the learning rate in an optimizer-specific manner.

In this one-dimensional symmetry group parameterized by $\theta$, to preserve the forward pass, regardless of the optimizer, apply
\begingroup
\begin{align*}
    &a_l \leftarrow a_l + \theta\MoveEqLeft[1]\\
    &b_l \leftarrow b_l - \theta.\\
\intertext{Then specific to the optimizer, to preserve the effect of the backwards pass, correct the learning rate according to}
    \textrm{SGD:}\quad&c_l \leftarrow c_l - 2\theta\\
    \textrm{Adam:}\quad&c_l \leftarrow c_l - \theta\\
    \textrm{Adafactor:}\quad&c_l \leftarrow c_l.
\end{align*}
\endgroup
In particular, under the right learning rates, our four parameterizations occupy two equivalence classes: standard and NTK are equivalent and muP and mean-field parameterization are equivalent. In \cref{fig:equivalence_classes}, we visualize the subspaces spanned by the equivalence classes for each layer type: each parameterization is plotted at $(a_l, b_l)$ for each layer $l$, and the dashed lines span the equivalence class $(a_l + \theta, b_l - \theta)$ for all values of $\theta$. We see that for both the embedding layer and the hidden layers, all four parameterizations have equivalent initializations. However, the readout layer has two distinct equivalence classes: standard and NTK parameterizations have $a_{L+1} + b_{L+1} = 1/2$, corresponding to constant scale logits at initialization, whereas muP and MFP have $a_{L+1} + b_{L+1} = 1$, resulting in logits that scale like $1/\sqrt{n}$ at initialization. This difference, specifically the shift by a factor of $\sqrt{n}$ in the readout layer, is the key difference between the standard + NTK equivalence class and the muP + MFP equivalence class.

In this paper, we will consider all four parameterizations separately, as these equivalences hold only under infinite precision, while neural networks regularly encounter finite-precision effects. These equivalences were observed for SGD and Adam in \citet{yang2021tensoriv} and \citet{yang2023tensorivb} respectively, and we propose this equivalence for Adafactor.

\newcolumntype{C}{>{\centering\arraybackslash}}

\begingroup
\renewcommand{\arraystretch}{2.5}
\begin{table*}[h!]
\centering
\adjustbox{scale=0.86}{
\begin{footnotesize}
\begin{tabular}{c | c | c| c}
 & \normalsize{\textbf{SGD}} & \normalsize{\textbf{Adam}} & \normalsize{\textbf{Adafactor}}\\ \hline
     \multirow{3}{*}[1.2ex]{\rotatebox[origin=c]{90}{\parbox[l]{1.75cm}{\centering{\textbf{Stability at initialization}}}}} & \multicolumn{3}{c}{$a_1 + b_1 = 0$}\\[-1.5ex]
     & \multicolumn{3}{c}{$a_l + b_l = 1/2$ for $l \in [2, L]$} \\[-1.5ex]
     & \multicolumn{3}{c}{$a_{L+1} + b_{L+1} \geq 1/2$} \\ \hline
   \multirow{4}{*}{\rotatebox[origin=c]{90}{\parbox[c]{2.4cm}{\centering\textbf{Stable activations during training}}}} & $r_1 \coloneqq g_1 + a_1 + c_1 \geq 0$    &  $r_1 \coloneqq a_1 + c_1 \geq 0$    &   $r_1 \coloneqq c_1 \geq 0$\\
  & $r_l \coloneqq \min \begin{cases}g_l + a_l + c_l - \alpha_l\\ g_l + a_l + c_l + r_{l-1} - u_l\\ 1/2 + r_{l-1} - \omega_l\end{cases}\hspace{-0.5em}\geq 0$ & $r_l \coloneqq \min \begin{cases}a_{l} + c_{l} - \alpha_l\\a_{l} + c_{l} + r_{l-1} - u_{l}\\1/2 + r_{l-1} - \omega_l\end{cases}\hspace{-0.5em} \geq 0$    &    $r_l \coloneqq \min \begin{cases}1/2 + c_{l} - \alpha_l\\1/2 + c_{l} + r_{l-1} - u_{l}\\ 1/2 + r_{l-1} - \omega_l\end{cases}\hspace{-0.5em}\geq 0$\\
  & \multicolumn{1}{l|}{where $g_i \coloneqq$} & & $c_l \geq 0$\\[-3ex]
  & $\max(a_{L+1} + b_{L+1}, 2a_{L+1} + c_{L+1}) + a_i$& & \\ \hline
     \multirow{2}{*}[2.5ex]{\rotatebox[origin=c]{90}{\parbox[c]{2.4cm}{\centering\textbf{Stable logits during training}}}} & $\min \begin{cases}a_{L+1} + b_{L+1} + r_L - \omega_{L+1}\\ 2a_{L+1} + c_{L+1} - \alpha_{L+1} \\ 2a_{L+1} + c_{L+1} + r_L - u_{L+1} \end{cases}\hspace{-0.5em}\geq 0$ \rule{0pt}{8ex} &  $\min \begin{cases}a_{L+1} + b_{L+1} + r_L - \omega_{L+1}\\a_{L+1} + c_{L+1} - \alpha_{L+1}\\a_{L+1} + c_{L+1} +r_L - u_{L+1} \end{cases}\hspace{-0.5em}\geq 0$     &   $\min \begin{cases}a_{L+1} + b_{L+1} + r_L - \omega_{L+1}\\a_{L+1} + b_{L+1} + c_{L+1} - \alpha_{L+1}\\a_{L+1} + b_{L+1} + c_{L+1} +r_L - u_{L+1} \end{cases}\hspace{-0.5em}\geq 0$\\
  & & & $c_{L+1} \geq 0$ \\\hline
\end{tabular}
\end{footnotesize}}
\vspace{2pt}
\caption{Constraints for stable parameterizations at initialization and during training.}
\label{tab:stability_training_constraints}
\end{table*}
\endgroup

\subsection{Alignment-General Space of Parameterizations}
We now propose a general space of parameterizations where we define three alignment variables and derive the space of parameterizations that are stable and nontrivial as a function of these variables. We define a parameterization to be \emph{stable} if the activations have exactly constant scale and the logits have at most constant scale throughout training, and to be \emph{nontrivial} if the change in logits after initialization has at least constant scale.

The activations change after initialization due to both changes in the parameters in the previous layer and changes in the activations immediately prior to that layer. Specifically, starting from the second forward pass, the activations for layer $l$ are computed as
\begin{align*}
    z_l &= n^{a_l} (W_l + \Delta W_l)(z_{l-1} + \Delta z_{l-1})\\
    &= n^{a_l} (W_lz_{l-1} + \Delta W_lz_{l-1} + W_l\Delta z_{l-1} + \Delta W_l\Delta z_{l-1}).
\end{align*}
The first term $W_l z_{l-1}$ in the expanded sum contains the initial random parameters and initial activations, which are not aligned. The remaining three terms in the sum may have alignment as they result from updates that might be aligned to the data distribution or other parameters in the model.

We will define $\alpha_l, \omega_l,$ and $u_l$ to be the exponents of the alignment contributions from these three terms, where the alignment exponent quantifies how the norm of the product scales compared to the norm of the factors. We define
\begin{itemize}
    \item $\alpha_l$ to be the alignment exponent for $\Delta W_l z_{l-1}$,
    \item $\omega_l$ to be the alignment exponent for $W_l \Delta z_{l-1}$, and
    \item $u_l$ to be the alignment exponent for $\Delta W_l \Delta z_{l-1}$.
\end{itemize}
so that, for example, $\norm{\Delta W_l z_{l-1}}$ scales like $n^{\alpha_l} \norm{\Delta W_l} \norm{z_{l-1}}$. With this definition, $\alpha_l = 1/2$ corresponds to no alignment in the $\Delta W_l z_{l-1}$ term and $\alpha_l=1$ corresponds to high alignment.

We next define a feature learning residual quantity $r_l$ that measures how far the parameterization is from the feature learning regime. For each layer $l$ in $[1, L]$, we define $r_l$ as the negative exponent of the scale of $\Delta z_l$, where $\Delta z_l$ is the change in activations following layer $l$ during training. To preserve stability, this change cannot exceed constant scale, so $r_l$, as the negative exponent, cannot be less than zero. Feature learning, where the change in activations immediately prior to the readout layer has constant scale, then corresponds to $r_L = 0$ exactly. Conceptually, feature learning occurs if at least one of the embedding or hidden layers contributes at least one constant scale term to the activations. Lastly, for convenience, we define $g_l$ to denote the negative exponent of the gradient scale.

\subsubsection{Stability at initialization}
To derive the stability constraints, which are shown in \cref{tab:stability_training_constraints}, we will first derive constraints on the initialization scale and parameter multipliers so that the parameterization is stable during the first forward pass. In the next subsection, we will derive constraints for stability during training that ensure the \emph{change} in activations after every layer has \emph{at most} constant scale and, similarly, we will bound the change in logits to ensure they do not exceed constant scale. For a more detailed derivation of the stability constraints, see \cref{app:theory}. For nontriviality, we include the derivation and constraints in \cref{app:theory_nontriviality} and \ref{app:theory_summary_constraints}.

\label{sec:stability_init}
During the first forward pass, the conditions for stability depend only on the parameter initialization and the parameter multipliers, and not on the optimizer, learning rates or alignment variables because no updates have occurred yet. For all optimizers, for input data $x$ with constant scale, the constraints for stability at initialization are:
\begingroup
\begin{align*}
    a_1 + b_1 &= 0\\
    a_l + b_l &= 1/2, \hquad l \in [2, \ldots, L]\\
    a_{L+1} + b_{L+1} &\geq 1/2
\end{align*}
\endgroup

\subsubsection{Stability during training}
\label{sec:stability_training}
During the second and subsequent forward passes, the constraints for stability are specific to the optimizer. In \cref{app:theory}, we derive the constraints on the parameter multipliers, parameter initialization, and learning rates under each optimizer that ensure stability during training, that is, constant activations and at most constant logits. In our notation, recall that $r_l$ is the negative exponent of the scale of $\Delta z_l$ and $g_l$ to is the negative exponent of the gradients with respect to parameters $W_l$.

For intuition, this derivation first considers the second forward pass, and assuming the constraints for stability at initialization hold, iteratively adds constraints starting with the embedding layer and working up to the readout layer, where each constraint ensures stability of that layer assuming stability of all earlier layers. We compute the scale of the change in activations in each layer, and bound its exponent to be at most zero. Note that in our constraints, this is written to bound the \emph{negative} exponent to be \emph{at least} zero. Similarly, we constrain the logits to stay at most constant scale in the second forward pass. The constraints derived from the second forward pass are included in \cref{app:theory_second_forward}.

Next, we consider the third forward pass, because the readout parameters may change in scale between the first update and second update. This slightly modifies the constraints to use the maximum possible scale of the readout parameters, which may come from the readout initialization or the readout update depending on the parameterization. This produces the final set of stability constraints, as after the third forward pass the scale of each quantity remains the same over a constant number of training steps.

We present these constraints for stability during training in Table \ref{tab:stability_training_constraints}. The set of constraints has some similarity across optimizers. The change in activations following a hidden layer $l$ is computed as the sum of three terms, $\Delta W_l z_{l-1}$, $W_l \Delta z_{l-1}$ and $\Delta W_l \Delta z_{l-1}$. As the term with the maximum scale dominates the exponent, the value of $r_l$ is the minimum over three expressions each coming from one of these terms. The constraint $r_l \geq 0$ then preserves stability by preventing any of these terms from exceeding constant scale, where when $r_L = 0$ exactly, we are in the feature learning regime. To highlight the differences across optimizers, we note some of the SGD constraints include the term $g_l$, as the SGD update depends on the scale of the gradient. Compared to SGD, the Adam constraints result from removing the contribution of $g_l$ due to the normalization of the gradient in Adam, even in the readout layer where $g_{L+1} = a_{L+1}$. Then, compared to Adam, the constraints for Adafactor have additional appearances of $b_l$ as a result of the parameter scaling; in some places this simplifies due to substituting $a_1 + b_1 = 0$ or $a_l + b_l = 1/2$. Lastly, there is an additional constraint $c_{l} \geq 0$ required for Adafactor to prevent the parameters from growing exponentially with the number of training steps as a result of parameter scaling.

\subsubsection{Prior Work as a Special Case}
We exactly recover the stability and nontriviality constraints in \citet{yang2021tensoriv} for SGD and \citet{yang2023tensorivb} for Adam\footnotemark{} if and only if
\begin{itemize}
    \item $\alpha_l = 1 \quad \forall l \in [2, L+1]$,
    \item $\omega_l = 1/2 \quad \forall l \in [2, L]$, \quad and
    \item $\omega_{L+1} = 1$.
\end{itemize}

The choice of $\alpha_l = 1$ on all layers depends on the assumption that updates to the parameters $\Delta W_l$ are aligned with the activations $z_{l-1}$ due to alignment between the parameter updates and the data distribution. We will investigate settings with and without this assumption.

\footnotetext{\citet{yang2023tensorivb} carefully considers the role of the epsilon hyperparameter in Adam. This correspondence holds if we assume what they call \emph{faithfulness}, which is equivalent to setting epsilon following the per-layer epsilon prescription we implement in \sref{sec:results_epsilon}. It also holds if we disregard epsilon (i.e. consider epsilon to be zero) and view Adam as perfectly scale-invariant.}

We note that the assumption $\omega_{L+1} = 1$, which we will relax, is at the very core of the theoretical motivation for muP. Recall that $\omega_{L+1}$ is the alignment exponent for $W_{L+1} \Delta z_{L}$, where $W_{L+1}$ is the readout layer initialization and $\Delta z_{L}$ is the change in activations immediately prior to the readout layer resulting from changes in the parameters in earlier layers. In theory, this alignment could develop when the gradients propagated to earlier layers contain information from the initialization parameters in later layers. However, we also note the assumption of $\omega_l = 1/2$ on all layers except the readout layer assumes that the analogous alignment does not develop in any earlier layers.

A key consequence of the $\omega_{L+1} = 1$ assumption is that neither standard nor NTK parameterizations are able to achieve feature learning \emph{regardless of what learning rate prescription is used}. Recall that feature learning corresponds to $r_L = 0$, where the change in activations $\Delta z_L$ is exactly constant scale, and that standard and NTK parameterizations both have $a_{L+1} + b_{L+1} = 1/2$. Therefore, due to the constraint on the logits that $a_{L+1} + b_{L+1} + r_L - \omega_{L+1} \geq 0$, it is not possible when $\omega_{L+1}=1$ for standard and NTK parameterizations to have $r_L = 0$ regardless of how it might be induced. However, the shift in the readout layer in muP and MFP so that $a_{L+1} + b_{L+1} = 1$ allows these parameterizations to attain feature learning when $\omega_{L+1} = 1$.

\subsubsection{Maximum Stable Learning Rate Exponents}
We will next define two specific sets of alignment assumptions in terms of $\alpha_l$, $\omega_l$, and $u_l$ and then derive the maximum stable learning rates shown in \cref{tab:common_parameterizations} for each layer and parameterization under these assumptions. To select these assumptions, we first consider the types of alignment measured by each variable and how the alignment assumptions dictate the maximum stable learning rates.

The $\Delta W_{l} z_{l-1}$ term, whose alignment is measured by $\alpha_l$, may have alignment between the parameter updates and data distribution. If we unroll the layers in the term $\Delta W_l  z_{l-1} = \Delta W_l \cdot W_{l-1} \cdot W_{l-2} \ldots W_1 \cdot x$, since the parameters $W_1, \ldots, W_{l-1}$ are from the random initialization, any alignment comes from the matrix multiplication between the parameter updates $\Delta W_{l}$ and the data $x$.

In contrast, for the $W_{l} \Delta z_{l-1}$ term, whose alignment is measured by $\omega_l$, the alignment occurs not between parameters and data, but between the updates to parameters in earlier layers and the initialization parameters in the later layer $l$. Recall that this alignment could develop as the backpropagated gradients, which are used to update the earlier layers, depend on the initialization parameters in the later layers. The final term $\Delta W_{l} \Delta z_{l-1}$, whose alignment is measured by $u_l$, may contain both kinds of alignment, between the parameters and data or between parameters in different layers.

In our alignment settings, we will assume that the alignment between parameters in different layers stays small, and focus on exploring the range of possible degrees of parameter-to-data alignment. Specifically, we assume that $\omega_l = 1/2$ on all layers including the readout layer $\omega_{L+1} = 1/2$. As a consequence, we will be able to consider feature learning versions of all four parameterizations.

For the assumptions on $\alpha_l$ and $u_l$ that measure alignment between parameters and data, we note that in the stability constraints, $\alpha_l$ and $u_l$ always appear in tandem in pairs of constraints, where in fact the constraints could be rewritten entirely in terms of the single quantity $\max(\alpha_l, u_l - r_{l-1})$. When using the maximum stable learning rates on earlier layers, $r_{l-1}$ will be zero, so the learning rate exponents are constrained by the maximum of $\alpha_l$ and $u_l$. We will therefore consider the two extremes, and make two choices of alignment assumptions: ``full alignment" where $\alpha_l = u_l = 1$ and $\omega_l = 1/2$ and ``no alignment" where $\alpha_l = u_l = \omega_l = 1/2$. Intermediate choices where $\alpha_l = u_l$ takes a value between $1/2$ and $1$ would also be interesting for future work, but scaling studies at practical sizes may not have sufficient resolution to distinguish smaller variations in the exponents.

In \cref{tab:common_parameterizations}, we compute the maximum stable per-layer learning rate exponents under these two assumptions of full alignment and no alignment. Our full alignment per-layer learning rate prescriptions for standard and NTK differ from prior work and are in feature learning limits under our assumptions. For muP and MFP, our full alignment assumptions result in learning rate exponents that coincide with prior work as relaxing the $\omega_{L+1}$ constraint does not impact these parameterizations. Our no alignment prescriptions are also in feature learning limits for all parameterizations, and again differ from prior work that assumed $\alpha_l = 1$.

\begin{figure*}[ht!]
    \centering
    \includegraphics[clip=true,trim={6 15 6 8},width=\textwidth]{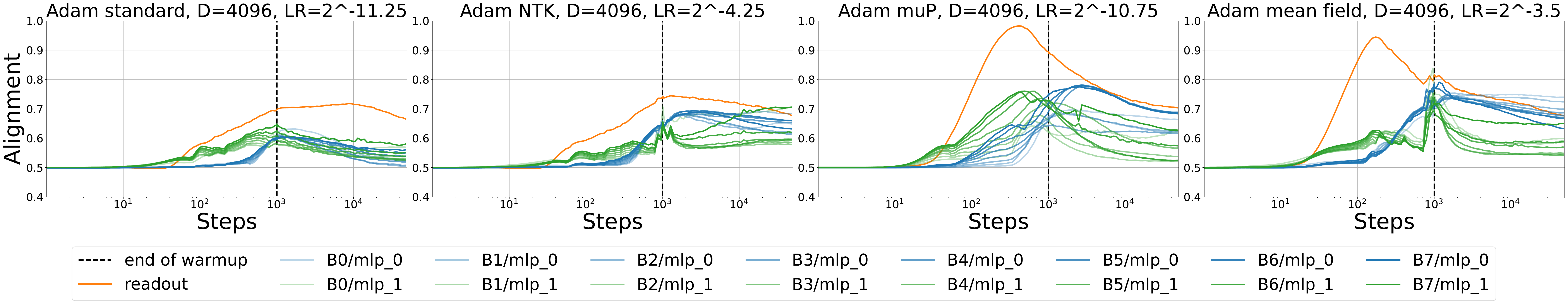}
    \caption{\textbf{Alignment is intermediate and highly dynamic throughout training, with parameterization-specific patterns.} The log alignment ratio metric in readout and hidden (MLP) layers across training steps for each parameterization, for Adam $1.9B$ parameter models ($H=32, D=4096, B=256$) using optimal global learning rates. Blue and green curves are for the first and second MLP layers, respectively, in each Transformer block. Transformer blocks are denoted B0 through B7 in the legend. Orange curves are the readout layer.}
    \vspace{-8pt}
    \label{fig:alignment}
\end{figure*}

\subsection{Alignment Ratio}
 The alignment variables $\alpha_l, \omega_l,$ and $u_l$ quantify the alignment contributions to the activations $z_l$ from the individual terms in the expanded sum $(W_l + \Delta W_l)(z_{l-1} + \Delta z_{l-1}) = W_lz_{l-1} + \Delta W_lz_{l-1} + W_l\Delta z_{l-1} + \Delta W_l\Delta z_{l-1}$. However, we note that in practice the alignment in the terms in this sum may interfere constructively or destructively, and the single alignment quantity that actually governs the scale of the activations $z_l$ is the alignment between $(W_l + \Delta W_l)$ and $(z_{l-1} + \Delta z_{l-1})$.
 
 We therefore propose a metric that measures this alignment, between $(W_l + \Delta W_l)$ and $(z_{l-1} + \Delta z_{l-1})$, in order to understand empirically how alignment that accumulates during training is contributing to the activation scales throughout the model. This metric is defined on each dense layer and quantifies the contribution from alignment between the current parameters and the current pre-layer activations on the scaling exponent of the post-layer activations.

Consider a neural network layer $l$ at training step $t$ with parameters $W_l^t \in \R^{\textrm{fan-out} \times \textrm{fan-in}}$, and pre-layer activations $z_{l-1}^t \in \R^{\textrm{fan-in}}$. The alignment between the pre-layer activations and the rows of the parameter matrix dictates the scaling term contributed when summing over the fan-in dimension in the matrix multiplication, giving an alignment term with the fan-in as the base of the exponent.

\begingroup
\definition We define the \emph{log alignment ratio} as\begin{tightcenter}$A_l^t = \log_{\textrm{fan-in}} \frac{\norm{ W_l^t z_{l-1}^t}_{RMS}}{\norm{W_{l}^t}_{RMS}\norm{z_{l-1}^t}_{RMS}} \in \mathbb{R}$,\end{tightcenter} where the norm is the RMS norm\footnote{An equivalent definition for the log alignment ratio in terms of Frobenius norms is $A_l^t = 1 + \log_{\textrm{fan-in}} \frac{\norm{ W_l^t z_{l-1}^t}_{F}}{\norm{W_{l}^t}_{F}\norm{z_{l-1}^t}_{F}}$.}.
\endgroup

We note that while our theoretical derivations consider exponents in the infinite-width limit, this metric is intentionally defined at finite size so that we can measure it in practice. We will see in \cref{sec:results_alignment} that this metric shows convergence within our model sizes suggesting that in practical settings this finite size approximation is indicative of large-width behavior.

\section{Experiments}
We investigate the role of alignment, per-layer learning rate exponents and constant factors, and the epsilon hyperparameter by running \emph{tens of thousands} of experiments in a Transformer language model across all combinations of the three optimizers, four parameterizations, learning rate sweeps with a granularity of $2^{0.25}$ or $2^{0.5}$, and fourteen model widths ranging up to $26.8$ billion parameters.

We use the NanoDO decoder-only Transformer architecture~\citep{nanodo} employing learned positional embeddings, pre-layer norm~\citep{ba2016layer,xiong2020layer}, and GeLU nonlinearity~\citep{hendrycks2016gaussian}. All models are trained on the C4 dataset~\citep{t5}. To scale the width, we fix the attention head dimension $h = 128$ and co-scale the model dimension $D$ and the number of attention heads $H$ so that $D = H \times h$. All experiments use a fixed batch size of 256, context length of 512 and depth of 8 Transformer blocks. Our fourteen model widths range from $D=128$ to $D=16,384$ corresponding to a range of $9.9$ million to $26.8$ billion parameters.

All experiments in the main text of the paper train for $50{,}000$ training steps and do not use weight decay. The learning rate schedule for all experiments uses linear warmup of $1{,}000$ steps followed by a cosine decay schedule, with initial and final learning rates of $0.0$. We include additional experiments for compute-optimal training horizons in in \cref{app:fixed_vs_compute_opt} that show that moving from the fixed step setting to the compute optimal setting likely requires sharper decay in the learning rate exponents. In addition, we include experiments in \cref{app:weight_decay} that suggest our conclusions should transfer to settings using small amounts of weight decay. See \cref{app:expt_details} for additional experiment and implementation details.

\subsection{Alignment Experiments}
\label{sec:results_alignment}

We measure the log alignment ratio throughout training for three model sizes for each parameterization $\times$ optimizer, using a global learning rate that is close to optimal. For Adam, alignment values for the readout and MLP layers are shown for model dimension $D = 4096$ in Figure \ref{fig:alignment} and full results including the other optimizers and the dense layers within the attention block (query, key, value, and output projection dense layers) are included in \cref{app:alignment_expts}. All experiments use batch size $B=256$. As expected, the measured alignment values start at 0.5 due to the independence between the random initialization and the data, and change during training as parameters become aligned with the data or parameters in earlier layers.

The alignment values vary significantly across the training horizon and the trajectories depend heavily on the parameterization and layer type. We see similar values across three model sizes, suggesting that these model sizes are sufficiently large for our measurements to be indicative of large-width behavior. For SGD, the results show high instability and are difficult to interpret; one consistent pattern is that NTK and MFP have almost no alignment and STP and muP have low amounts. Adam and Adafactor show matching trends: the readout layer has the highest peak among the layers, with high peak readout alignment above 0.9 in muP and mean-field parameterization and only moderate peak readout alignment between 0.7 and 0.8 in standard and NTK parameterizations. In the MLP layers, alignment in all parameterizations is moderate and does not exceed 0.8.

\begin{figure*}[ht]
    \centering
    \includegraphics[width=\textwidth]{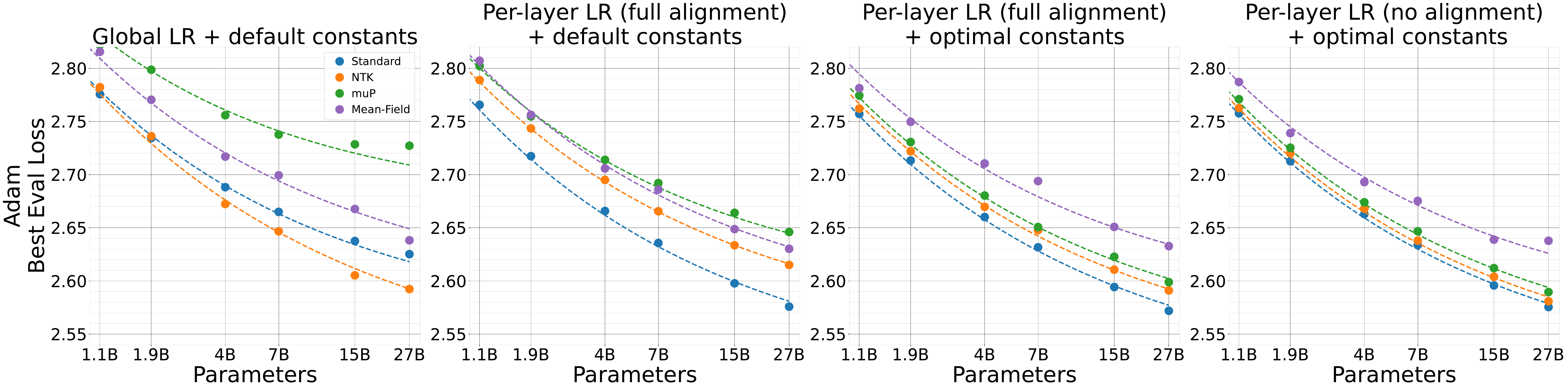}
    \caption{\textbf{All parameterizations for Adam benefit from per-layer learning rates and tuning per-layer constant learning rate multipliers.} Eval loss comparisons for all parameterizations using Adam across a sequence of interventions. From left to right panels: (a) global lr exponents + default constants, (b) per-layer lr exponents assuming full alignment + default constants, (c) per-layer lr exponents assuming full alignment + optimal constants, (d) per-layer lr exponents assuming no alignment + optimal constants.}
    \label{fig:adam_additive_interventions}
\end{figure*}

The high readout alignment early in training that is specific to muP and MFP parameterizations may result from the relationship between the readout updates and readout initialization. In standard and NTK parameterizations, the scale of the readout update matches the scale of the readout initialization, whereas in muP and MFP, the scale of the update is larger than the scale of the initialization. At each step, the parameter updates may be highly aligned with the data distribution, resulting in high readout alignment early in training because these highly aligned updates dominate the readout parameters in muP and MFP. However, as more of these updates accumulate, the alignment actually decreases to a moderate level, as the sum of the accumulated updates may be less aligned than the individual updates. In contrast, in standard and NTK parameterizations, the individual updates may still be highly aligned with the data distribution without resulting in high readout alignment as they don't dominate over the readout initialization.

Overall, these results indicate that the alignment assumptions in ~\citet{yang2021tensoriv,yang2023tensorivb} may be overly conservative; if in practice alignment contributes less than one to the activation exponents, then the learning rate exponents can be larger. However, even given these measurements for alignment, it is not obvious how exactly the learning rate exponents that control the base or peak learning rate should be adjusted. First, we see that alignment is a dynamical quantity that varies widely throughout training: even when the readout alignment is close to maximum early in training for muP and MFP, it is much lower for the vast majority of training steps. In addition, alignment varies across layers within the same layer type, which typically use the same learning rate or at least the same learning rate exponent. Further, the interaction between the alignment measurement and learning rate schedule is complex. The learning rate schedule likely influences the alignment measurements, and alignment likely influences the optimal learning rate schedule: one possible role of learning rate schedules is that the decay counteracts alignment that develops later in training. Even if we used alignment measurements from experiments under one set of learning rates to inform adjustments to the learning rate exponents, this would induce an iterative loop where the adjusted learning rates would then affect the alignment. We will therefore take an empirical approach to determine how alignment should influence the learning rate exponents, and consider several choices of alignment assumptions for the per-layer learning rate experiments in the following section.

\subsection{Per-layer Learning Rates}
\label{sec:results_per_layer}
All parameterizations have per-layer learning rate exponents specific to the optimizer and the choice of alignment assumption, as shown in \cref{tab:common_parameterizations}. In this section, we empirically validate these theoretical learning rate exponent prescriptions and investigate the impact of tuning the per-layer constant factors on all combinations of parameterizations $\times$ optimizers. We compare against global learning rates as a baseline: while global learning rates are in most cases not theoretically principled, they are the overwhelmingly dominant paradigm in practice. In most settings, the theoretically ideal learning rate exponents differ across layers, implying a mismatch in at least some layers when using global learning rates. However, there are certain cases, such as muP + SGD + full alignment, or Adafactor + any parameterization + no alignment, where the theoretically motivated per-layer exponents happen to be the same in all layers so that global learning rate actually coincides with the theoretical prescription. We include experiments using the theoretical prescriptions from both the full alignment and no alignment settings since our empirical alignment measurements show intermediate and highly dynamic values. We first present the series of experiments for Adam, followed by the results for SGD+momentum and Adam+parameter scaling. 

Additional results are included in the appendix including additional ablations (\cref{sec:app_per_layer_lr_results}), eval losses for all settings for the six largest model sizes (\cref{tab:appendix_table}) and learning rate sweeps for all settings for SGD (\sref{sec:app_lr_sweeps_sgd}), Adam (\sref{sec:app_lr_sweeps_adam}) and Adam + parameter scaling (\sref{sec:app_lr_sweeps_adam_ps}). 

\begin{figure*}[ht]
    \centering
    \includegraphics[width=\textwidth]{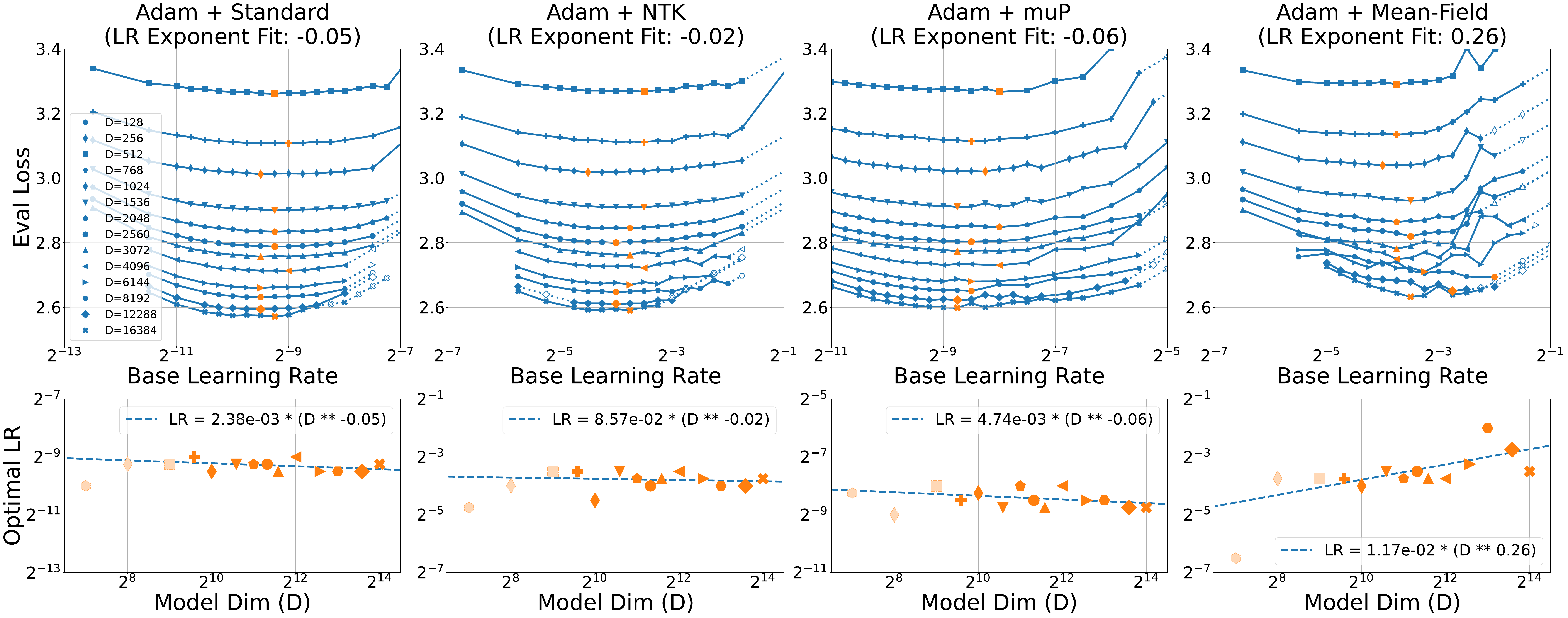}
    \vspace{2pt}
    \caption{\textbf{All parameterizations can perform hyperparameter transfer with the right per-layer learning rate exponents.} Top row = learning rate sweep for all parameterizations using Adam with per-layer learning rates assuming full alignment and optimal constants. The LR scaling is fully encapsulated by the per-layer LR exponents so the base learning rate is consistent across model widths. Bottom row = power law fit of optimal LR vs model dim, with exponents close to zero indicating the same base LR can be reused at all model widths. Only mean-field parameterization deviates slightly from zero, which is improved by addressing epsilon underflow in \cref{sec:results_epsilon}.}
    \label{fig:adam_hyperparameter_transfer}
\end{figure*}

\subsubsection{Adam}

Our first experiment compares per-layer learning rates when assuming full alignment against the baseline of global learning rates. For all experiments, we select a base model dim of $b = 1024$ and define the learning rate in layer $l$ as $\eta_l = \beta_n \cdot \frac{n}{b} ^ {-c_l}$. We perform a one-dimensional sweep of $\beta_n$ at each model dim $n$ to determine the best value and report the eval loss from this best $\beta_n$. For global learning rates, $c_l = 0$ for all $l$, and for per-layer learning rates, $c_l$ follows \cref{tab:common_parameterizations}. Since there is no contribution from the learning rate exponents at the base model size, the global and per-layer learning rate settings coincide exactly at the base model dim and differ only when scaling away from the base model size. For Adam, the per-layer results in \cref{fig:adam_additive_interventions}(b), compared with the global learning rate results in \cref{fig:adam_additive_interventions}(a), show that standard, muP and mean-field parameterizations all improve significantly with per-layer learning rates. In contrast, NTK actually performs worse with the prescribed per-layer exponents: we note that these exponents assume full alignment and return to this result later in this section.

We next introduce constant multiplicative factors to the per-layer learning rates and propose a hyperparameter transfer strategy where we tune these constants at small scale and reuse them across model sizes. Again using the base model dim of $b = 1024$, we now define the learning rate in layer $l$ as $\eta_l = \beta_n \cdot \gamma_l \cdot \frac{n}{b} ^ {-c_l}$ where $\gamma_l$ is this constant factor that should be determined empirically. We use the same learning rate within each layer type, so we need to determine $\gamma_1$ for the embedding layer, $\gamma_h$ for the hidden layers, and $\gamma_{L+1}$ for the readout layer. The previous experiments correspond to setting these constants equal to one by default. We continue to sweep one dimension at each model size to determine $\beta_n$, so we note the choice of $(\gamma_1, \gamma_h, \gamma_{L+1})$ really defines two ratios as a common factor can be absorbed by $\beta_n$. We tune $(\gamma_1, \gamma_h, \gamma_{L+1})$ at the base model dim using a three-dimensional hyperparameter search described in \sref{app:tune_constant_factors} and reuse the best set of values across all model sizes to give the eval losses in \cref{fig:adam_additive_interventions}(c). We include the values found for these estimated optimal constant factors for all optimizers and parameterizations in \sref{app:tune_constant_factors}.

\begingroup
\renewcommand{\arraystretch}{1}
\begin{table*}[ht]
\centering
\vspace{4pt}
\adjustbox{scale=0.8}{
\begin{tabular}{r wl{2cm} p{2.45cm} p{2.45cm} p{2.45cm} p{2.45cm}}
\toprule
& & \multirow{2}{=}{Global LR + default constants} & \multirow{3}{=}{Per-layer LR (full align) + default constants} & \multirow{3}{=}{Per-layer LR (full align) + optimal constants} & \multirow{3}{=}{Per-layer LR\\ (no align) + optimal constants}\\
\\
\\
\midrule
\multirow{4}{*}{SGD}    & STP & 3.657 & 3.510 & 3.385 & \textbf{3.318}\\ 
                         & NTK & 3.732 & 3.627 & \textcolor{red}{\textbf{3.313}} & 3.324 \\
                         & muP & 4.224 & 4.184 & \textbf{3.809} & 4.222 \\
                         & MFP & 4.092 & 4.319 & 4.092 & \textbf{3.898} \\
\midrule
\multirow{4}{*}{Adam}    & STP & 2.625 & 2.576 & \textcolor{red}{\textbf{2.572}} & 2.575\\ 
                         & NTK & 2.592 & 2.615 & 2.591 & \textbf{2.581} \\
                         & muP & 2.727 & 2.646 & 2.599 & \textbf{2.590} \\
                         & MFP & 2.638 & \textbf{2.630} & 2.633 & 2.638 \\
\midrule
\multirow{4}{*}{Adam+PS} & STP & 2.580 & 2.613 & 2.675 & \textbf{2.577}      \\ 
                         & NTK & 2.570 & 2.623 & 2.667 & \textcolor{red}{\textbf{2.566}}      \\
                         & muP & \textbf{2.574} & 2.655 & 2.606 & 2.575      \\
                         & MFP & 2.624 & 2.640 & 2.772 & \textbf{2.623}      \\
\bottomrule
\end{tabular}}
\vspace{4pt}
\caption{\textbf{Best eval losses for 26.8B parameter models.} For each optimizer $\times$ parameterization $\times$ setting, we sweep the base learning rate at each model size and use the eval loss from the best base learning rate. The best setting for each optimizer $\times$ parameterization is bold; the best parameterization + setting for each optimizer is bold and red. For Adam, the best model is standard parameterization + per-layer ``full alignment" + optimal constants, where the embedding LR scales like $O(1)$ and hidden and readout LRs scale like $O(1/n)$. For Adam+parameter scaling, the best model is NTK per-layer ``no alignment" + optimal constants, where all LRs scale like $O(1)$.}
\label{tab:eval_losses_26B}
\end{table*}
\endgroup

Using these optimal constants and the per-layer learning rate exponents that assume full alignment, we see that \emph{all parameterizations can perform hyperparameter transfer across width}. First, the eval loss improves across all scales when using the optimal constants instead of the default constants, indicating successful transfer of these constant multipliers. For the base model dim $b=1024$, tuning these constants can only improve the performance, but comparing \cref{fig:adam_additive_interventions}(c) to \ref{fig:adam_additive_interventions}(b) shows that for Adam these constants improve performance substantially for all model sizes across all parameterizations. If instead, these constants improved performance only at or near the model sizes where they were tuned, it would be less practical to include them for large model training because the three-dimensional sweep would be prohibitively expensive at large scale. The only exception is that these constants may not transfer well to the mean-field parameterization models above 2B parameters: we will show in \cref{sec:results_epsilon} that this is likely due to epsilon underflow and is addressed with our epsilon mitigations. These results indicate that our recipe for these per-layer constant multiplicative factors is both essential and practical: the performance gains are substantial and the hyperparameter transfer makes them feasible.

Second, the optimal base learning rate is consistent across scale in all parameterizations. In \cref{fig:adam_hyperparameter_transfer}, we first find the optimal learning rate at each model dim $n$ with a sweep of the base learning rate $\beta_n$, then fit a power law to the optimal base learning rate vs model dim. If the base learning rate were exactly the same at all scales, we would see a power law exponent of zero. We find exponents for standard, NTK, and muP of -0.05, -0.02, and -0.06, respectively, where these exponents close to zero indicate that the base learning rate is almost perfectly scale-invariant. For mean-field parameterization there is slight deviation from zero with an exponent of 0.26. This illustrates hyperparameter transfer for all parameterizations: when the constant factors are consistent across scale, the scaling prescription has correctly encapsulated the dependence on the scaling dimension into the prescribed exponents. We can therefore find the optimal base learning rate on a smaller model and reuse it on all model sizes.

Finally, we investigate the impact of the alignment assumptions, still using per-layer learning rates and optimal constants. In \cref{fig:adam_additive_interventions}(d), we use the learning rate exponents from the right side of \cref{tab:common_parameterizations} derived from assuming no alignment rather than full alignment. For NTK, muP and MFP, we see slight improvements in the eval losses across model sizes when using the no alignment exponents. On the surface, this improved performance would indicate that the no alignment exponents are preferable. However, when we look at the learning rate sweeps in \cref{fig:app_hparam_transfer_align_comparison}, the power law exponents fit to the optimal base learning rates for standard, NTK and muP are in the range of -0.58 to -0.69. Since these exponents are not close to zero, despite the modest performance improvements using the no alignment exponents over the full alignment exponents, the no alignment exponents do not appear to capture the learning rate scaling behavior as well as the full alignment exponents. For mean-field parameterization, it is less clear what the optimal learning rate exponents are as both the full alignment and no alignment settings have power law exponents slightly above zero. This indicates that in practice, there may be some nuance in selecting the alignment assumptions that are optimal for a particular use case and that there may be multiple interesting choices of exponents for a given parameterization within our more general space of parameterizations.

We note these empirical results across alignment assumptions could occur due to a mechanism other than the activation or logit growth with respect to width that is predicted by alignment in dense layers. For example, prior work notes training instabilities in Transformers due to attention logit growth~\citep{dehghani2023scaling,zhai2023stabilizing} or output logit divergence~\citep{chowdhery2023palm} and that these instabilities are sensitive to the learning rate~\citep{wortsman2023small}. We may also miss slight undercorrection of the alignment in our finite size models: if the true alignment is slightly larger than the learning rate exponents account for, we could have slow growth of activations or logits with respect to width that does not harm performance in our models but would eventually induce instability at sufficient width. However, our largest model with $26.8B$ parameters has a model dimension of $16{,}384$ which encompasses the width of many even larger models. As such, even if additional training instabilities may occur, we expect these per-layer learning rate and constant factor prescriptions to be relevant in practice to width scaling in large Transformers.

While muTransfer~\citep{yang2022tensorv} emphasizes that muP is the unique parameterization that allows hyperparameter transfer across width, our results show that all parameterizations can perform hyperparameter transfer across width when each parameterization uses theoretically motivated per-layer learning rate exponents. In addition, our eval loss results for the full alignment exponents contrast with the empirical results in \citet{yang2022tensorv} where muP outperforms standard parameterization in a $6.7B$ parameter GPT-3 model~\citep{brown2020gpt3}. We note, however, that their comparison across parameterizations has several elements that favor muP. First, the muP experiments use per-layer learning rate scaling while the standard parameterization experiments use a global learning rate. Second, the muP results tune a handful of constant factors at small scale and transfer them to large scale. While the specific constant factors differ from our setting, this is comparable to using our optimal constant learning rate factors instead of the default constant factors. As such, their comparison of muP and standard parameterization is analogous to comparing muP in our third experiment (\cref{fig:adam_additive_interventions}(c), green curve) against standard parameterization in our first experiment (\cref{fig:adam_additive_interventions}(a), blue curve), which indeed shows a benefit for muP. Instead, we argue that the fair comparison across parameterizations would use per-layer learning rates and optimal constants for both, that is, to consider both parameterizations in our third experiment \cref{fig:adam_additive_interventions}(c). There, we see that standard parameterization outperforms muP and in fact substantially so: the \emph{second largest} standard parameterization model with $15.3B$ parameters outperforms the \emph{largest} muP model with $26.8B$ parameters despite having 57\% as many parameters. We therefore recommend our full alignment per-layer learning rate prescription for standard parameterization for Adam where the embedding layer learning rate scales like $O(1)$ and the hidden and readout layer learning rates scale like $O(1/n)$.

\begin{figure*}[ht]
    \centering
    \includegraphics[width=\textwidth]{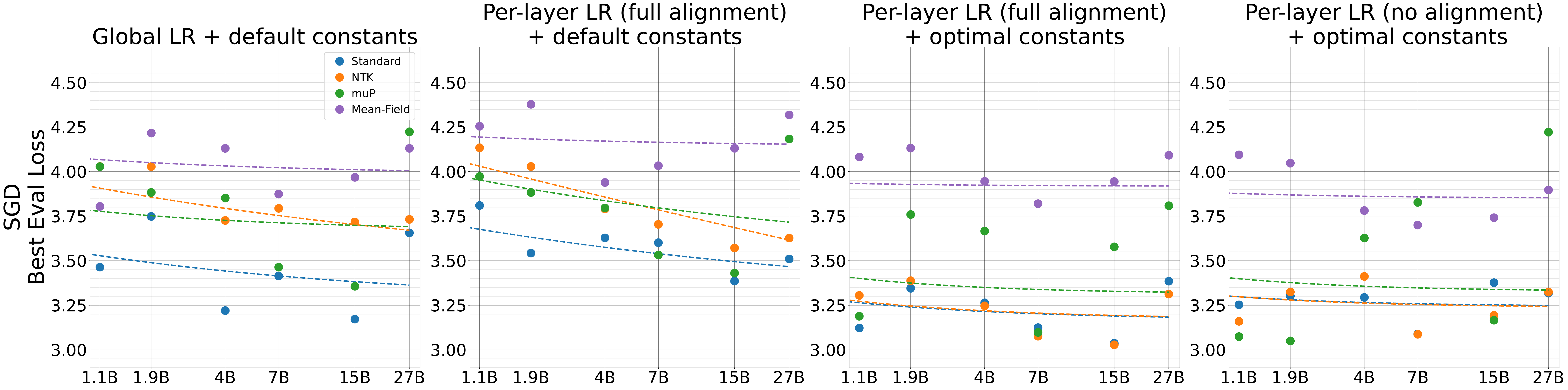}
    \vspace{2pt}
    \includegraphics[width=\textwidth]{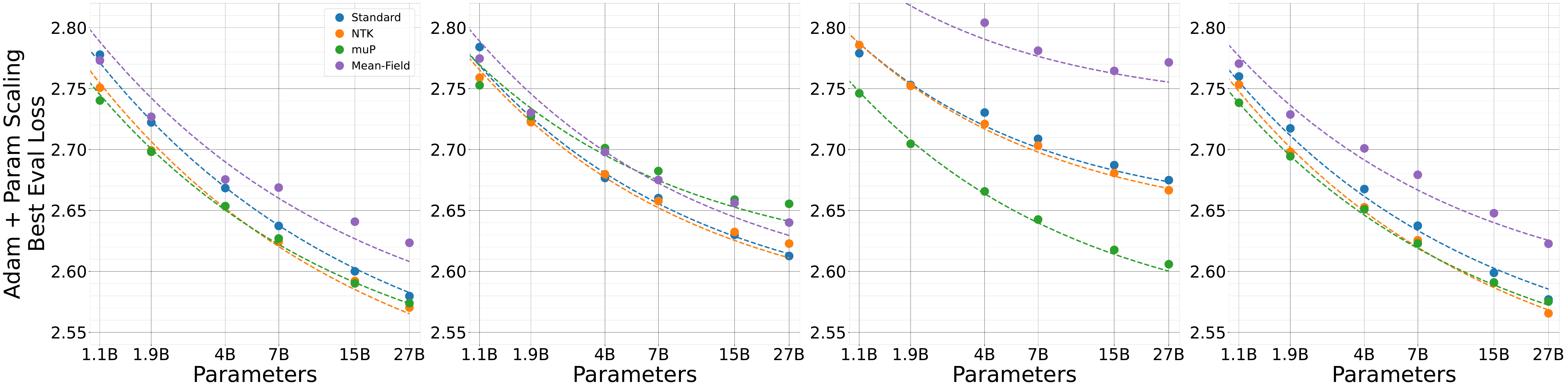}
    \caption{For SGD + momentum (top row) and Adam + parameter scaling (bottom row), eval loss comparisons for all parameterizations across a sequence of interventions. From left to right columns: (a) global LR exponents + default constants, (b) per-layer LR exponents assuming full alignment + default constants, (c) per-layer LR exponents assuming full alignment + optimal constants, (d) per-layer LR exponents assuming no alignment + optimal constants.}
    \label{fig:sgd_adam_ps_additive_interventions}
\end{figure*}

\subsubsection{SGD and Adam + Parameter Scaling}
We next present the results for the same series of experiments for SGD and Adam + parameter scaling in \cref{fig:sgd_adam_ps_additive_interventions}, where we use (a) global learning rates + default constants, (b) per-layer learning rates (full alignment) + default constants, (c) per-layer learning rates (full alignment) + optimal constants, and (d) per-layer learning rates (no alignment) + optimal constants. As we might expect, the results for SGD have significantly worse performance than the other optimizers, and the eval loss values are quite noisy. The impact of the different learning rate exponents is difficult to distinguish from noise for SGD, but there are visible performance improvements when using the optimal constants over the default constants.

For the per-layer learning rate experiments in the Adafactor family of optimizers, we use Adam + parameter scaling as the optimizer. The Adafactor optimizer was proposed in \citet{shazeer2018adafactor} and includes three types of changes to Adam. First, it reduces the memory requirements for the optimizer state by using a low-rank approximation of the gradient second moment and eliminating the exponential moving average on the first moment. Second, it introduces a parameter scaling term in the update rule. Third, it perform update clipping rather than gradient clipping, which stabilizes the update when the second moment estimate is out-of-date due to the moving average.

From the perspective of width-scaling, the most important of these differences is the parameter scaling, which multiplies the normalized gradients by the norm of the existing parameters. To focus our investigation on the impact of this parameter scaling term, we use Adam + parameter scaling as the optimizer in this section, so that the parameter scaling term is the sole change from our implementation of Adam. Additionally, the low-rank approximation in Adafactor introduces changes in the tensor shapes, which caused issues out-of-the-box with our implementation of fully sharded data parallelism (FSDP)~\citep{rajbhandari2020zero}. The choice of Adam + parameter scaling avoids these issues. As a cross-check, in \cref{app:adafactor_adam_ps} we compare Adafactor and Adam + parameter scaling and show that the performance differences are small and that overall these two optimizers occupy the same width-scaling regime.

Overall, we see very good empirical performance from Adam + parameter scaling. When comparing the best setting for each parameterization across Adam with or without parameter scaling, the addition of parameter scaling slightly improves the eval loss for all parameterizations except standard. In particular, the best performing model across all parameterizations and optimizers is quite unexpected: it is Adam + param scaling + NTK + no alignment! This result further validates that it is critical to consider our more general space of parameterizations rather than making specific alignment assumptions upfront.

For the experiments in \cref{fig:sgd_adam_ps_additive_interventions} across different settings with global, full alignment per-layer and no alignment per-layer learning rate exponents, we first note that for Adam + parameter scaling, the no alignment setting is equivalent to the global learning rate setting. Concretely, when we compute the maximum stable learning rate in the no alignment setting in the rightmost column of \cref{tab:common_parameterizations}, for all parameterizations we get $O(1)$ scaling in all layers. This means the per-layer exponents in the no alignment setting do not modify the base learning rate, which is equivalent to the global learning rate setting that does not apply these per-layer exponents. Conceptually, this illustrates an interesting theoretical property of parameter scaling: in a sense, it accomplishes with the optimizer alone what the width-scaling theory of parameterization intends when carefully selecting the learning rates to preserve the scale of the activations and logits. Recall that the stability constraints ensure that, at initialization, the parameters contribute to constant activations and at-most-constant logits. If we neglect for a moment the contribution of alignment, parameter scaling makes the updates to each layer exactly the ``right" scale by matching the existing parameter norms, allowing every layer to use a constant scale learning rate regardless of the parameterization. If we do account for alignment, rather than using constant learning rates, we reduce the hidden and/or readout layer learning rates to $1/\sqrt{n}$ to account for the extra $\sqrt{n}$ contribution from alignment. The parameter scaling therefore makes all parameterizations behave similarly, which we see in Figures \ref{fig:lr_sweep_adam_ps_full_align} and \ref{fig:lr_sweep_adam_ps_no_align} where all parameterizations use similar values for the base learning rate and have similar scale-dependence of the optimal learning rate under each set of learning rate exponents.

With this equivalence of the no alignment and global learning rate settings in mind, we can now compare the no alignment and full alignment settings for Adam + parameter scaling. For three reasons, the no alignment exponents appear preferable to the full alignment exponents. First, the eval losses from the no alignment (or global) exponents in \cref{fig:sgd_adam_ps_additive_interventions}(a) and (d) outperform the full alignment exponents in \cref{fig:sgd_adam_ps_additive_interventions}(b) and (c). Moreover, the performance gap increases with scale suggesting the no alignment exponents are truly superior. Second, the full alignment experiments have very high learning rate sensitivity: this is seen in the higher curvature in the eval loss vs learning rate curves in \cref{sec:app_lr_sweeps_adam_ps} where a small deviation from the optimal learning rate causes a larger loss in performance. In addition, the optimal learning rate is quite close to the maximum stable learning rate, which is a risky scenario for training stability. In general, high learning rate sensitivity is an undesirable property: when all else is equal, it is preferable to have a model that can train well with a larger range of learning rates than one that requires a very narrow range of learning rates for success.~\citep{wortsman2023small} Third, the optimal constant multipliers transfer across scales well for the no alignment exponents but actually harm performance compared to the default constants for full alignment. Recall that the optimal constants should transfer well across scale when the learning rate scaling is well-encapsulated in the per-layer learning rate exponents. The positive transfer of these constants under the no alignment exponents compared to the negative transfer under the full alignment exponents indicates that the no alignment exponents better encapsulate the learning rate scaling into the exponents.

However, despite these reasons to prefer the no alignment exponents for Adam + parameter scaling, we see an interesting phenomenon when we look at the power law exponents on the optimal learning rate in the full alignment and no alignment settings. Across all combinations of the parameterization, default or optimal constants, and full or no alignment, in Figures \ref{fig:lr_sweep_adam_ps_full_align} and \ref{fig:lr_sweep_adam_ps_no_align} we see similar power law exponents in the range of $-0.29$ to $0.05$. In particular, both the full alignment and no alignment settings are close to scale-invariant, but neither is clearly more scale-invariant than the other. This contrasts with the Adam setting, where the full alignment setting had optimal learning rate power law exponents close to zero, and the no alignment setting had power law exponents close to $-0.5$. In that setting, the exponent difference of approximately $0.5$ corresponded with the $\sqrt{n}$ difference in the prescribed learning rate exponents between the two alignment settings. This showed that the full alignment setting was encapsulating the scaling behavior into the prescribed exponents whereas the no alignment setting required the base learning rate to change across scale to absorb what wasn't captured by the prescribed learning rate exponents. In contrast, for Adam + parameter scaling, the power law exponents that are slightly less than zero in both the full alignment and no alignment settings indicate that parameter scaling may introduce factors that influence the optimal learning rate scaling that are more complex than the width-scaling dependence we analyze here.

Although Adafactor has been much less widely adopted than Adam, it has been used successfully in large model training~\citep{t5,chowdhery2023palm,du2022glam,fedus2022switch,zoph2022st}. However, several papers have reported difficulties with training stability and brittleness to the learning rate or other hyperparameters~\citep{rae2021scaling,zhai2021scaling}. Both the empirical success and this brittleness are consistent with our results. In our experiments with parameter scaling, we observe eval losses that are similar or better than for Adam, and in particular the best eval loss across all optimizers and parameterizations uses Adam + parameter scaling and NTK. In addition, we note two features that may contribute to the empirical success of parameter scaling. First, the global learning rate setting coincides with the theoretically motivated no alignment exponents. Second, the default constants equal to one are not far from the optimal constants (see \sref{tab:optimal_per_layer_multipliers}), which range in value from 0.5 to 2.6. This contrasts with the larger range in the optimal constants for Adam that span 0.15 to 11.7. These two properties may help adaptive optimizers with parameter scaling perform well in the typical setting in practice that uses global learning rates and no per-layer learning rate constants. At the same time, we observe high learning rate sensitivity that may contribute to training stability difficulties or brittleness to optimizer hyperparameter choices. In particular, in our largest models, the optimal learning rate is close to the maximum stable learning rate which may cause issues with training stability.

In conclusion, for adaptive optimizers with parameter scaling, we see similar width-scaling behavior from all parameterizations. We recommend using the no alignment, or equivalently, global learning rate exponents, for improved performance, and close to scale-invariance in the base learning rate. Unlike Adam, tuning the per-layer constant multipliers has minor performance impact and should not be considered essential. In addition, future work is needed to more clearly understand the impact of parameter scaling on the training stability and hyperparameter sensitivity.

\begin{figure*}[ht]
    \centering
    \includegraphics[height=4.1cm]{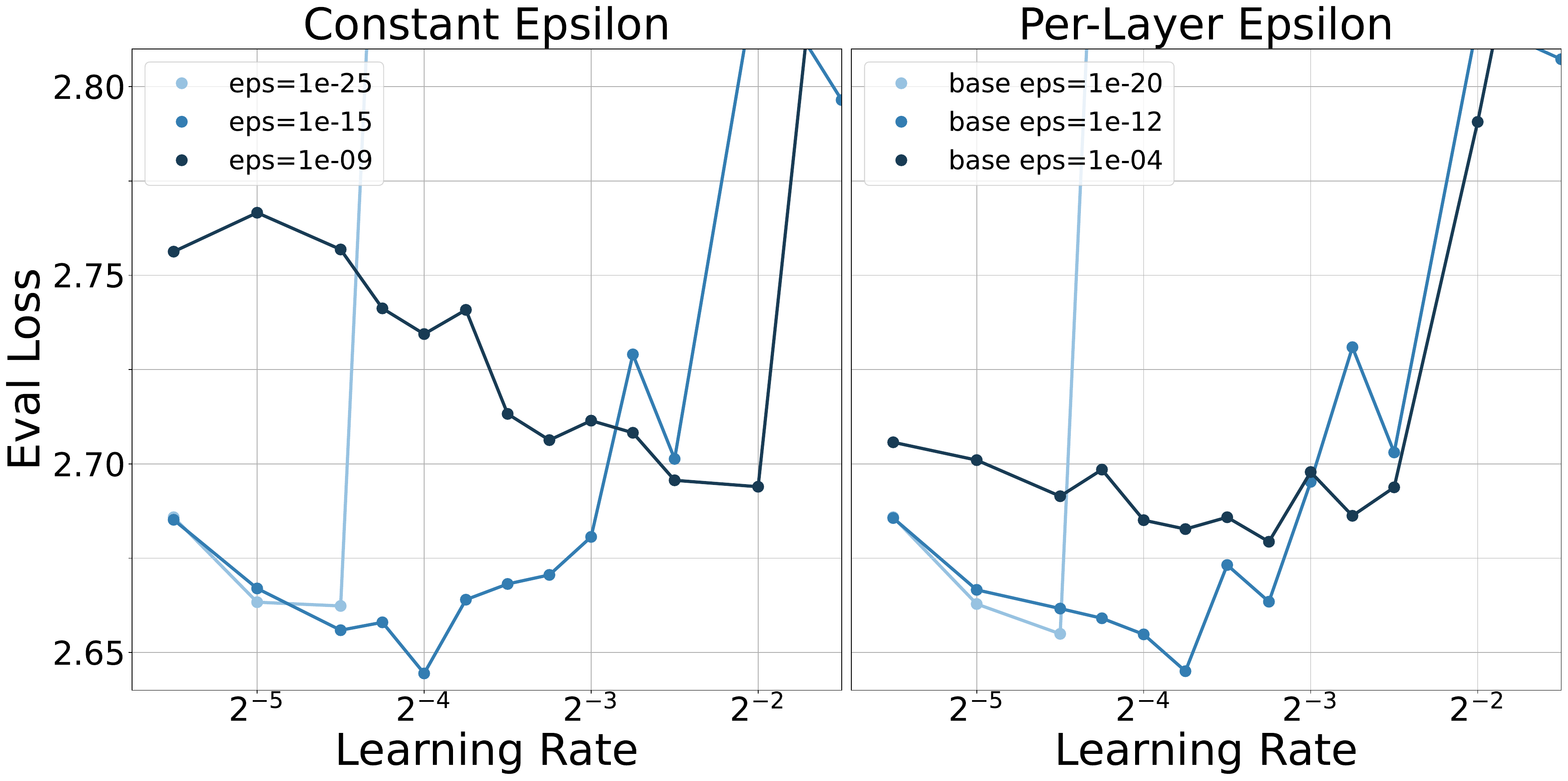}
    \includegraphics[height=4.1cm]{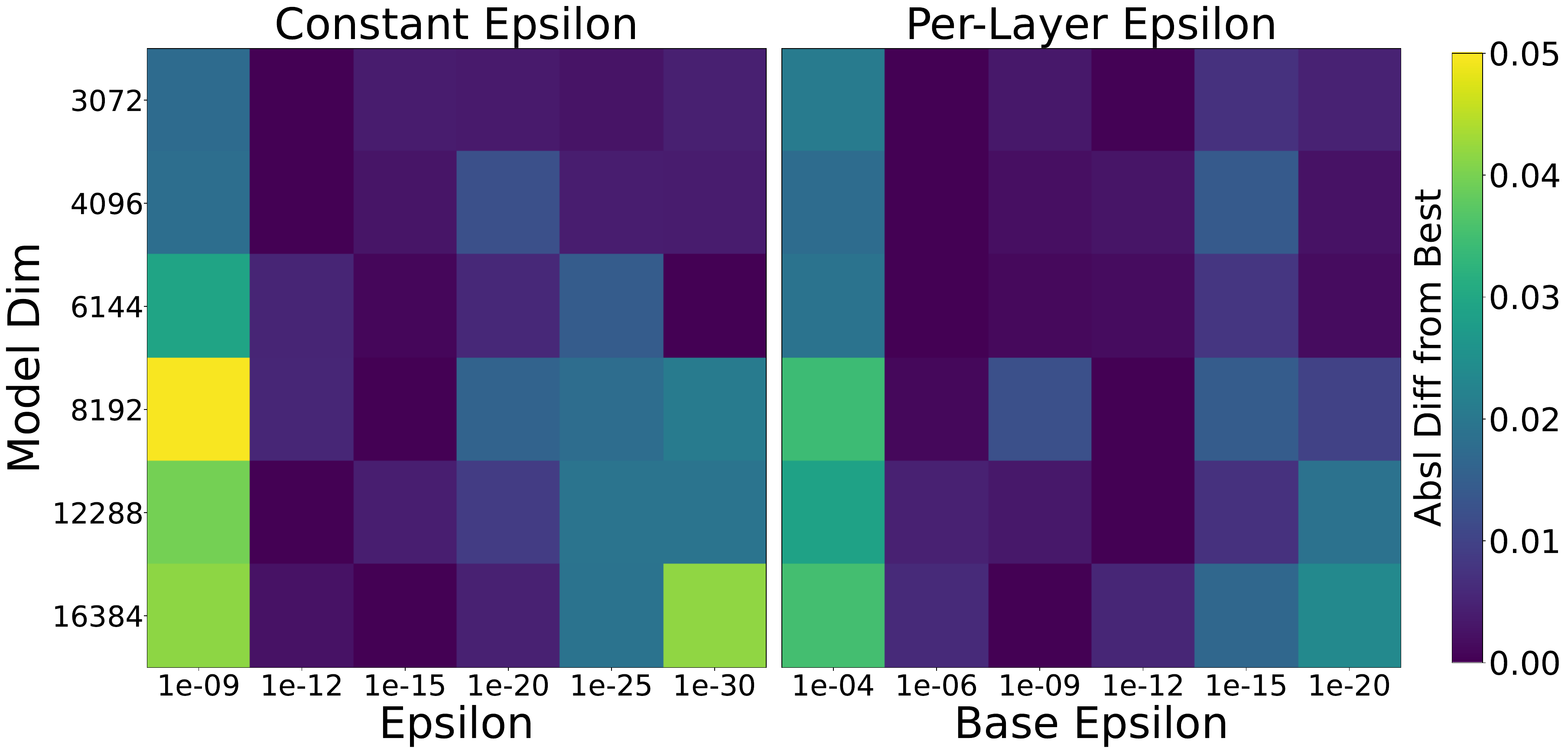}
    \caption{\textbf{Epsilon in Adam requires tuning for both constant and per-layer epsilons.} (a-b) Learning rate sweep of epsilon constant factors at model dim $D=4096$ for mean field parameterization. Epsilon too small (light blue) = instability at low learning rate, epsilon too large (dark blue) = suboptimal performance, epsilon just right (medium blue) = optimal performance. (c-d) Heatmaps for epsilon constant factors for mean field parameterization in six model sizes: color indicates the absolute difference in eval loss from the best constant value for that setting. Dark blue columns in the middle indicate good performance whereas smaller or larger values have suboptimal performance.}
    \label{fig:epsilon_constants}
\end{figure*}

\subsection{Epsilon Underflow in Adaptive Optimizers}
\label{sec:results_epsilon}

In adaptive optimizers like Adam, the denominator of the update rule adds a small epsilon parameter to the gradient second-order moment, originally intended to regularize against division by zero when the gradients are very small~\citep{duchi2011adaptive,hinton2012neural}. More recently, \citet{choi2019empirical} shows that epsilon is a hyperparameter that requires tuning. Despite its small value, typically around 1e-8~\citep{paszke2019pytorch,deepmind2020jax}, the epsilon parameter prevents Adam from being perfectly scale-invariant in that multiplying the gradients by a constant would not alter the resulting update. In particular, if the gradient scale drops below the size of epsilon then epsilon dominates the gradients instead of acting as a negligible additive constant. Since gradients decrease in scale with model width as in \cref{tab:common_parameterizations}, in theory, for any constant value of epsilon there exists a sufficiently large model that will encounter this scenario.

Two recent works note this phenomenon and propose possible mitigations. From an empirical perspective, ~\citet{wortsman2023small} observes that gradient norms in standard parameterization models decrease with model size and approach 1e-8 for 1.2B parameter models, suggesting this epsilon underflow is relevant in practice. As a mitigation, they propose using a smaller constant value for epsilon and show that decreasing epsilon from 1e-8 to 1e-15 improves the loss in a 4.6B parameter model. From a theoretical perspective, ~\citet{yang2023tensorivb} notes that epsilon should be treated as part of the parameterization and propose per-layer epsilon scaling. For each layer, epsilon is proportional to the parameterization and layer-specific gradient scale shown in \cref{tab:common_parameterizations}. Similar to our per-layer learning rate experiments, we implement this as
\begingroup
\setlength{\abovedisplayskip}{4pt}
\setlength{\belowdisplayskip}{6pt}
\begin{align*}
    \epsilon_l = \textrm{base epsilon} \cdot (\frac{n}{b})^{-g_l}
\end{align*}
\endgroup
for each layer $l$, where $g_l$ is the negative exponent of the gradient scale, the base model dim $b$ is $1024$, and the base epsilon is determined empirically. This approach has not been empirically validated and adds significant implementation complexity, but should ensure that epsilon and the gradients scale in tandem across width.

We investigate the practical impact of epsilon across parameterizations. As seen in the gradients column of \cref{tab:common_parameterizations}, gradients decrease as model width increases with an exponent specific to both the parameterization and layer. As such, we expect that different parameterizations will encounter epsilon underflow at different model sizes: in particular, the steepest exponent is the mean-field parameterization hidden layer, which scales like $1/n^{1.5}$. This suggests that mean-field parameterization should encounter epsilon underflow at smaller model sizes than other parameterizations.

Both constant epsilon and per-layer epsilon require hyperparameter tuning, to select the constant or the base constant multiplier, respectively. As shown in \cref{fig:epsilon_constants} for mean field parameterization and \cref{fig:epsilon_appendix_heatmaps} for all parameterizations, constant values that are too large lead to suboptimal performance; values that are too small lead to instability, presumably by failing to prevent the numerical instability epsilon was originally intended to prevent. Rather than providing numerical stability for small number division with an additive constant in the denominator that breaks scale-invariance, we propose \emph{Adam-atan2}: a variant of Adam that replaces the standard division operation in the update rule with the standard library function $\atan2$. The function $\atan2(x,y)$ returns $\arctan(x/y)$ in the appropriate quadrant, which is approximately equal to $x/y$ due to small-angle approximation when $x/y$ is close to zero and asymptotically approaches $\pm\,\pi/2$ as the argument goes to $\pm\,\infty$. In particular, the $\atan2$ function is defined even at $(0,0)$ exactly and is scale-invariant up to precision limits. The single-line code change in \cref{sec:adam_atan_code} to use arctangent in the Adam update equation eliminates the epsilon hyperparameter entirely and restores the scale invariance to Adam.

\begin{figure*}[ht]
    \centering
    \includegraphics[height=4.1cm]{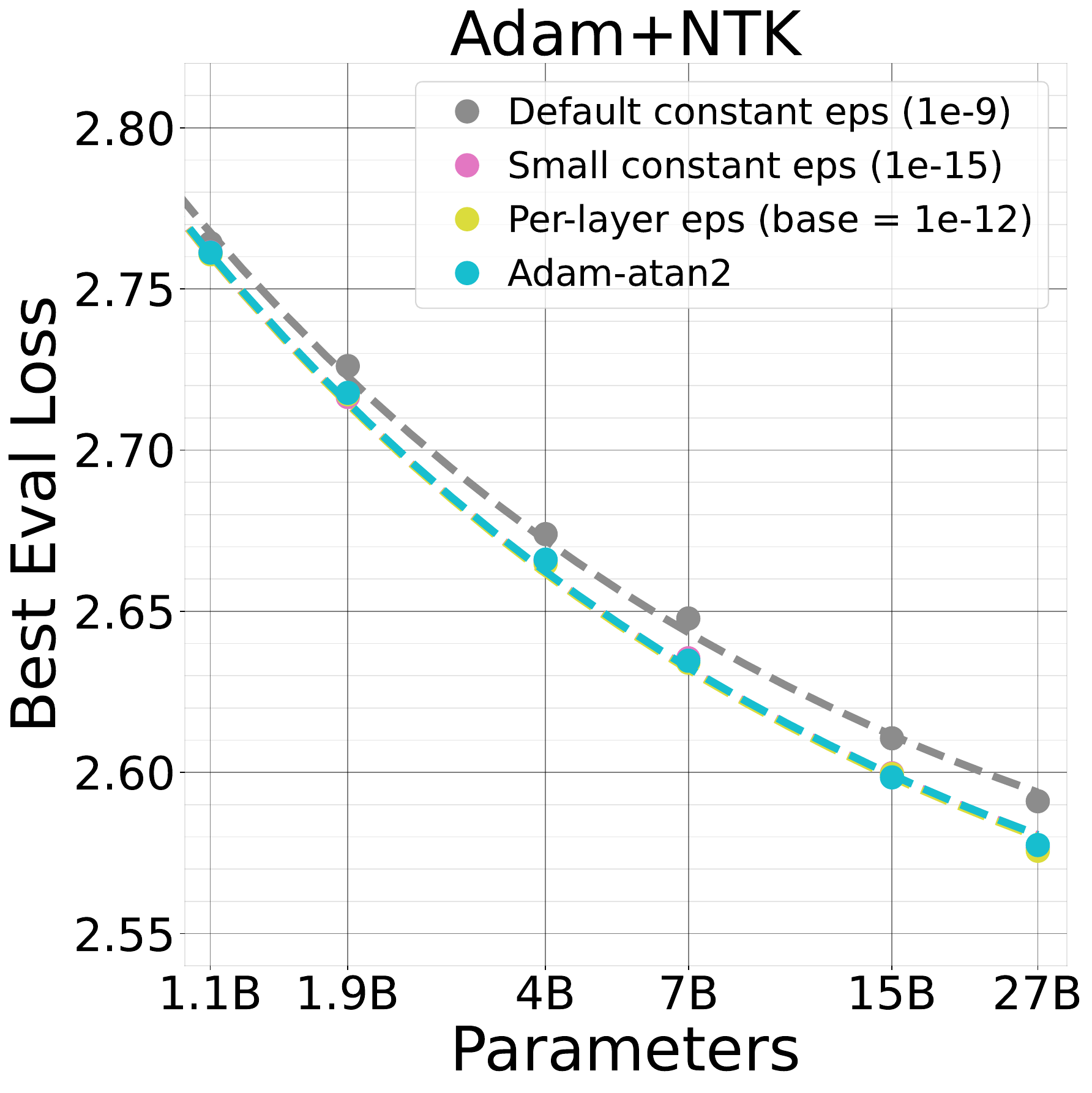}
    \includegraphics[height=4.1cm]{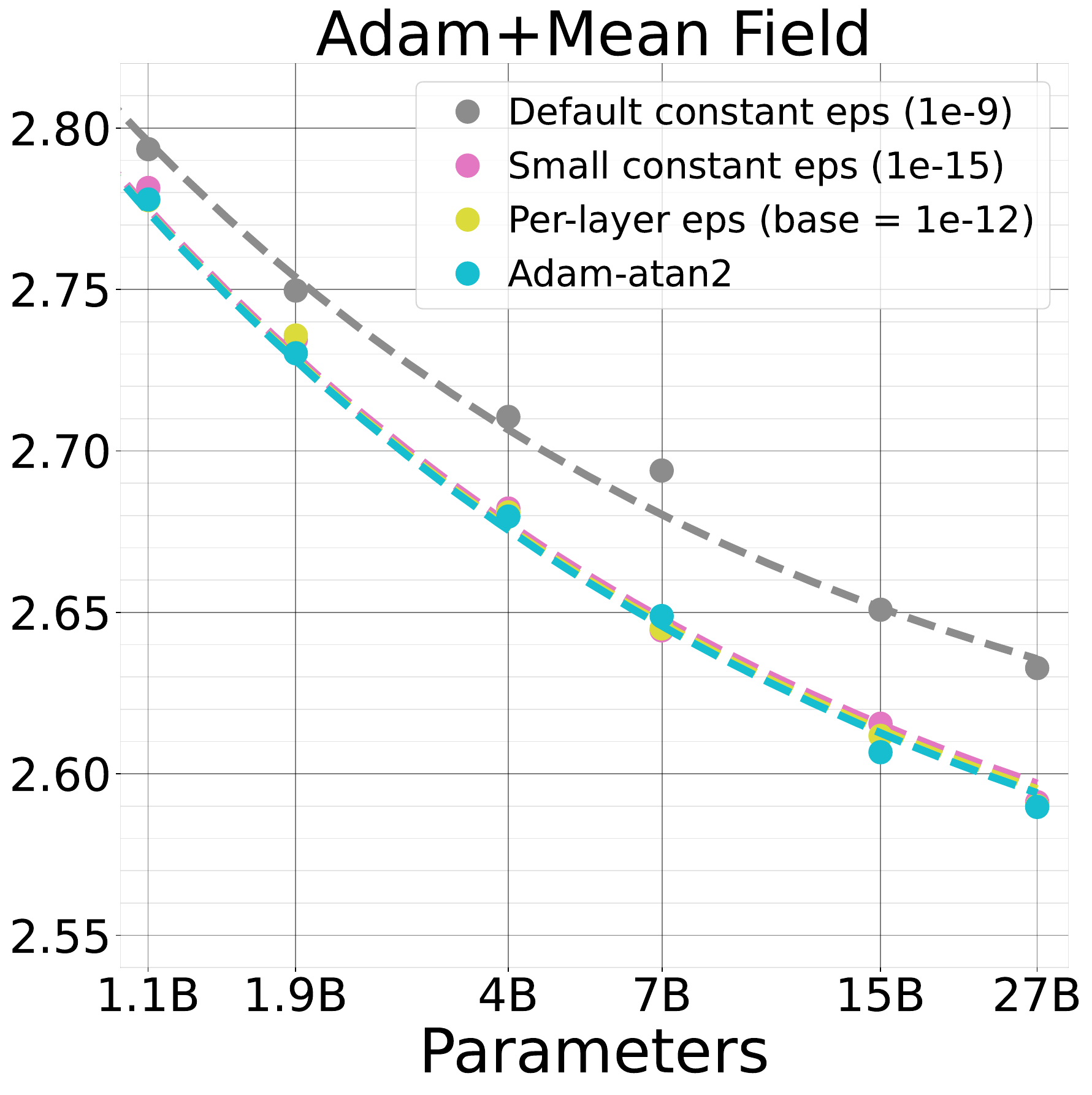}
    \includegraphics[height=4.1cm]{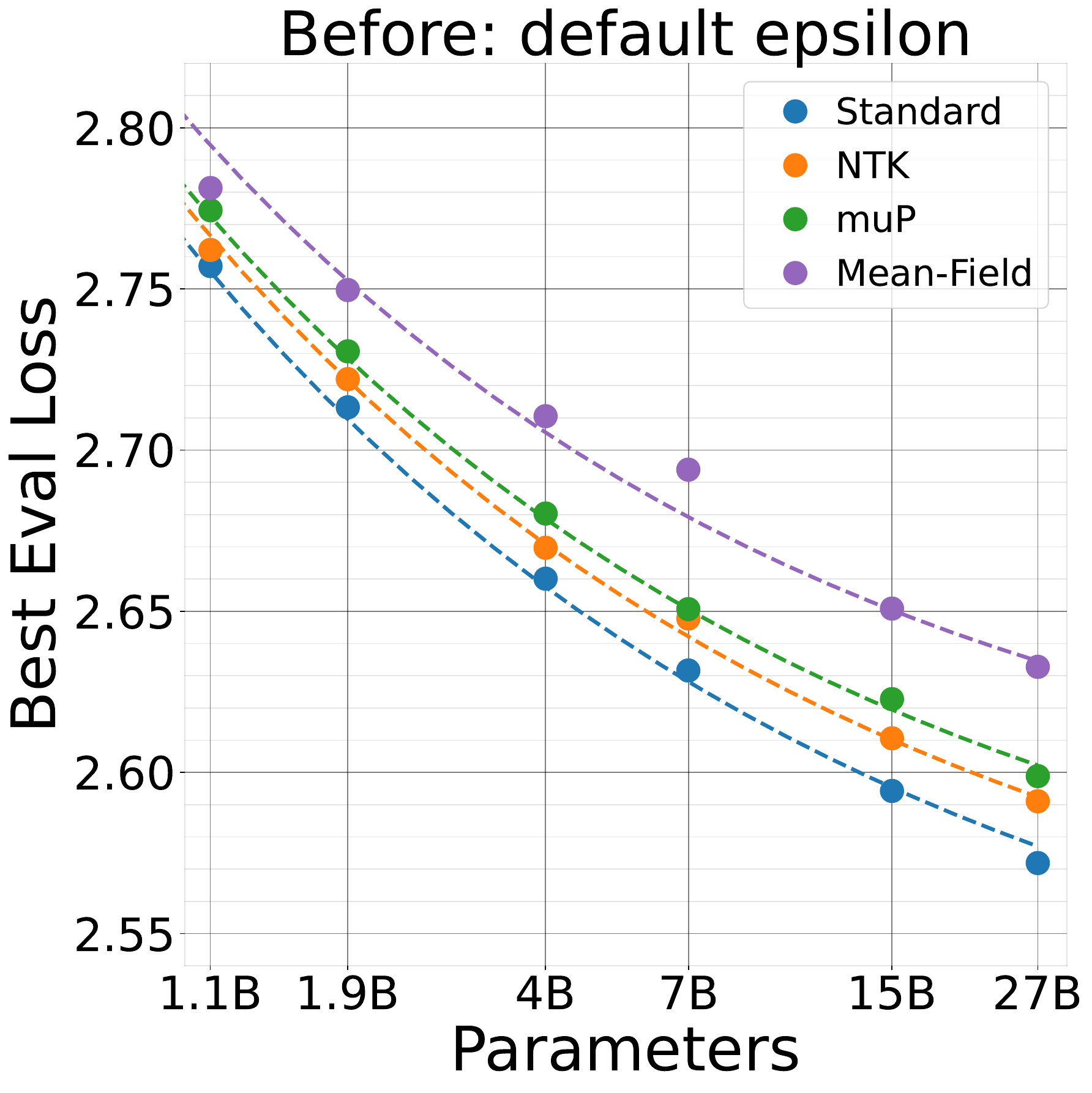}
    \includegraphics[height=4.1cm]{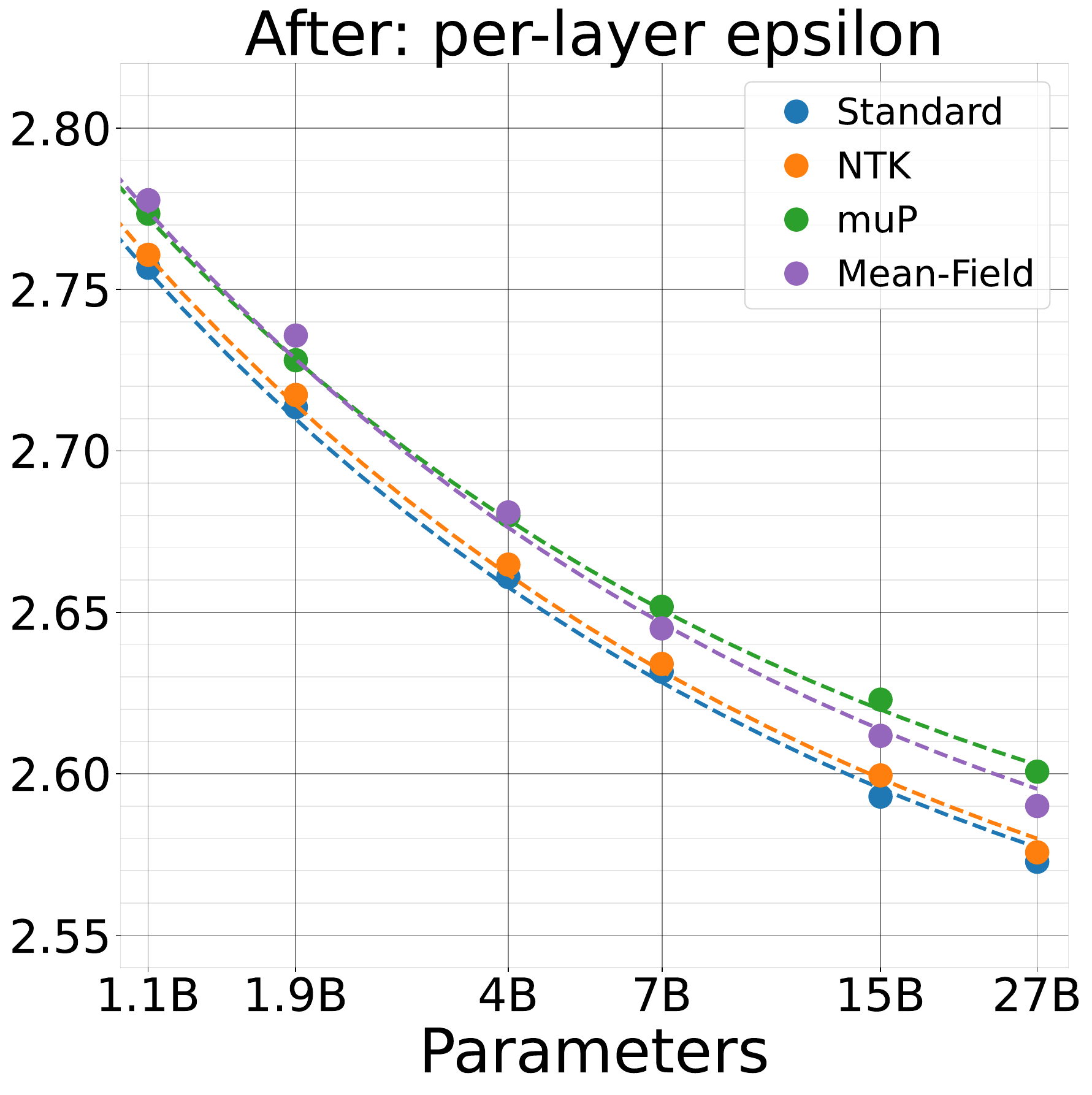}
    \caption{\textbf{All epsilon mitigations improve performance similarly for NTK and MFP.} Epsilon mitigations for Adam using the best choice of constants in each mitigation setting, showing all three mitigations equally and substantially improve performance in (a) NTK and (b) Mean Field. (c) Before fixing epsilon, equivalent parameterizations have different performance. (d) After fixing epsilon, shown here with per-layer epsilon, equivalent parameterizations (STP+NTK, muP+MFP) have approximately equivalent performance.}
    \label{fig:epsilon_mitigations}
\end{figure*}

For all parameterizations, we compare the three mitigations using the best choice of constant in each setting against a baseline epsilon (1e-9): small constant epsilon (1e-15), per-layer epsilon scaling (base epsilon = 1e-12), and our proposal \emph{Adam-atan2}. In \cref{fig:epsilon_mitigations}(a) and (b), we see that all three mitigations have modest improvements in the eval loss for NTK and and substantial improvements for mean-field parameterization. In addition, the gap in performance increases with model size. The other two parameterizations show no performance changes in our model sizes (see \cref{fig:epsilon_appendix_scaling_plots}) with any of the epsilon mitigations. The particular sensitivity of mean-field parameterization to epsilon is consistent with our theoretical motivation: due to its hidden layer gradients scaling like $1/n^{1.5}$, we expect that among the parameterizations, mean-field will encounter epsilon underflow at the smallest model sizes and benefit most from these mitigations.

We therefore recommend care when setting epsilon. For standard parameterization models with up to a billion parameters, the typical default value of 1e-8 is likely acceptable but slightly smaller values of 1e-12 or 1e-15 may be preferable. For larger models or other parameterizations, using epsilon requires smaller constants or per-layer epsilon scaling, with tradeoffs between implementation complexity and, at least theoretically, hyperparameter tuning costs. In principle, the theoretical prescription for per-layer epsilon encapsulates the epsilon scaling in the exponents, allowing hyperparameter transfer of the constant multiplier similar to other parameterized quantities like learning rates. In contrast, the optimal constant epsilon should be scale-dependent in theory, but we note that with our model sizes and constant search resolution we did not see scale dependence of the optimal epsilon constant. However, this epsilon hyperparameter can be eliminated entirely with \emph{Adam-atan2} with a one-line code change and the same improved performance.

Moreover, epsilon illustrates that finite precision plays an important role in parameterization in practice. Recall that standard and NTK parameterizations, and similarly muP and mean-field parameterizations, are theoretically equivalent under the equivalence relations in \cref{app:theory_equivalence_classes}, if we overlook the contribution of epsilon. However, using default epsilons in \cref{fig:epsilon_mitigations}(c), we see significant performance gaps between equivalent parameterizations that are closed when epsilon underflow is mitigated in \cref{fig:epsilon_mitigations}(d): now the pairs standard + NTK and muP + mean-field show approximately equal performance. It is plausible that the more widespread usage of standard and muP parameterizations over their equivalents has been influenced by this phenomenon.

Lastly, we can now compare the performance across equivalence classes rather than individual parameterizations: we see that the standard parameterization equivalence class outperforms the muP equivalence class. This narrows the possible explanations for the performance differences to the elements that we saw distinguish the equivalence classes in \cref{sec:equivalence_relations}: namely, the shift in the readout layer in muP and mean-field parameterization that initializes the logits to scale like $1/\sqrt{n}$ appears to harm the empirical performance. Instead, to obtain the best performance for Adam, we recommend the parameterizations that initialize the logits to be constant scale, in particular our prescription for standard parameterization with per-layer learning rates where the embedding layer learning rate scales like $O(1)$ and hidden and readout layer learning rates scale like $O(1/n)$.

\section{Related Work}

In addition to the prior work in~\citet{yang2021tensoriv,yang2023tensorivb} discussed earlier in the paper, the literature on width-scaling parameterizations~\citep{lee2017deep,matthews2018gaussian,jacot2018neural} includes the neural tangent kernel parameterization~\citep{jacot2018neural}, a modification to standard parameterization to enable a consistent infinite-width limit~\citep{sohl2020infinite}, and a mean-field limit~\citep{mei2018mean,geiger2020disentangling,rotskoff2018parameters,chizat2018global,araujo2019mean} for single-hidden-layer MLPs. \citet{bordelon2022self} proposed the mean-field parameterization by extending this mean-field limit to deep neural networks using self-consistent dynamical field theory. \citet{yaida2022meta} proposed a one-parameter family of hyperparameter scaling strategies that interpolates between the neural tangent scaling and mean-field or muP scaling. This meta-parameterized scaling strategy has been used to propose width-scaling initialization and training hyperparameters in Transformers~\cite{dinan2023effective}. Various ``unit scaling'' strategies proposed by \citet{noam_unit_scaling,kaplan2019notes} are similar to NTK parameterization. In addition, \citet{blake2023unit} also modifies the per-layer gradients to keep the activations and parameters close to unit scale regardless of model size. Throughout this paper, as in \citet{yang2021tensoriv} and \citet{yang2023tensorivb}, we use the RMS norm to quantify the scale of activations and logits, but other choices of norm are possible including the spectral norm used in \citet{yang2023spectral} to provide a spectral perspective on muP, and a modular norm proposed in \citet{large2024scalable}. \citet{ishikawa2023parameterization} extends muP to second-order optimizers (K-FAC, Shampoo).

Empirical evaluations of muP using Adam in Transformers were first presented in muTransfer~\citep{yang2022tensorv} on models up to $6.7B$ parameters. Cerebras-GPT~\citep{dey2023cerebras} found that muP with per-layer learning rates aided hyperparameter transfer and outperformed standard parameterization with global learning rates in compute-optimal models up to 2.7B parameters. In addition,~\citet{lingle2024large} showed that the optimal learning rate transfer across width for muP held on ablations of many architectural and optimal choices in models up to 1.2B parameters with learning rate granularity of $4\times$, but that the optimal learning rate drifted by a factor of $2\times$ between $40M$ parameter and $10B$ parameter models.

Other important scaling dimensions include depth and batch size. For depth scaling, \citet{yang2023tensorvi}, \citet{bordelon2023depthwise}, and \citet{chizat2023steering} make similar proposals that add a $1/\sqrt{L}$ scaling factor to the residual branch of a ResNet or Transformer where $L$ is the depth. Batch size scaling is investigated in~\citep{shallue2019measuring,mccandlish2018empirical,zhang2019algorithmic,kaplan2020scaling}. It is common to increase the batch size with model size as in~\citep{brown2020gpt3,rae2021gopher,hoffmann2022training,chowdhery2023palm,workshop2022bloom,dey2023cerebras,bi2024deepseek} but there are exceptions that use fixed batch sizes~\citep{thoppilan2022lamda,touvron2023llama,touvron2023llama2} across scale.

\section{Limitations}
This paper focuses specifically on width scaling, while in practice there are a number of dimensions that need to scale in tandem to achieve optimal performance in large models. In particular, large models typically co-scale at least the width, depth, batch size, training horizon, weight decay, and learning rate. For specific pairs or subsets of these scaling dimensions, such as batch size and learning rate or the aspect ratio between width and depth, the literature contains some theoretical insights and empirical validation, but determining how to optimally co-scale this entire set of dimensions is a complex and open problem. As we discuss in \cref{app:fixed_vs_compute_opt}, co-scaling the width and training horizon as in the compute-optimal regime requires adjustments to the learning rate exponents, but developing theory with realistic implications would require alternate approaches.

In addition, our theory considers the input and output dimensionality, corresponding to the vocabulary size in a Transformer language model, to be an $O(1)$ constant with respect to width. However, our experiments use a vocabulary size of $32{,}000$ that is typical for our range of model sizes. We do hold the vocabulary size fixed across our model sizes, but the magnitude of $32{,}000$ is comparable to model widths that range up to $D=16{,}384$, and it might be more realistic to consider the vocabulary size to be $O(n)$ rather than $O(1)$.

\section{Conclusions and Future Work}
From our broad perspective across parameterizations and optimizers, we find that key assumptions about alignment in prior work require additional consideration. Using empirical measurements from our alignment ratio metric, we find that alignment is a dynamical quantity that depends significantly on the training step, parameterization and layer, and less heavily on the optimizer. The alignment measured during training gives intermediate values that indicate that alignment assumptions in prior work may be overly conservative, suggesting that a larger set of parameterizations is more interesting than previously thought.

By considering a more general space of parameterizations with respect to the alignment, we show that all parameterizations benefit from theoretically motivated learning rate exponent prescriptions. We also demonstrate that several hyperparameters should be chosen carefully. First, we show the necessity and practicality of tuning the constant factors in per-layer learning rate prescriptions. These constants transfer well across model sizes, showing that all parameterizations can perform hyperparameter transfer under the right theoretical prescription. Second, the epsilon hyperparameter in adaptive optimizers induces gradient underflow using typical defaults at realistic model sizes, in particular for mean-field parameterization. Theoretically, epsilon should be considered part of the parameterization and scaled per-layer, but practically, small constant values can perform just as well when selected carefully. To eliminate epsilon entirely, we propose making Adam scale-invariant with \emph{Adam-atan2}.

Future work might consider alignment-aware learning rate schedules or alignment-aware optimizers. In addition, since the characterization of parameterizations into feature learning and kernel limits is specific to the alignment assumptions, this characterization could be extended to the general alignment setting. Beyond width scaling, future work should investigate the other scaling dimensions that are necessary for large model training, in particular depth and batch size.

\clearpage

\section*{Acknowledgements}
The authors would like to thank Yasaman Bahri and Kevin Yin for detailed feedback on paper drafts, and Kevin Yin for the suggestion to scale the arctangent function as $\lambda \cdot \textrm{atan2}(x, \lambda y)$. We also thank Ben Adlam, Kelvin Xu, Justin Gilmer, Gamaleldin Elsayed, Mark Kurzeja, Jiri Hron, Noah Fiedel, Zachary Nado, Justin Austin, Ben Poole, Clare Lyle and Tomás Lozano-Pérez for technical discussions and are grateful to four anonymous ICML reviewers who provided especially high-quality reviews and discussion during the rebuttal period.

\section*{Impact Statement}
This paper presents work whose goal is to advance the field of Machine Learning. There are many potential societal consequences of our work, in particular due to the large model sizes considered in this work, but we do not feel there are specific aspects of this work with broader impacts beyond the considerations relevant to all large machine learning models.

\bibliography{references}
\bibliographystyle{icml2024}

\newpage
\setcounter{section}{0}

\appendix
\onecolumn
\appendix
\pdfbookmark{Appendix}{unnumbered}
\section*{Appendix}\label{sec:appendix}
\setcounter{section}{0}
\renewcommand{\thesubsection}{\Alph{subsection}}

\counterwithin{figure}{subsection}
\counterwithin{table}{subsection}
\renewcommand\thefigure{\thesubsection\arabic{figure}}
\renewcommand\thetable{\thesubsection\arabic{table}}

\subsection{Author Contributions}
Katie, Lechao, Jaehoon and Jeffrey were the four core project contributors. All core contributors were closely involved throughout the duration of the project and made contributions to developing the theory, analyzing and debugging experiments, framing the narrative, reviewing code and giving feedback on writing. Katie led the project and led the theory, implemented and ran all experiments, produced all figures, and wrote the paper. Lechao made particular contributions to theory. Jaehoon made particular contributions to experimental design and framing the paper in relation to prior work, including drafting the related work section. Jeffrey made particular contributions to theory, experiment analysis, and paper framing and provided the primary advising on the project.

The experiments were implemented on top of a base Transformer model codebase. Peter led the development of this base model codebase and Roman, Mitchell, Jaehoon, Lechao and Katie made significant contributions to this codebase.

In addition, Mitchell contributed expertise on optimizers and weight decay. Alex contributed expertise on quantifying the uncertainty in exponent measurements. Roman implemented a small library used for reparameterization and gradient scaling. Jascha contributed to early discussions on parameterization and gradient scaling, contributed ideas about weight decay, and gave feedback on writing. Izzeddin contributed to technical discussions on alignment. Leslie contributed to technical discussions and gave feedback on paper framing and writing.

\subsection{Theoretical Details}
\label{app:theory}
In this section, we provide formal definitions and a complete derivation of the constraints for our alignment-general space of parameterizations.

\subsubsection{Model}
\label{app:theory_model}
Following a similar model and notation as \citet{yang2021tensoriv}, we consider a multilayer perceptron with $L$ hidden layers, input and output dimensionality $d$, hidden layer dimensionality $n$, and nonlinearity $\phi: \R \rightarrow \R$. The weight matrices are denoted:
\begin{itemize}
    \item $W_1 \in \R^{n \times d}$ for the embedding layer
    \item $W_2, \ldots W_L \in \R ^ {n \times n}$ for the hidden layers, and
    \item $W_{L+1} \in \R^{d \times n}$ for the readout layer.
\end{itemize}

The parameterization for each layer $l$ is specified by three values $\{a_l, b_l, c_l\}$, where:
\begin{itemize}
    \item the parameter multiplier is $n^{-a_l}$,
    \item the parameter initialization is $W_l \sim \mathcal{N}(0, n^{-2b_l})$, and
    \item  the learning rate $\eta_l \propto n^{-c_l}$ with width-independent constant of proportionality that we omit here.
\end{itemize}

For an input $x \in \R^d$, the model has activations $z_1, \ldots z_L$ and outputs logits $z_{L+1}$:
\begingroup
\begin{align*}
    z_1 &= \phi(n^{-a_1} W_1 \cdot x)\\
    z_l &= \phi(n^{-a_l} W_l \cdot z_{l-1}), \quad l \in [2, L]\\
    z_{L+1} &= n^{-a_{L+1}} W_{L+1} \cdot z_L
\end{align*}
\endgroup
There can be additional values prescribed in a width-scaling parameterization beyond the initialization scale, parameter multipliers and learning rate. For example, \citet{yang2023tensorivb} includes the epsilon hyperparameter, gradient clipping and weight decay in the parameterization for adaptive optimizers like Adam.

\subsubsection{Equivalence classes}
\label{app:theory_equivalence_classes}
These parameterizations occupy equivalence classes because in any layer we can ``factor out'' a constant term from the parameter initialization into the parameter multiplier, which exactly preserves the output of the forward pass while multiplying the gradients by this constant. This change in the gradients can then be ``corrected for'' by modifying the learning rate in an optimizer-specific manner.

In this one-dimensional symmetry group parameterized by $\theta$, to preserve the forward pass, regardless of the optimizer apply
\begin{align*}
    &a_l \leftarrow a_l + \theta\MoveEqLeft[1]\\
    &b_l \leftarrow b_l - \theta.\\
\intertext{Then specific to the optimizer, to preserve the effect of the backwards pass, correct the learning rate according to}
    \textrm{SGD:}\quad&c_l \leftarrow c_l - 2\theta\\
    \textrm{Adam:}\quad&c_l \leftarrow c_l - \theta\\
    \textrm{Adafactor:}\quad&c_l \leftarrow c_l.
\end{align*}

In particular, under the right learning rates, our four parameterizations occupy two equivalence classes: standard and NTK are equivalent and muP and mean-field parameterization are equivalent. In this paper, we will consider all four parameterizations separately, as these equivalences hold only under infinite precision, while neural networks regularly encounter finite-precision effects. These equivalences were observed for SGD and Adam in \citet{yang2021tensoriv} and \citet{yang2023tensorivb} respectively, and we propose this equivalence for Adafactor.

\subsubsection{Defining ``scale"}
\label{app:theory_scale}
Throughout this derivation, we are interested in the ``scale" of various quantities in the infinite-width limit, specifically the exponent with respect to width $n$ as the width becomes large.

\begin{appendixdef}
We say that the \emph{scale} of a quantity $U$ is $n^v$ if 
\begin{align*}
v = \lim_{n \rightarrow \infty} \log_n \norm{U}_{RMS}
\end{align*}
\end{appendixdef}
where the norm is the root-mean-square (RMS) norm. Intuitively, the RMS norm describes the size of ``typical" entries in a matrix: if all entries were the same then the RMS norm would match the value of each entry.

We use standard Big O notation or the $\sim$ symbol to denote this asymptotic behavior and write:

\begin{align*}
U = \Theta(n^v) \textrm{ or } U \sim n^v \textrm{ if } v =  \lim_{n \rightarrow \infty} \log_n \norm{U}_{RMS}\\
U = O(n^v) \textrm{ if } v \geq  \lim_{n \rightarrow \infty} \log_n \norm{U}_{RMS}\\
U = \Omega(n^v) \textrm{ if } v \leq  \lim_{n \rightarrow \infty} \log_n \norm{U}_{RMS}
\end{align*}

\subsubsection{Assumptions and Notation}
\label{app:theory_assumptions}
We will assume the following:
\begin{itemize}
    \setlength\itemsep{-0.2em}
    \item The input data is $\Theta(1)$.
    \item The number of layers $L$ is $O(1)$.
    \item The number of training steps $T$ is $O(1)$.
    \item The batch size is one.
    \item The input and output dimensionality $d$ is $O(1)$.
    \item The nonlinearity $\phi$ has bounded (weak) derivative, so the derivative of the nonlinearity does not contribute to the exponent in the infinite-width limit. As such, we omit the nonlinearity in the following calculations, equivalent to assuming $\phi$ is the identity function.
    \item We assume that the derivative of the loss with respect to the logits $\nabla_{z_{L+1}} \loss$ is $\Theta(1)$. Since the output dimensionality $d$ is $O(1)$, this assumption holds for many common loss functions.
    \item Our theoretical derivations use real numbers, i.e. assuming infinite precision, despite possible effects from finite precision in practice.
    \item We denote the difference in a quantity after initialization as $\Delta \bullet^t \coloneqq \bullet ^t - \bullet ^0$, for example $\Delta z_l^t \coloneqq z_l ^t - z_l ^0$.
    \item When the timestep is clear from the context, we omit the superscripts.
    \item Unless otherwise stated, a norm refers to the RMS norm.
    \item The learning rate $\eta_l$ is proportional to $n^{-c_l}$ with a width-independent proportionality constant that is typically determined empirically. For our derivations we will omit the proportionality constant and write $\eta_l = n^{-c_l}$.
    \item We use $\nabla_{W_l} \loss$ and $\partialloss{W_l}$ interchangeably.
\end{itemize}

\subsubsection{Defining stability and nontriviality}
\label{app:theory_stability_nontriviality_def}

We will use the definitions from \citet{yang2021tensoriv} for stability and nontriviality:
\begin{appendixdef}
A parameterization is \emph{stable} if the activations have exactly constant scale, i.e. $z_l^t = \Theta(1) \enspace \forall l \in [1,L]$ and the logits are at most constant scale, i.e. $z_{L+1} = O(1)$, at all timesteps $0 \leq t \leq T$ during training.
\end{appendixdef}
\begin{appendixdef}
A parameterization is \emph{nontrivial} if the change in logits after initialization is at least constant scale, i.e. $z_{L+1}^t - z_{L+1}^0 = \Omega(1)$ for some timestep $0 \leq t \leq T$ during training.
\end{appendixdef}
The specific choice to require exactly constant scale activations and at most constant scale logits should be thought of as a design choice from which theoretical results follow rather than a theoretical result itself.

\subsubsection{First Forward Pass: Stability at initialization}
\label{app:theory_first_forward}
The stability constraints at initialization ensure that all intermediate activations $z_l$ are $\Theta(1)$ and the logits $z_{L+1}$ are $O(1)$. The constraints apply iteratively across $O(1)$ layers: since the input $x$ is $O(1)$, the constraint on the first layer ensures that $z_1$ is $O(1)$, then the constraint on layer $l$ ensures that $z_l$ is $O(1)$ assuming the previous layer $l-1$ constraint are satisfied so that $z_{l-1}$ is $O(1)$.

This gives the constraints for stability at initialization:
\begin{align*}
    &a_1 + b_1 = 0\\
    &a_l + b_l = 1/2,\quad l \in [2, \ldots, L]\\
    &a_{L+1} + b_{L+1} \geq 1/2
\end{align*}

\subsubsection{Gradients at Initialization}
\label{app:theory_gradients_init}
At initialization, the gradients for each layer can be calculated using straightforward application of the chain rule. We first define $g_l^t$ as the negative exponent of the gradient scale of the loss with respect to the parameters in layer $l$ at timestep $t$.

\begin{appendixdef}
Let $\displaystyle g_l^t = -\lim_{n \rightarrow \infty} \log_n \norm{\frac{\partial \mathcal{L}}{\partial W_l^t}}$ so that $\frac{\partial \mathcal{L}}{\partial W_l} = \Theta(n ^ {-g_l})$.
\end{appendixdef}

Then by the chain rule, the gradient decomposes as
\begin{align*}
    \partialloss{W_l} = \partialloss{z_{L+1}} \partialfrac{z_{L+1}}{z_L} \cdots \partialfrac{z_{l+1}}{z_l} \partialfrac{z_l}{W_l}
\end{align*}
where $\partialfrac{z_{L+1}}{z_L} = \Theta(1)$ by assumption, $\partialfrac{z_{l}}{z_{l-1}} \sim n^{-a_{l}} \cdot n^{-b_{l}}$ and $\partialfrac{z_l}{W_l} \sim n^{-a_l}$.

After taking the logarithm and flipping the negative signs, this gives
\begin{align*}
    g_l = a_{L+1} + b_{L+1} + \left(\: \sum_{i = l+1} ^{L} (a_i + b_i - 1/2) \right) + a_l
\end{align*}

If we then assume the stability at initialization constraints, the terms inside the sum for all hidden layers cancel, leaving that the gradients at initialization are:
\begin{align*}
    g_l &= a_l + a_{L+1} + b_{L+1} \textrm{ for } l \in [1, \ldots, L]\\
    g_{L+1} &= a_{L+1}
\end{align*}

\subsubsection{Optimizer Update Rules}
\label{app:theory_optimizer_updates}
We write out the version of the update rules that we use for these derivations for each optimizer family, which include the aspects that are essential to the scaling exponents but omit more specific features like momentum or moving averages, learning rate schedules, weight decay, clipping, and low-rank factoring. For intuition, it is useful to consider the relationship between the scale of the updates, gradients, parameters, and learning rate for each optimizer. In SGD, the scale of the update matches the scale of the learning rate times the scale of the gradients. In Adam, or similar adaptive optimizers that normalize by the gradient scale, the scale of the updates match the scale of the learning rate regardless of the gradient scale. In Adafactor, Adam with parameter scaling, or similar optimizers that normalize by the gradient scale and then multiply by the parameter scale, the scale of the updates matches the scale of the learning rate times the scale of the parameters.

\begin{align*}
    \textrm{SGD:}\hquad &\Delta W_l = \eta_l \cdot \nabla_{W_l} \loss\\
    \textrm{Adam:}\hquad &\Delta W_l = \eta_l \cdot \frac{\nabla_{W_l} \loss}{\norm{\nabla_{W_l} \loss}}\\
    \textrm{Adafactor:}\hquad &\Delta W_l = \eta_l \cdot \norm{W_l} \cdot \frac{\nabla_{W_l} \loss}{\norm{\nabla_{W_l} \loss}}\\
\end{align*}

\subsubsection{First Backward Pass}
\label{app:theory_first_backward}
Using the update rules in the previous section, we write out the update for each optimizer during the first backward pass. We note here that so far the calculations and constraints have been the same for all optimizers, and this step is the first one that is specific to the optimizer based on its update rule.

\begin{align*}
\textrm{SGD:} \hquad &\Delta W_l = \eta_l \cdot \nabla_{W_l} \loss \sim n^{-c_l} \cdot n^{-g_l},\\
\textrm{where}\hquad &g_{L+1} = a_{L+1} \textrm{ or } g_l = a_l + a_{L+1} + b_{L+1},\\
\textrm{so}\hquad &\Delta W_l \sim n^{-a_{L+1} - b_{L+1} - a_l - c_l},\\
&\Delta W_{L+1} \sim n^{-a_{L+1} - c_{L+1}}\\ \\
\textrm{Adam: }\hquad &\Delta W_l = \eta_l \cdot \frac{\nabla_{W_l} \mathcal{L}}{\norm{\nabla_{W_l} \mathcal{L}}} \sim n^{-c_l} \cdot 1 \\
  \textrm{so } &\Delta W_l \sim n^{-c_l}\\ \\
\textrm{Adafactor:}\hquad &\Delta W_l = \eta_l \cdot \norm{W_l} \cdot \frac{\nabla_{W_l} \mathcal{L}}{\norm{\nabla_{W_l} \mathcal{L}}} \sim n^{-c_l} \cdot n^{-b_l} \cdot 1\\
    \textrm{so } &\Delta W_l \sim n^{-c_l - b_l}\\
\end{align*}

\subsubsection{Defining the activation update residual}
We next define a feature learning residual quantity $r_l$ that measures how far the parameterization is from the feature learning regime. For each layer $l$ in $[1, L]$, we define $r_l$ as the negative exponent of the scale of $\Delta z_l$, where $\Delta z_l$ the change in activations following layer $l$ during training. To preserve stability, this change cannot exceed constant scale, so $r_l$, as the negative exponent, cannot be less than zero. Feature learning, where the change in activations immediately prior to the readout layer has constant scale, then corresponds to $r_L = 0$ exactly. Conceptually, feature learning occurs if at least one of the embedding or hidden layers contributes at least one constant scale term to the activations.

\begin{appendixdef}
For all $l$ in $[1, L]$, let $\displaystyle r_l \coloneqq - \lim_{n \rightarrow \infty} \log_n \norm{\Delta z_l}$, so that $\Delta z_l \sim n^{-r_l}$.
\end{appendixdef}

\subsubsection{Defining Alignment Variables}
\label{app:theory_align_vars}

In this section, we will define three alignment variables $\alpha_l, \omega_l$ and $u_l$ that are the exponents of the alignment contributions from the $\Delta W_l \cdot z_{l-1} $, $W_l \cdot \Delta z_{l-1}$ and $ \Delta W_l \cdot \Delta z_{l-1}$ terms respectively.

Starting in the second forward pass, each activation for layer $l$ in $[2,L+1]$ expands into four terms:
\begin{align*}
    z_l &= n^{-a_l}(W_l + \Delta W_l)(z_{l-1} + \Delta z_{l-1})\\
    &= n^{-a_l}( W_l \cdot z_{l-1} +  \cdot W_l \Delta z_{l-1}  + \Delta W_l \cdot z_{l-1}  + \Delta W_l \cdot \Delta z_{l-1})
\end{align*}

Due to the random initialization, the $W_l \cdot z_{l-1}$ term has no alignment. For the remaining three terms, we introduce the following alignment variables.
\begingroup
\begin{appendixdef} We define the alignment variables as
\begin{align*}
\alpha_l &=  \lim_{n \rightarrow \infty} \log_n \frac{\norm{ \Delta W_l z_{l-1}}}{\norm{\Delta W_l} \norm{z_{l-1}}}\quad\textrm{so that}\quad\Delta W_l z_{l-1} \sim n^{\alpha_l} \norm{\Delta W_l}\norm{z_{l-1}},\\
\omega_l &=  \lim_{n \rightarrow \infty} \log_n \frac{\norm{W_l \Delta z_{l-1}}}{\norm{W_l}\norm{\Delta z_{l-1}}}\quad\textrm{so that}\quad W_l\Delta z_{l-1} \sim n^{\omega_l} \norm{ W_l}\norm{\Delta z_{l-1}},\\
u_l &= \lim_{n \rightarrow \infty} \log_n \frac{\norm{\Delta W_l \Delta z_{l-1}}}{\norm{\Delta W_l}\norm{\Delta z_{l-1}}}\quad\textrm{so that}\quad \Delta W_l \Delta z_{l-1} \sim n^{u_l} \norm{\Delta W_l}\norm{\Delta z_{l-1}}.\\
\end{align*}
\end{appendixdef}
\endgroup
We have omitted the timestep superscripts above, but alignment is a dynamic quantity so more formally we have 
\begin{align*}
\alpha_l^t \coloneqq  \lim_{n \rightarrow \infty} \log_n \frac{\norm{ \Delta W_l^t z_{l-1}^t}}{\norm{\Delta W_l^t} \norm{z_{l-1}^t}}
\end{align*}
and similarly for $\omega_l^t$ and $u_l^t$.

Note that for many quantities in our notation, we define the variable to be a negative exponent, but for these alignment variables we are defining $\alpha_l$, $\omega_l$, $u_l$ as positive exponents so they take on values between $0$ and $1$.

\subsubsection{Second Forward Pass: Stability During Training}
\label{app:theory_second_forward}
To derive stability constraints for the second forward pass that ensure all intermediate activations are exactly constant scale and the logits are at most constant scale, we will proceed starting from the embedding layer $l$, followed by the hidden layers $l$ in $[2,L]$ and finally the readout layer $L+1$. These constraints work iteratively across layers: the first constraints will ensure $z_1 = \Theta(1)$, and then the subsequent constraints will ensure $z_{l} = \Theta(1)$ assuming that $z_{l-1} = \Theta(1)$, and finally the readout constraints will ensure that $z_{L+1} = O(1)$ assuming $z_{L} = \Theta(1)$.

In the second forward pass, the embedding layer activations are
\begin{align*}
    z_1^1 &= n^{-a_1} (W_1^0 + \Delta W_1^1)x\\
    &= n^{-a_1}W_1^0x + n^{-a_1}\Delta W_1^1x\\
    &= z_1^0 + n^{-a_1}\Delta W_1^1x.\\
\end{align*}
Recall that the input $x$ is $O(1)$ and that the input dimensionality $d$ that is the interior dimension in the $W_1 \cdot x$ term is $O(1)$. Since $z_1^0 = \Theta(1)$ by the stability at initialization constraints, we have $\Delta z_1^1 = n^{-a_1}\Delta W_1^1x = O(1) \iff z_1^1 = \Theta(1)$. Then by plugging in $\Delta W_1^1$ for each optimizer from \sref{app:theory_first_backward}, we have

\begingroup
\renewcommand{\arraystretch}{1.5}
\begin{table}[h!]
\centering
\begin{tabularx}{\textwidth}{>{\raggedleft\arraybackslash}X V{1}>{\raggedright\arraybackslash}X V{1}>{\raggedright\arraybackslash}X V{1}>{\raggedright\arraybackslash}X}
 & \multicolumn{1}{c|}{SGD} & \multicolumn{1}{c|}{Adam} & \multicolumn{1}{c}{Adafactor}\\ \hline
$\Delta W_1^1 \sim$ & $n^{-a_{L+1}-b_{L+1} - a_l - c_l}$    &  $n^{-c_1}$    &   $n^{-b_1 -c_1}$       \\
$\Delta z_1^1 = n^{-a_1}\Delta W_1^1x \sim$ &  $n^{-a_{L+1} - b_{L+1} - 2a_1 - c_1}$   &  $n^{-a_1 - c_1}$    &    $n^{-a_1 - b_1 - c_1}$\\
$\Delta z_1^1 = O(1) \Leftrightarrow$ & $a_{L+1} + b_{L+1} + 2a_1 + c_1 \geq 0$    &  $a_1 + c_1 \geq 0$    &   $c_1 \geq 0$ since $a_1 + b_1 = 0$       \\ 
\end{tabularx}
\end{table}
\endgroup

\FloatBarrier

Next, for the hidden layer activations we have
\begin{align*}
    z_l^1 &= n^{-a_l}W_l^1 z_{l-1}^1 = n^{-a_l}(W_l^0 + \Delta W_l^1)(z_{l-1}^0 + \Delta z_{l-1}^1)\\
    &= n^{-a_l}W_l^0 z_{l-1}^0 + n^{-a_l}W_l^0 \Delta z_{l-1}^1 + n^{-a_l}\Delta W_l^1 z_{l-1}^0 + n^{-a_l}\Delta W_l^1 \Delta z_{l-1}^1
\end{align*}
where $z_l^0 = \Theta(1)$ by the stability at initialization constraints and we assume that $z_{l-1}^1 = \Theta(1)$ by these constraints on the previous layer. This gives us four terms to bound, and in the table below we write one row for each term and in the columns we write the constraints needed to bound that term for the relevant optimizer.

\begingroup
\renewcommand{\arraystretch}{2.5}
\begin{table}[h!]
\centering
\adjustbox{scale=0.9}{
\begin{tabular}{V{1}>{\raggedleft\arraybackslash}r V{1}>{\raggedright\arraybackslash}l V{1}>{\raggedright\arraybackslash}l V{1}>{\raggedright\arraybackslash}l V{1}}
\hline
 & \multicolumn{1}{c|}{SGD} & \multicolumn{1}{c|}{Adam} & \multicolumn{1}{c|}{Adafactor}\\ \hline
$n^{-a_l} W_l^0 z_{l-1}^0$ & \multicolumn{3}{c|}{$a_l + b_l - 1/2 = 0$ by stability at init so no constraint required}       \\ \hline
$n^{-a_l} W_l^0 \Delta z_{l-1}^1$ &  \multicolumn{3}{c|}{$1/2 + r_{l-1} - \omega_l \geq 0$} \\ \hline
$n^{-a_l} \Delta W_l^1 z_{l-1}^0$ & $a_{L+1} + b_{L+1} + 2a_l + c_l - \alpha_l \geq 0$    &  $a_l + c_l - \alpha_l \geq 0$   &   $1/2 + c_l - \alpha_l \geq 0$       \\ \hline
$n^{-a_l} \Delta W_l^1 \Delta z_{l-1}^1$ & $a_{L+1} + b_{L+1} + 2a_l + c_l + r_{l-1} - u_l \geq 0$    &  $a_l + c_l + r_{l-1} - u_l \geq 0$   &   $1/2 + c_l + r_{l-1} - u_l \geq 0$       \\ \hline
\end{tabular}}
\end{table}
\endgroup
\FloatBarrier
Finally, for the logits we have
\begin{align*}
    z_{L+1}^1 &= n^{-a_{L+1}}W_{L+1}^1 z_{L}^1 = n^{-a_{L+1}}(W_{L+1}^0 + \Delta W_{L+1}^1)(z_{L}^0 + \Delta z_{L}^1)\\
    &= n^{-a_{L+1}}W_{L+1}^0 z_{L}^0 + n^{-a_{L+1}}W_{L+1}^0 \Delta z_{L}^1 + n^{-a_{L+1}}\Delta W_{L+1}^1 z_{L}^0 + n^{-a_{L+1}}\Delta W_{L+1}^1 \Delta z_{L}^1
\end{align*}
where $z_L^0 = \Theta(1)$ by stability at initialization and $z_L^1 = \Theta(1)$ by the constraints on the hidden layers, and we want to find the constraints so that $z_{L+1}^1 = O(1)$.

Similar to the hidden activations, we have four terms to bound and show the constraints for each term and optimizer in the following table:

\begingroup
\begin{table}[h!]
\renewcommand{\arraystretch}{2.5}
\centering
\adjustbox{scale=0.8}{
\begin{tabular}{V{1}>{\raggedleft\arraybackslash}r V{1}>{\raggedright\arraybackslash}l V{1}>{\raggedright\arraybackslash}l V{1}>{\raggedright\arraybackslash}l V{1}}
\hline
 & \multicolumn{1}{c|}{SGD} & \multicolumn{1}{c|}{Adam} & \multicolumn{1}{c|}{Adafactor}\\ \hline
$n^{-a_{L+1}} W_{L+1}^0 z_{L}^0$ & \multicolumn{3}{c|}{$a_{L+1} + b_{L+1} - 1/2 \geq 0$ by stability at init so no constraint required}       \\ \hline
$n^{-a_{L+1}} W_{L+1}^0 \Delta z_{L}^1$ &  \multicolumn{3}{c|}{$a_{L+1} + b_{L+1} + r_{L} - \omega_{L+1} \geq 0$} \\ \hline
$n^{-a_{L+1}} \Delta W_{L+1}^1 z_{L}^0$ & $2a_{L+1} + c_{L+1} - \alpha_{L+1} \geq 0$    &  $a_{L+1} + c_{L+1} - \alpha_{L+1} \geq 0$   &   $a_{L+1} + b_{L+1} + c_{L+1} - \alpha_{L+1} \geq 0$       \\ \hline
$n^{-a_{L+1}} \Delta W_{L+1}^1 \Delta z_{L}^1$ &  $2a_{L+1} + c_{L+1} + r_L - u_{L+1} \geq 0$    &  $a_{L+1} + c_{L+1} + r_L - u_{L+1} \geq 0$   &   $a_{L+1} + b_{L+1} + c_{L+1} + r_L - u_{L+1} \geq 0$ \\ \hline
\end{tabular}}
\end{table}
\endgroup
\FloatBarrier
\subsubsection{Third and Subsequent Forward Passes: Stability During Training}
\label{app:theory_third_forward}
For the third and subsequent forward passes, there are slight modifications required to the stability constraints from the second forward pass. Since we require the activations to be exactly constant scale at initialization, the parameter updates for the embedding and hidden layers are never larger in scale than the initial parameters and therefore never dominate the contribution from the initial parameters to the activations following that layer. However, the readout parameters might have updates that are larger in scale than the initialization, so we need to calculate the scale of the readout parameters after the first update and then consider how this changes the constraints on each optimizer.

For SGD, after the first update we have $W_{L+1}^1 = W_{L+1}^0 + \Delta W_{L+1}^1 \sim \max(-b_{L+1}, -a_{L+1} - c_{L+1})$. This changes the gradients for all layers before the readout layer, which were $g_l^0 = a_{L+1} + b_{L+1} + a_l$, and are now $g_l^1 = \max(a_{L+1} + b_{L+1}, 2a_{L+1} + c_{L+1}) + a_l$. We account for this by replacing the constraints
\begin{align*}
    \begin{cases}
    &g_1^0 + a_1 + c_1 = a_{L+1} + b_{L+1} + 2a_1 + c_1 \geq 0\\
    &g_l^0 + a_l + c_l - \alpha_l = a_{L+1} + b_{L+1} + 2a_l + c_l - \alpha_l \geq 0\\
    &g_l^0 + a_l + c_l + r_{l-1} - u_l = a_{L+1} + b_{L+1} + 2a_l + c_l + r_{l-1} - u_l \geq 0\\
    \end{cases}
\end{align*}
with 
\begin{align*}
    \begin{cases}
    &g_1^1 + a_1 + c_1 = \max(a_{L+1} + b_{L+1}, 2a_{L+1} + c_{L+1}) + 2a_1 + c_1 \geq 0\\
    &g_l^1 + a_l + c_l - \alpha_l = \max(a_{L+1} + b_{L+1}, 2a_{L+1} + c_{L+1}) + 2a_l + c_l - \alpha_l \geq 0\\
    &g_l^1 + a_l + c_l + r_{l-1} - u_l = \max(a_{L+1} + b_{L+1}, 2a_{L+1} + c_{L+1}) + 2a_l + c_l + r_{l-1} - u_l \geq 0\\
    \end{cases}
\end{align*}

For Adam, even if the readout parameters do increase in scale after initialization, leading to increased gradient scales, the Adam update scale does not depend on the gradient scale so the existing constraints are sufficient.

For Adafactor, similar to Adam we do not require an additional constraint as a result of a change in gradient scales, but there is one additional constraint required due to the parameter scaling: we require $c_{l} \geq 0$ to avoid exponential growth as $n^{-c_{l} \cdot t}$ across steps $t$.

Finally, by induction over the steps, combining all the above constraints ensures stability for any time $t \leq T$. Note that it is essential that we assumed the number of training steps $T$ is $O(1)$ so that this induction step does not introduce any width dependence.

\subsubsection{Nontriviality}
\label{app:theory_nontriviality}
Recall that a parameterization is nontrivial if the change in logits after initialization is at least constant scale. This corresponds to exact equality on one of the stability constraints on the logits, specifically
\begingroup
\renewcommand{\arraystretch}{1.5}
\begin{table}[h!]
\centering
\adjustbox{scale=0.9}{
\begin{tabularx}{\textwidth}{>{\centering\arraybackslash}X V{1}>{\centering\arraybackslash}X V{1}>{\centering\arraybackslash}X}
 \multicolumn{1}{c|}{SGD} & \multicolumn{1}{c|}{Adam} & \multicolumn{1}{c}{Adafactor}\\ \hline
 $a_{L+1} + b_{L+1} + r_L - \omega_{L+1} = 0$    &  $a_{L+1} + b_{L+1} + r_L - \omega_{L+1} = 0$     &   $a_{L+1} + b_{L+1} + r_L - \omega_{L+1} = 0$        \\
 or & or & or \\
  $2a_{L+1} + c_{L+1} - \alpha_{L+1} = 0$   &  $a_{L+1} + c_{L+1} - \alpha_{L+1} = 0$    &    $a_{L+1} + b_{L+1} + c_{L+1} - \alpha_{L+1} = 0$\\
  or & or & or \\
  $2a_{L+1} + c_{L+1} + r_L - u_{L+1} = 0$   &  $a_{L+1} + c_{L+1} + r_L - u_{L+1} = 0$    &    $a_{L+1} + b_{L+1} + c_{L+1} + r_L - u_{L+1} = 0$
\end{tabularx}}
\end{table}
\endgroup

\subsubsection{Summary of Constraints}
\label{app:theory_summary_constraints}
In Table 2, we summarize the full set of stability and nontriviality constraints derived in the previous sections, which define the alignment-general space of parameterizations.
\clearpage
\newcommand{\STAB}[1]{\begin{tabular}{@{}c@{}}#1\end{tabular}}

\begingroup
\renewcommand{\arraystretch}{2.5}
\begin{table*}[h!]
\centering
\adjustbox{scale=0.86}{
\begin{footnotesize}
\begin{tabular}{c | c | c| c}
 & \normalsize{\textbf{SGD}} & \normalsize{\textbf{Adam}} & \normalsize{\textbf{Adafactor}}\\ \hline
 \multirow{3}{*}[1.2ex]{\rotatebox[origin=c]{90}{\parbox[l]{1.75cm}{\centering{\textbf{Stability at initialization}}}}} & \multicolumn{3}{c}{$a_1 + b_1 = 0$}\\[-1.5ex]
     & \multicolumn{3}{c}{$a_l + b_l = 1/2$ for $l \in [2, L]$} \\[-1.5ex]
     & \multicolumn{3}{c}{$a_{L+1} + b_{L+1} \geq 1/2$} \\ \hline
  \multirow{4}{*}{\rotatebox[origin=c]{90}{\parbox[c]{2.4cm}{\centering\textbf{Stable activations during training}}}} & $r_1 \coloneqq g_1 + a_1 + c_1 \geq 0$    &  $r_1 \coloneqq a_1 + c_1 \geq 0$    &   $r_1 \coloneqq c_1 \geq 0$\\
  & $r_l \coloneqq \min \begin{cases}g_l + a_l + c_l - \alpha_l\\ g_l + a_l + c_l + r_{l-1} - u_l\\ 1/2 + r_{l-1} - \omega_l\end{cases}\hspace{-0.5em}\geq 0$ & $r_l \coloneqq \min \begin{cases}a_{l} + c_{l} - \alpha_l\\a_{l} + c_{l} + r_{l-1} - u_{l}\\1/2 + r_{l-1} - \omega_l\end{cases}\hspace{-0.5em} \geq 0$    &    $r_l \coloneqq \min \begin{cases}1/2 + c_{l} - \alpha_l\\1/2 + c_{l} + r_{l-1} - u_{l}\\ 1/2 + r_{l-1} - \omega_l\end{cases}\hspace{-0.5em}\geq 0$\\
  & \multicolumn{1}{l|}{where $g_i \coloneqq$} & & $c_l \geq 0$\\[-3ex]
  & $\max(a_{L+1} + b_{L+1}, 2a_{L+1} + c_{L+1}) + a_i$& & \\ \hline
  \multirow{2}{*}[2.5ex]{\rotatebox[origin=c]{90}{\parbox[c]{2.4cm}{\centering\textbf{Stable logits during training}}}} & $\min \begin{cases}a_{L+1} + b_{L+1} + r_L - \omega_{L+1}\\ 2a_{L+1} + c_{L+1} - \alpha_{L+1} \\ 2a_{L+1} + c_{L+1} + r_L - u_{L+1} \end{cases}\hspace{-0.5em}\geq 0$ \rule{0pt}{8ex} &  $\min \begin{cases}a_{L+1} + b_{L+1} + r_L - \omega_{L+1}\\a_{L+1} + c_{L+1} - \alpha_{L+1}\\a_{L+1} + c_{L+1} +r_L - u_{L+1} \end{cases}\hspace{-0.5em}\geq 0$     &   $\min \begin{cases}a_{L+1} + b_{L+1} + r_L - \omega_{L+1}\\a_{L+1} + b_{L+1} + c_{L+1} - \alpha_{L+1}\\a_{L+1} + b_{L+1} + c_{L+1} +r_L - u_{L+1} \end{cases}\hspace{-0.5em}\geq 0$\\
  & & & $c_{L+1} \geq 0$ \\\hline
  \multirow{3}{*}{\STAB{\rotatebox[origin=c]{90}{\textbf{Nontriviality}}}} & $a_{L+1} + b_{L+1} + r_L - \omega_{L+1} = 0$   &  $a_{L+1} + b_{L+1} + r_L - \omega_{L+1} = 0$     &   $a_{L+1} + b_{L+1} + r_L - \omega_{L+1} = 0$        \\
 & or \hquad $2a_{L+1} + c_{L+1} - \alpha_{L+1} = 0$   &  or \hquad $a_{L+1} + c_{L+1} - \alpha_{L+1} = 0$    &    or \hquad $a_{L+1} + b_{L+1} + c_{L+1} - \alpha_{L+1} = 0$\\
  & or \hquad $2a_{L+1} + c_{L+1} + r_L - u_{L+1} = 0$   &  or \hquad$a_{L+1} + c_{L+1} + r_L - u_{L+1} = 0$    &    or \hquad$a_{L+1} + b_{L+1} + c_{L+1} + r_L - u_{L+1} = 0$\\
\end{tabular}
\end{footnotesize}}
\vspace{2pt}
\caption{Summary of stability and nontriviality constraints for our alignment-general space of parameterizations.}
\label{tab:app_stability_training_constraints}
\end{table*}
\endgroup

\subsubsection{Tensor Programs as a Special Case}

When we assume $\alpha_l = 1 \;\forall l \in [2, L+1]$, $\omega_l = 1/2$ for $l \in [2, L]$, and $\omega_{L+1} = 1$, by plugging these values into our constraints we recover exactly the stability and nontriviality constraints in~\citet{yang2021tensoriv,yang2023tensorivb}. These assumptions are the necessary and sufficient conditions to recover their constraints exactly. In particular, their constraints imply no assumption on $u_l$ as their $\alpha_l = 1$ is maximal so $\alpha_l \geq u_l$ in all cases and the $\Delta z_{l-1} \Delta W_l$ term never dominates the $z_{l-1} \Delta W_l$ term.

\subsubsection{Maximum Stable Learning Rates for All Parameterizations}

In \cref{tab:common_parameterizations} (repeated here), we compute the maximum stable per-layer learning rate exponents under two specific alignment assumptions: ``full alignment" where $\alpha_l = u_l = 1$, and ``no alignment" where $\alpha_l = u_l = 1/2,\,l \in [2, L+1]$. In both of these settings, we assume $\omega_{l} = 1/2,\, l\in[2, L+1]$. This $\omega_{L+1}$ term is the alignment exponent on the $\Delta z_{L} W_{L+1}$ term, which quantifies the alignment between parameter updates in earlier layers that contribute to $\Delta z_L$ and the initialization in the readout layer $W_{L+1}$. Our $\omega_{L+1} = 1/2$ relaxes the $\omega_{L+1} = 1$ assumption in \citet{yang2021tensoriv,yang2023tensorivb}.

The maximal learning rate exponents follow by first plugging in the values for $\alpha_l, \omega_l$, and $u_l$ and then solving for the minimal value of $c_l$ (where minimal $c_l$ corresponds to the maximal learning rate, as $c_l$ is the negative exponent) that satisfies the stability constraints in each layer $l$. Due to the relaxation with $\omega_{L+1} = 1/2$, for all parameterizations and optimizers in both our alignment settings, this results in values of $c_l$ that make $r_l = 0$ for all $l \in [2,L+1]$, indicating that all layers are being updated maximally and that the parameterization is in a feature learning limit.

For standard and NTK parameterizations, our full alignment per-layer learning rate prescriptions differ from prior work, and can attain feature learning. For muP and MFP, our full alignment per-layer learning rates coincide exactly for SGD and Adam in \citet{yang2021tensoriv} and \citet{yang2023tensorivb} respectively as the $\omega_{L+1}$ term does not constrain the learning rates in the embedding and hidden layers in those parameterizations.

\begingroup
\captionsetup[table]{labelformat = repeattable}
\setcounter{stashtablecounter}{\value{table}}
\setcounter{table}{\therepeattablecounter}
\renewcommand\thetable{\arabic{table}}
\begin{table*}
  \caption{Left: Parameterizations and gradients at initialization for width $n$. Middle: Max stable per-layer learning rate scaling for each optimizer assuming $\alpha_l = 1, \omega_l = 1/2$ for all layers $l$. Right: Max stable learning rates assuming $\alpha_l = \omega_l = u_l = 1/2$ for all layers.}
  \vspace{2pt}
  \centering
  \begin{footnotesize}
    \setlength\extrarowheight{2.5pt}
    \adjustbox{scale=0.8}{
    \begin{tabular}
        {r | l | p{1.45cm}| p{1.3cm} | p{1.1cm} || p{1.3cm} 
        | p{1.3cm} 
        | p{1.8cm}  || p{1.3cm}
        | p{1.3cm}
        | p{1.8cm} |}
        \hline
        \multicolumn{2}{ V{1} r V{1}}{}
        & Initialization Variance & Parameter Multiplier & Gradient & SGD LR, Full Align & Adam LR, Full Align & Adafactor LR, Full Align & SGD LR, No Align & Adam LR, No Align & Adafactor LR, No Align
        \\ 
        \hline
        \multicolumn{1}{ V{1} r V{1}}{\multirow{3}{*}{Standard}} & Embedding & 
        $1\hphantom{/n}$    & $1\hphantom{/\sqrt{n}}$                  & $1/\sqrt{n}$                  & $\sqrt{n}\hphantom{/\sqrt{n}}$         & $1\hphantom{/\sqrt{n}}$ & 1  & $\sqrt{n}\hphantom{/\sqrt{n}}$         & $1\hphantom{/\sqrt{n}}$ & 1              
        \\ 
        \multicolumn{1}{ V{1} r V{1}}{}                                   & Hidden    & 
        $1/n$               & $1\hphantom{/\sqrt{n}}$                  & $1/\sqrt{n}$                  & $1/\sqrt{n}$                           & $1/n\hphantom{n}$               & $1/\sqrt{n}$ & $1\hphantom{/\sqrt{n}}$  & $1/\sqrt{n}$ & 1 
        \\ 
        \multicolumn{1}{ V{1} r V{1}}{}                                   & Readout   & 
        $1/n$               & $1\hphantom{/\sqrt{n}}$                  & $1\hphantom{/\sqrt{n}}$       & $1/n\hphantom{n}$                      & $1/n\hphantom{n}$                & $1/\sqrt{n}$ & $1/\sqrt{n}$ & $1/\sqrt{n}$ & 1 
        \\ 
        \hline
        \multicolumn{1}{ V{1} r V{1}}{\multirow{3}{*}{NTK}}      & Embedding & 
        $1\hphantom{/n}$    & $1\hphantom{/\sqrt{n}}$                  & $1/\sqrt{n}$                  & $\sqrt{n}\hphantom{/\sqrt{n}}$         & $1\hphantom{/\sqrt{n}}$  & 1 & $\sqrt{n}\hphantom{/\sqrt{n}}$         & $1\hphantom{/\sqrt{n}}$  & 1               
        \\ 
        \multicolumn{1}{ V{1} r V{1}}{}                                   & Hidden    & 
        $1\hphantom{/n}$    & $1/\sqrt{n}$                             & $1/n\hphantom{n}$             & $\sqrt{n}\hphantom{/\sqrt{n}}$         & $1/\sqrt{n}$      & $1/\sqrt{n}$ & $n\hphantom{/\sqrt{n}}$ & $1\hphantom{/\sqrt{n}}$ & 1   
        \\ 
        \multicolumn{1}{ V{1} r V{1}}{}                                   & Readout   & 
        $1\hphantom{/n}$    & $1/\sqrt{n}$                             & $1/\sqrt{n}$                  & $1\hphantom{/\sqrt{n}}$                & $1/\sqrt{n}$        & $1/\sqrt{n}$ & $\sqrt{n}\hphantom{/\sqrt{n}}$ & $1\hphantom{/\sqrt{n}}$ & 1 
        \\ 
        \hline
        \multicolumn{1}{ V{1} r V{1}}{\multirow{3}{*}{muP}}      & Embedding & 
        $1/n$               & $\sqrt{n}\hphantom{/\sqrt{n}}$           & $1/\sqrt{n}$                  & $1\hphantom{/\sqrt{n}}$                & $1/\sqrt{n}$     & 1    & $1\hphantom{/\sqrt{n}}$                & $1/\sqrt{n}$     & 1 
        \\ 
        \multicolumn{1}{ V{1} r V{1}}{}                                   & Hidden    & 
        $1/n$               & $1\hphantom{/\sqrt{n}}$                  & $1/n\hphantom{n}$             & $1\hphantom{/\sqrt{n}}$                & $1/n\hphantom{n}$       & $1/\sqrt{n}$    & $\sqrt{n}\hphantom{/\sqrt{n}}$ &  $1/\sqrt{n}$ & 1      
        \\ 
        \multicolumn{1}{ V{1} r V{1}}{}                                   & Readout   & 
        $1/n$               & $1/\sqrt{n}$                             & $1/\sqrt{n}$                  & $1\hphantom{/\sqrt{n}}$                & $1/\sqrt{n}$         & $1$ & 1 & $1\hphantom{/\sqrt{n}}$ & 1 
        \\ 
        \hline
        \multicolumn{1}{ V{1} r V{1}}{\multirow{3}{*}{MFP}}      & Embedding & 
        $1\hphantom{/n}$    & $1\hphantom{/\sqrt{n}}$                  & $1/n\hphantom{n}$             & $n\hphantom{/\sqrt{n}}$                & $1\hphantom{/\sqrt{n}}$     & 1       & $n\hphantom{/\sqrt{n}}$                & $1\hphantom{/\sqrt{n}}$     & 1       
        \\ 
        \multicolumn{1}{ V{1} r V{1}}{}                                   & Hidden    & 
        $1\hphantom{/n}$    & $1/\sqrt{n}$                             & $1/n^{1.5}\hspace{-2pt}$      & $n\hphantom{/\sqrt{n}}$                & $1/\sqrt{n}$         & $1/\sqrt{n}$ & $n^{1.5}\hphantom{/\sqrt{n}}$  & $1\hphantom{/\sqrt{n}}$ & 1 
        \\ 
        \multicolumn{1}{ V{1} r V{1}}{}                                   & Readout   & 
        $1\hphantom{/n}$    & $1/n\hphantom{n}$                        & $1/n\hphantom{n}$             & $n\hphantom{/\sqrt{n}}$                & $1\hphantom{/\sqrt{n}}$       & $1$          & $n\hphantom{/\sqrt{n}}$  & $\sqrt{n}\hphantom{/\sqrt{n}}$ & 1 
        \\ 
        \hline
    \end{tabular}
    }
    \vspace{-12pt}
    \end{footnotesize}
\end{table*}
\endgroup
\setcounter{table}{\thestashtablecounter}

We note here that throughout the paper, we consider from a theoretical perspective what the maximum stable learning rate exponents should be, but empirically we are interested in the optimal learning rate. It is not necessarily the case that the maximum stable learning rate and optimal learning rate scale with the same exponents, and future work could more carefully investigate the relationship between these two entities.
\clearpage

\subsection{Experimental Details}\label{app:expt_details}

\subsubsection{Architecture and training details}
All experiments use the NanoDO~\citep{nanodo} decoder-only Transformer architecture employing learned positional embeddings, pre-layer norm~\citep{xiong2020layer}, and GeLU nonlinearity~\citep{hendrycks2016gaussian} with no tying of the embedding and readout parameters. We do not use bias terms for weight parameters or Layernorm, following~\citet{chowdhery2023palm}. Layernorm has a learnable scale parameter. We do not use dropout. All experiments are implemented in Flax~\citep{flax2020github} on top of JAX~\citep{jax2018github} and use Optax optimizers~\citep{deepmind2020jax}. For all optimizers except Adafactor, we use ZeRO3~\citep{rajbhandari2020zero} fully-sharded data parallelism (FSDP). Our FSDP implementation did not work with Adafactor out-of-the-box due to tensor shape mismatches as a result of the factored matrices in Adafactor so we omit it for that optimizer.

All models are trained on the C4 dataset~\citep{t5} encoded with the T5 SentencePiece~\citep{kudo2018sentencepiece} tokenizer, with an additional beginning-of-sequence (BOS) token, resulting in the vocabulary size of $V = 32,001$ ($32,000$ original vocabulary + $1$ BOS).\footnote{Effective vocabulary dimension in experiments is $32,101$ due to $100$ unused tokens.} Training inputs are sequence-packed, while evaluation inputs are padded.

We use a fixed batch size $256$, context length $512$ and depth $L=8$ for all experiments. The different model sizes considered are listed in \cref{tab:model_sizes}. Specifically, we fix the head dimension $h=128$ and co-scale the model dimension $D$, number of heads $H$ and MLP dimension $F$ such that $D = H \times h$ and $F = 4 \times D$ in all models. The resulting number of parameters is approximately $L\times 12 D^2 + 2 V D$, with exact parameter counts reported in \cref{tab:model_sizes}. The compute optimal experiments include models up to $H=32$ or $H=48$, and the fixed ($50{,}000$) step experiments include models up to $H=128$.

For each model size, we sweep the learning rate in increments of $2^{0.25}$ or $2^{0.5}$, with the largest stable learning rate determined by a heuristic: if the learning rate exceeds the optimal learning rate and the eval loss exceeds the minimum eval loss by more than $20\%$ or causes NaNs, we consider the learning rate unstable. We ensured that our learning rate sweeps covered this stability threshold so that the gap between the largest plotted learning rate and smallest unstable learning rate is at most $2^{0.5}$ and in many cases is $2^{0.25}$. The learning rate sweep plots show only the stable learning rates so learning rates larger than the rightmost point in each plot can therefore be considered unstable.

\begin{table}[h]
\centering
\caption{Model sizes used in experiments.}
\resizebox{0.9\columnwidth}{!}{%
\begin{tabular}{@{}n{3}{0}n{5}{0}n{5}{0}n{9}{0}n{11}{0}n{11}{0}@{}}
\toprule
\multicolumn{1}{l}{Number of heads} & \multicolumn{1}{l}{Model dimension} & \multicolumn{1}{l}{MLP width} & \multicolumn{3}{c}{Parameter Counts}                                                        \\
\multicolumn{1}{l}{$H$}            & \multicolumn{1}{l}{$D=128 H$}       & \multicolumn{1}{l}{$F = 4 D$} & \multicolumn{1}{r}{Embedding} & \multicolumn{1}{r}{Non-embedding} & \multicolumn{1}{r}{Total} \\ \midrule
1                                                   & 128                                                  & 512                                            & 4108928                       & 5749504                   & 9858432                   \\
2                                                   & 256                                                  & 1024                                           & 8217856                       & 14644736                  & 22862592                  \\
4                                                   & 512                                                  & 2048                                           & 16435712                      & 41872384                  & 58308096                  \\
6                                                   & 768                                                  & 3072                                           & 24653568                      & 81682944                  & 106336512                 \\
8                                                   & 1024                                                 & 4096                                           & 32871424                      & 134076416                 & 166947840                 \\
12                                                  & 1536                                                 & 6144                                           & 49307136                      & 276612096                 & 325919232                 \\
16                                                  & 2048                                                 & 8192                                           & 65742848                      & 469479424                 & 535222272                 \\
20                                                  & 2560                                                 & 10240                                          & 82178560                      & 712678400                 & 794856960                 \\
24                                                  & 3072                                                 & 12288                                          & 98614272                      & 1006209024                & 1104823296                \\
32                                                  & 4096                                                 & 16384                                          & 131485696                     & 1744265216                & 1875750912                \\
48                                                  & 6144                                                 & 24576                                          & 197228544                     & 3824357376                & 4021585920                \\
64                                                  & 8192                                                 & 32768                                          & 262971392                     & 6709755904                & 6972727296                \\
96                                                  & 12288                                                & 49152                                          & 394457088                     & 14896472064               & 15290929152               \\
128                                                 & 16384                                                & 65536                                          & 525942784                     & 26304413696               & 26830356480       
\\ \bottomrule
\end{tabular}
}
\label{tab:model_sizes}
\end{table}

\subsubsection{Parameterization Details}
This section includes details about the parameterization implementations for our Transformer model. For the purpose of parameterization, the \emph{embedding} layers include the embeddings, positional embeddings and the Layernorm scale parameter, the \emph{hidden} layers include the MLP layers in the Transformer block, the dense query, key and value layers, and the attention output projection layer, and the \emph{readout} layer is just the readout layer.

We use the variant of muP originally proposed by \citet{yang2021tensoriv}, which is also presented in Table 9 of \citet{yang2022tensorv}.

When the embedding initialization is a constant (i.e. has zero as the exponent), we use $0.01$ for the embedding and positional embedding initialization standard deviation. We otherwise omit constant factors from parameterized quantities unless otherwise specified.

The attention operator contains a tensor contraction between the query and key matrices, which induces another question about alignment: if we assume alignment between the query and key, then we should normalize by the head dimension $h$ and if we do not assume alignment then we should normalize by $\sqrt{h}$ inside the softmax. We follow convention and use $\sqrt{h}$ for standard and NTK parameterizations and $h$ for muP and mean-field. However, we note that due to our fixed head dimension that this difference amounts to only a constant factor.

\subsubsection{Optimizer Details}
\label{app:optim_details}
We list the default optimizer hyperparameters for each optimizer in \cref{tab:optim_hyperparameters}. We use these hyperparameters unless otherwise stated, for example in the epsilon experiments. Note that Adam + parameter scaling differs from Adam only by the parameter scaling. Adafactor uses the default optimizer hyperparameter values from the Optax implementation~\citep{deepmind2020jax}.

We do not use weight decay except for in the weight decay experiments in \cref{fig:appendix_adam_weight_decay} which use decoupled (independent) weight decay of 1e-4. The learning rate schedule for all experiments uses linear warmup of $1{,}000$ steps followed by a cosine decay schedule with initial and final learning rates of $0.0$.

\begin{table}[h!]
\centering
\caption{Default optimizer hyperparameters used in all experiments unless otherwise stated.}
\label{tab:optim_hyperparameters}
\begin{tabular}{lcccc} %
\toprule
\textbf{Hyperparameter} & \textbf{SGD} & \textbf{Adam} & \textbf{Adam + PS} & \textbf{Adafactor} \\ 
\midrule
Momentum / Beta1 (first moment exponential decay) & 0.9 & 0.9 & 0.9 &  1.0 \\ 
Beta2 (second moment exponential decay) &  & 0.98 & 0.98 & 0.8 \\ 
Epsilon &  & 1e-9 & 1e-9 & 1e-30 \\ 
Parameter Scaling &  & False & True & True \\ 
Factored Gradient RMS &  & False & False & True \\ 
Update Clipping (by block RMS) &  & None & None & 1.0 \\ 
\bottomrule
\end{tabular}
\end{table}
\clearpage
\subsubsection{Hyperparameter tuning for constant per-layer learning rate factors}
\label{app:tune_constant_factors}
When tuning the per-layer constant multiplicative factors defined in \cref{sec:results_per_layer}, we use a Bayesian optimization library~\citep{google_vizier} to perform a three-dimensional hyperparameter search for $(\gamma_1, \gamma_h, \gamma_{L+1})$ at the base model dim $b=1024$. Recall that we define the learning rate in layer $l$ as $\eta_l = \beta_n \cdot \gamma_l \cdot \frac{n}{b} ^ {-c_l}$ and sweep one dimension at all model sizes to determine $\beta_n$, so these values of $(\gamma_1, \gamma_h, \gamma_{L+1})$ define two ratios where any common factor can be absorbed by $\beta_n$.

For each optimizer $\times$ parameterization, we run 800 trials with at most 100 trials in parallel with a range set to $[1e-2, 1e2]$ for each constant. If the optimal value for any of the constants is at or near the edge of the range after this first search, we extend the range of the sweep for that constant to 0.01 and 100x the optimal value found in the original sweep and repeat the same tuning procedure.

Since the eval loss has some noise, we consider all trials that perform within $0.1\%$ relative eval loss of the best trial to be equivalently good, and determine the optimal constants using the average for each constant over this set of best trials.
\vspace{8pt}
\begin{table}[h]
\caption{Optimal per-layer learning rate multipliers found via three-dimensional hyperparameter search with Bayesian optimization for each optimizer and parameterization at the base model dim $b=1024$.}
\vspace{6pt}
\label{tab:optimal_per_layer_multipliers}
\centering
\adjustbox{scale=0.9}{
\begin{tabular}{lllll}
\toprule
\textbf{Optimizer} & \textbf{Parameterization} & \textbf{Embedding} & \textbf{Hidden} & \textbf{Readout} \\
\midrule
\multirow{4}{*}{SGD} &  STP  & 10.426 & 5.404 & 0.092 \\
 & NTK & 0.034 & 53.176 & 0.232 \\
 & muP & 1398.861 & 5.532 & 0.020 \\
 & MFP & 1.325 & 6.506 & 0.504 \\
\midrule
\multirow{4}{*}{Adam} & STP & 5.357 & 1.133 & 2.596 \\
 & NTK & 0.144 & 0.886 & 2.095 \\
 & muP & 5.303 & 1.258 & 11.723 \\
 & MFP & 0.351 & 0.939 & 2.814 \\
\midrule
\multirow{4}{*}{Adam+PS} &  STP  & 2.598 & 1.845 & 0.838 \\
 & NTK & 1.224 & 0.881 & 0.591 \\
 & muP & 0.503 & 0.794 & 0.914 \\
 & MFP & 2.339 & 1.060 & 0.516 \\
\midrule
 \multirow{4}{*}{Adafactor} & STP  & 0.918 & 1.048 & 0.703 \\
 & NTK & 0.837 & 0.955 & 0.765 \\
 & muP & 0.784 & 1.170 & 0.668 \\
 & MFP & 1.879 & 0.928 & 0.646 \\
\bottomrule 
\end{tabular}
}
\end{table}
\clearpage

\subsubsection{Adam-atan2 Code Change}
\label{sec:adam_atan_code}

To implement \emph{Adam-atan2} using the jax numpy package, imported here as \Verb +jnp+, we change a single line of code to replace the default Adam update rule:
\begin{center}
\begin{BVerbatim}[vspace=0pt,baselinestretch=0.0]
lambda m, v: m / (jnp.sqrt(v + eps_root) + eps)
\end{BVerbatim}
\end{center}
with the \emph{Adam-atan2} update rule:
\begin{center}
\begin{BVerbatim}[vspace=0pt,baselinestretch=0.0]
lambda m, v: a * jnp.arctan2(m, b * jnp.sqrt(v))
\end{BVerbatim}
\end{center}
where $a$ and $b$ are constants. We use $a = b = 1$ for all experiments, but other values might be considered in future work as discussed below.

The $\atan2(x,y)$ function is a standard library function that is typically a thin wrapper~\citep{numpy_atan2} around the arctangent function that determines the appropriate quadrant, handles zero / NaN / infinity values, and otherwise returns $\arctan(x/y)$. Recall the small-angle approximation $\tan \theta \approx \theta$ that applies for the inverse function $\arctan \theta \approx \theta$ as well. Therefore, for values of $x/y$ close to zero, $\atan2(x, y) = \arctan(x/y) \approx x/y$ which approximates the usual division operation in Adam.

For values away from zero, the arctangent function smoothly approaches $\pm\,\pi/2$ as the argument goes to $\pm\,\infty$. In particular, when the second-order moment estimate is out-of-date due to slow decay of the moving average, we may have $m > \sqrt{v}$ resulting in an argument with large magnitude that would be smoothly clipped by the arctangent function. This likely results in an effect similar to the update clipping used in Adafactor~\citep{shazeer2018adafactor}. However, unlike an additive constant epsilon, $\atan2$ is scale-invariant up to precision limits in that $\Adam-atan2(\lambda g) = \atan2(\lambda m, \lambda \sqrt{v}) = \atan2(m, \sqrt{v}) = \Adam-atan2(g)$ whereas $\Adamoptim(\lambda g) = \frac{\lambda m}{\lambda \sqrt{v} + \epsilon} \neq \frac{m}{\sqrt{v} + \epsilon} = \Adamoptim(g)$ where $g$ represents the gradients and $\lambda$ is a constant.

\begin{figure*}[h!]
\centering
\adjustbox{scale=0.95}{
    \begin{minipage}[b]{0.48\linewidth} %
        \centering
        \includegraphics[width=\linewidth, trim={0, 0, 0, 0},clip]{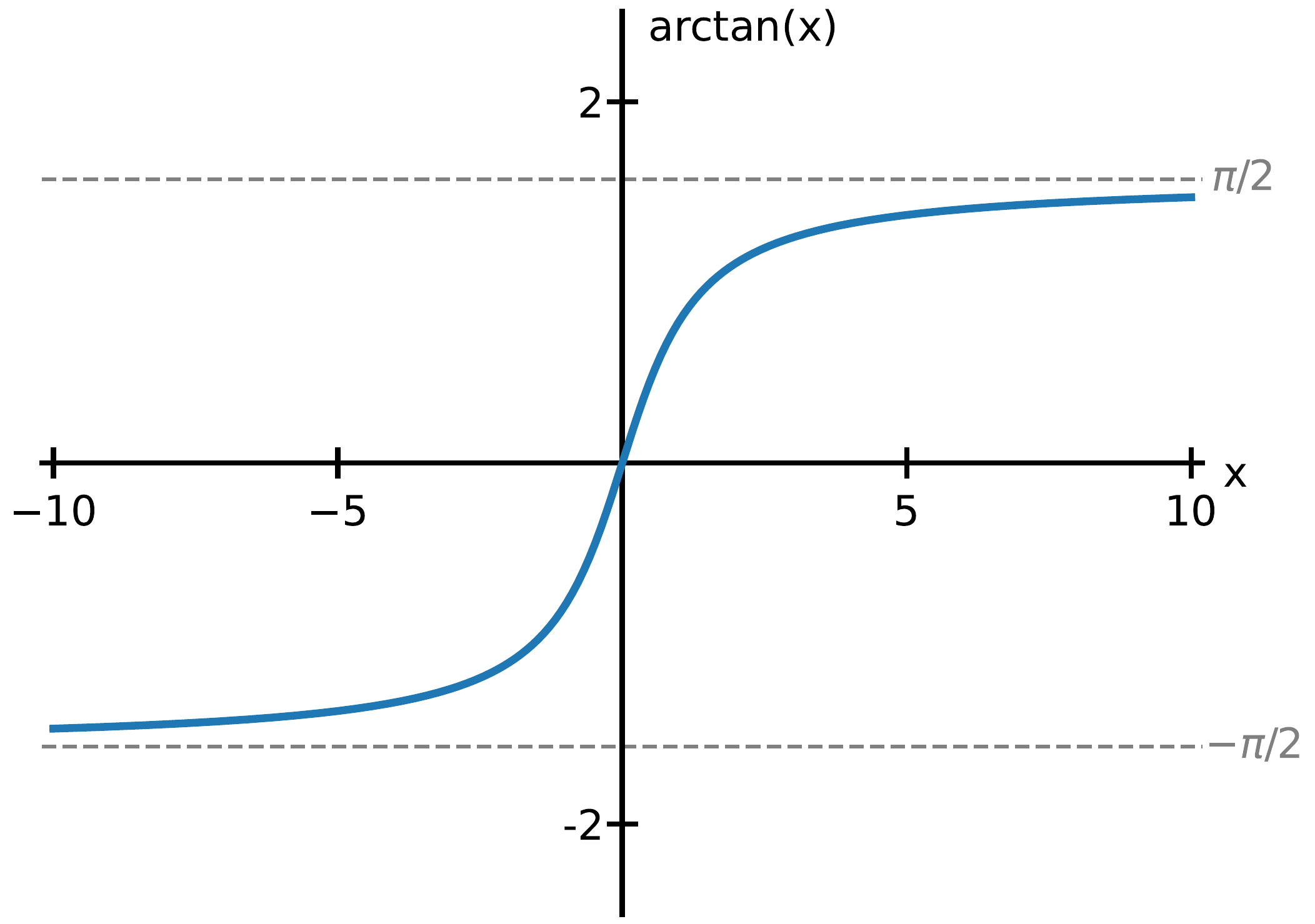}
        \caption{Arctangent function}
        \label{fig:appendix_arctan_function}
    \end{minipage}
    \hfill %
    \begin{minipage}[b]{0.48\linewidth} %
        \centering
        \begingroup
        \renewcommand{\arraystretch}{1.7}
        \begin{tabular}{|c|c|c|}
        \hline
         $b$ & $a$, if $m \ll \sqrt{v}$ & $a$, if $m \sim \sqrt{v}$ \\ \hline
         1 & 1 & $1/\arctan(1) = 1.27$ \\ \hline
         2 & 2 & $1/\arctan(\frac{1}{2}) = 2.16$ \\ \hline
         4 & 4 & $1/\arctan(\frac{1}{4}) = 4.08$ \\ \hline
         8 & 8 & $1/\arctan(\frac{1}{8}) = 8.04$ \\ \hline
         16 & 16 & $1/\arctan(\frac{1}{16}) = 16.02$ \\ \hline
         32 & 32 & $1/\arctan(\frac{1}{32}) = 32.01$ \\ \hline 
        \end{tabular}
        \vspace{12pt}
        \captionof{table}{Constants $a$ and $b$ for scaling arctangent.}
        \label{tab:arctan_constants}
        \endgroup
    \end{minipage}
    }
\end{figure*}

Regarding the constant values $a$ and $b$, using a value of $b > 1$ rescales the argument to be closer to zero which extends the region where arctangent acts as a small-angle approximation. For a given $b$, the choice of $a$ controls the approximation when the argument is far from zero. We note, however, that changing the value of $a$ simply rescales the effective learning rate and would be absorbed into a sweep of the base learning rate.

Depending on the relationship between $m$ and $v$, we derive different values of $a$ that would preserve the effective learning rate for Adam for a particular value of $b$. Recall that $m$ and $v$ are the first- and second-order moment estimates of the gradient, so when the moving average is up-to-date then $\norm{m} \leq \norm{\sqrt{v}}$ and the arctangent argument will be in the range $[-1/b, 1/b]$. If $m << \sqrt{v}$, to give an accurate small-angle approximation, we want $a=b$ so that the first term in the Taylor series $a \cdot \arctan(x, b\cdot y) = \frac{a}{by} x + \ldots$ matches the usual division operator that is linear in $x$ with coefficient $1/y$. If $m \sim \sqrt{v}$, to give an accurate approximation when $m / \sqrt{v}$ approaches $\pm 1$, we want $a = \frac{1}{arctan(1/b)}$ so that $a \cdot \arctan(m, b \cdot \sqrt{v}) = a \cdot arctan(1/b) = 1$. For $b=1$, this corresponds to $a = 4 / \pi \approx 1.27$. It is therefore possible that our choice of $a=1$ when $b=1$ may induce a change of up to this $\approx 1.27$ factor in the effective learning rate, but this change would be absorbed into our learning rate sweeps. We note in \cref{tab:arctan_constants} that as $b$ becomes larger, the values from $a$ converge between these two regimes.

\newcommand{\figwidth}{0.2\paperwidth}
\newcommand{\figvspace}{\vspace{0.5cm}}
\newcommand{\appsinglefigure}[1]{\includegraphics[width=\figwidth, trim={0, 0, 0, 0},clip]{#1}}

\newcommand{\appfigure}[2]{
    \centerline{
        \appsinglefigure{icml2024/appendix_figures/lr_sweep/#1+stp+50k_steps#2.pdf}
        \appsinglefigure{icml2024/appendix_figures/power_law/#1+stp+50k_steps#2.pdf}
        \appsinglefigure{icml2024/appendix_figures/lr_sweep/#1+stp+compute_opt#2.pdf}
        \appsinglefigure{icml2024/appendix_figures/power_law/#1+stp+compute_opt#2.pdf}
    }
    \centerline{
        \appsinglefigure{icml2024/appendix_figures/lr_sweep/#1+ntk+50k_steps#2.pdf}
        \appsinglefigure{icml2024/appendix_figures/power_law/#1+ntk+50k_steps#2.pdf}
        \appsinglefigure{icml2024/appendix_figures/lr_sweep/#1+ntk+compute_opt#2.pdf}
        \appsinglefigure{icml2024/appendix_figures/power_law/#1+ntk+compute_opt#2.pdf}
    }
    \centerline{
        \appsinglefigure{icml2024/appendix_figures/lr_sweep/#1+mup_table_9+50k_steps#2.pdf}
        \appsinglefigure{icml2024/appendix_figures/power_law/#1+mup_table_9+50k_steps#2.pdf}
        \appsinglefigure{icml2024/appendix_figures/lr_sweep/#1+mup_table_9+compute_opt#2.pdf}
        \appsinglefigure{icml2024/appendix_figures/power_law/#1+mup_table_9+compute_opt#2.pdf}
    }
    \centerline{
        \appsinglefigure{icml2024/appendix_figures/lr_sweep/#1+mfp+50k_steps#2.pdf}
        \appsinglefigure{icml2024/appendix_figures/power_law/#1+mfp+50k_steps#2.pdf}
        \appsinglefigure{icml2024/appendix_figures/lr_sweep/#1+mfp+compute_opt#2.pdf}
        \appsinglefigure{icml2024/appendix_figures/power_law/#1+mfp+compute_opt#2.pdf}
    }
}

\clearpage
\begin{figure*}[ht]
\subsection{Additional Alignment Experiments}
\label{app:alignment_expts}
\figvspace
\figvspace
    \begin{center}
        \textbf{SGD+Momentum Alignment Experiments}\\
        \figvspace
        
        \includegraphics[width=\linewidth, trim={0, 0, 0, 0},clip]{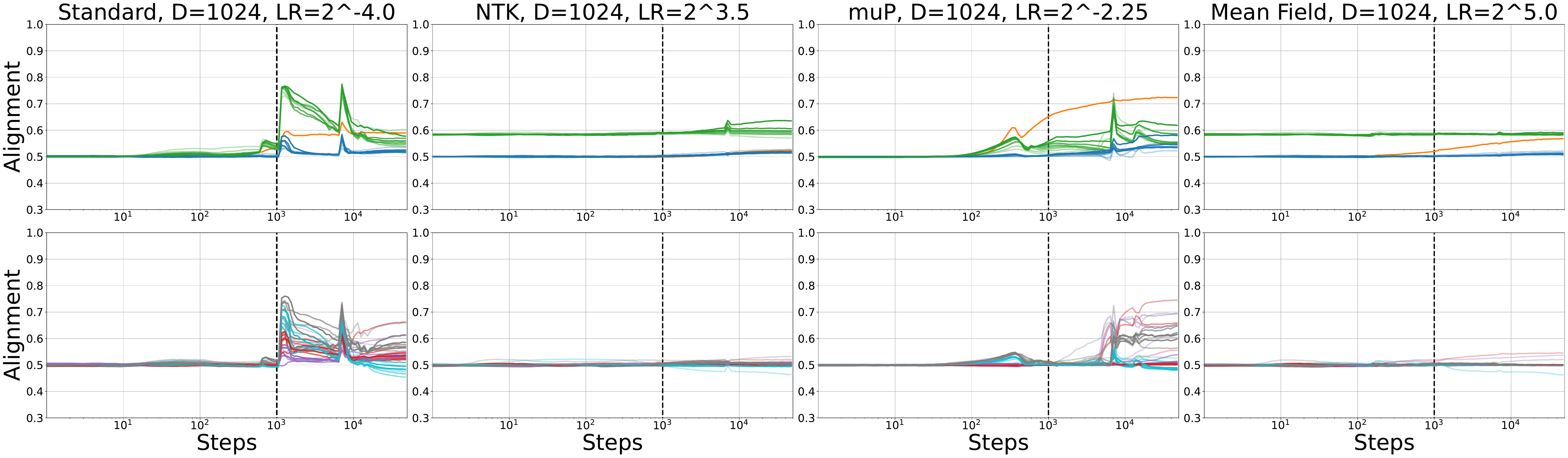}
       
        \figvspace
        \figvspace
       
        \includegraphics[width=\linewidth, trim={0, 0, 0, 0},clip]{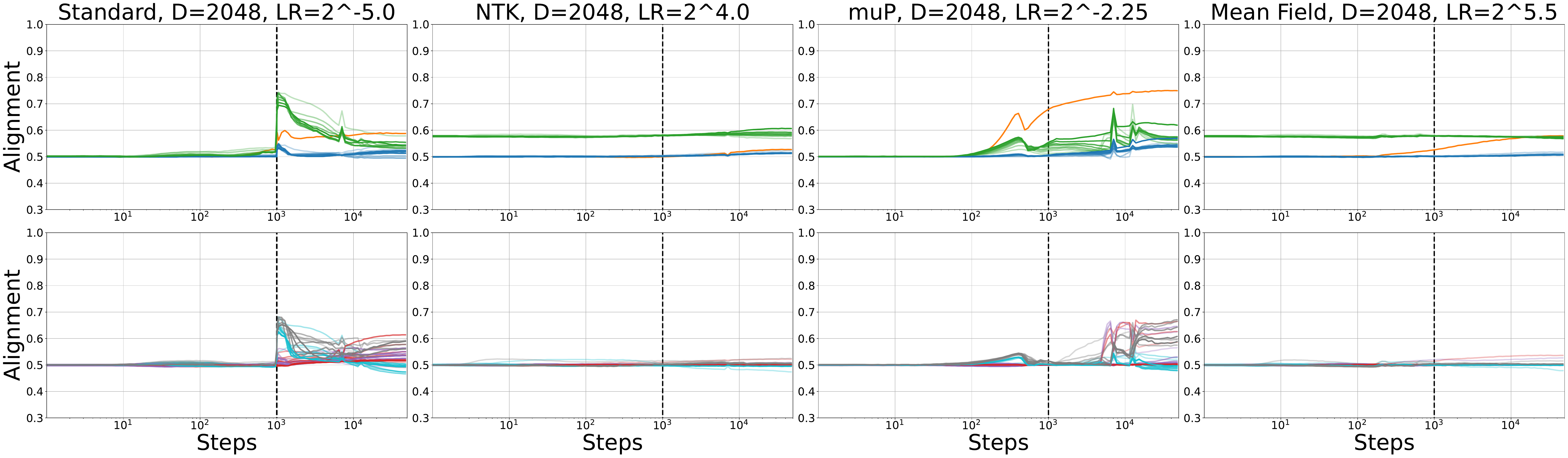}
       
        \figvspace
        \figvspace
       
        \includegraphics[width=\linewidth, trim={0, 0, 0, 0},clip]{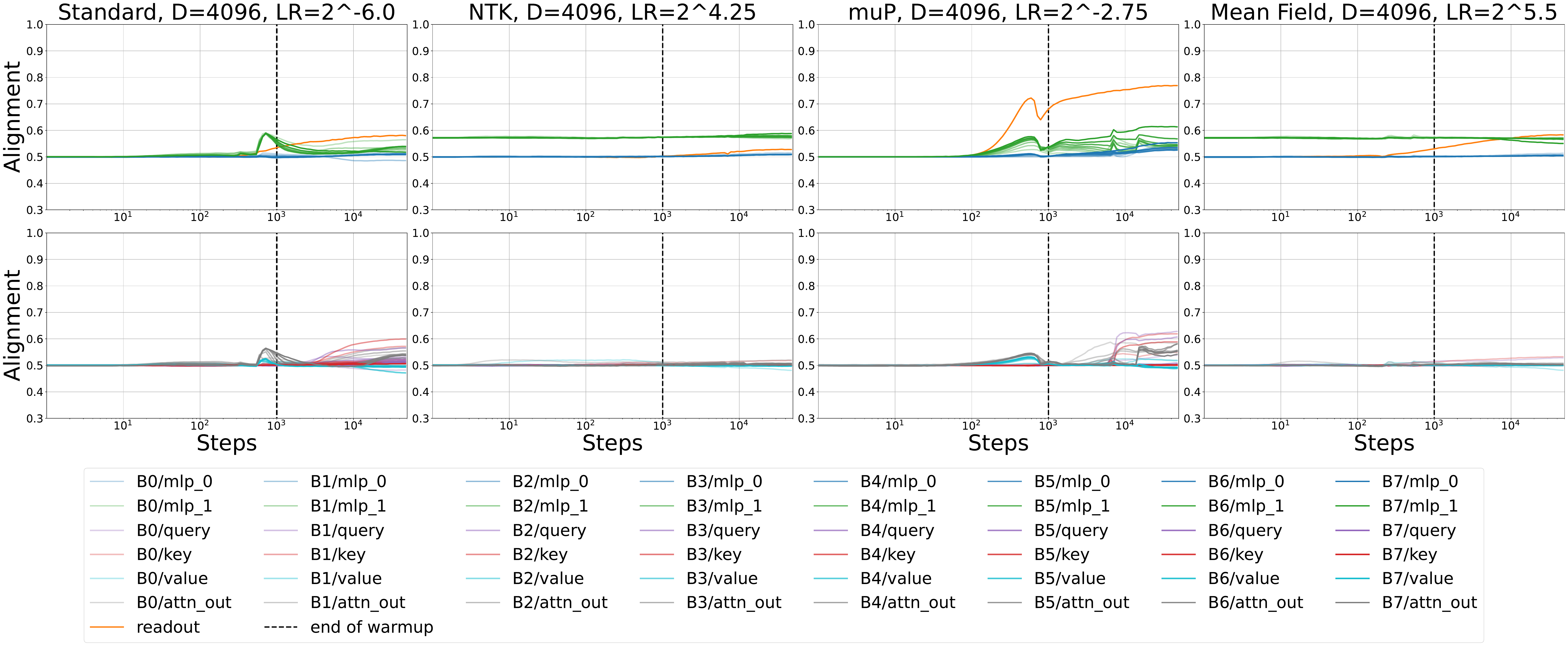}
        \caption{The log alignment ratio measured in all dense layers across training steps for SGD using a global learning rate at approximately the optimal learning rate for each setting. Top = $167M$ parameter model $(D=1024)$, middle = $535M$ parameter model $(D=2048)$, bottom = $1.9B$ parameter model $(D=4096)$. Note the log scale on the x-axis.}
        \figvspace
        \label{fig:appendix_alignment_sgd}
    \end{center}
\end{figure*}
\clearpage

\begin{figure*}[ht]
    \begin{center}
        \textbf{Adam Alignment Experiments}\\
        \figvspace
        
        \includegraphics[width=\linewidth, trim={0, 0, 0, 0},clip]{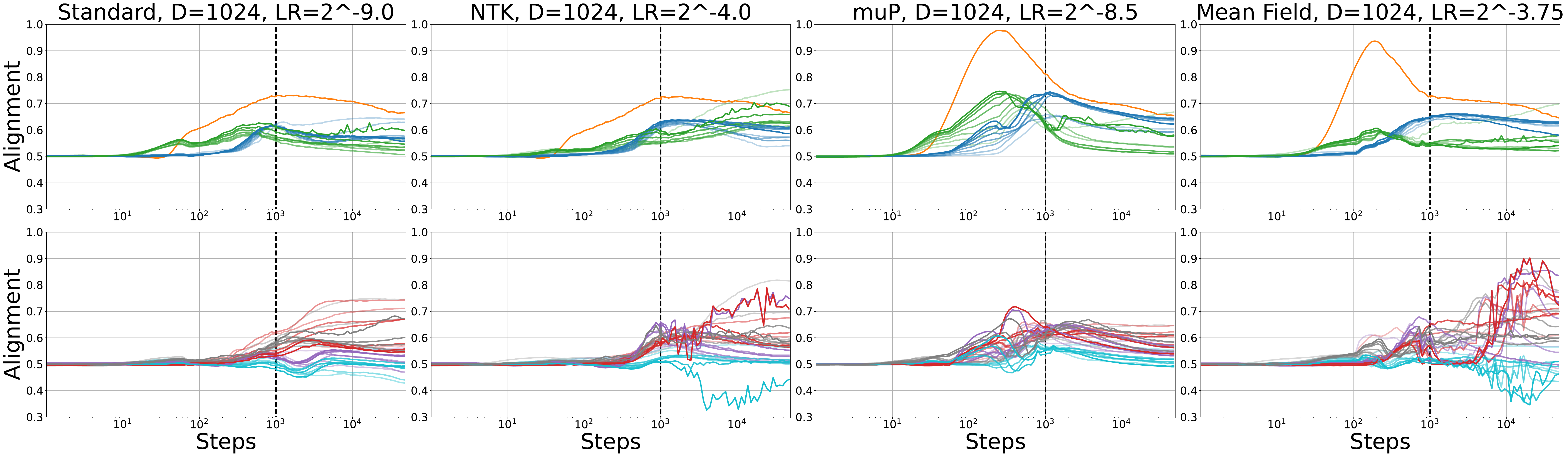}
       
        \figvspace
        \figvspace
       
        \includegraphics[width=\linewidth, trim={0, 0, 0, 0},clip]{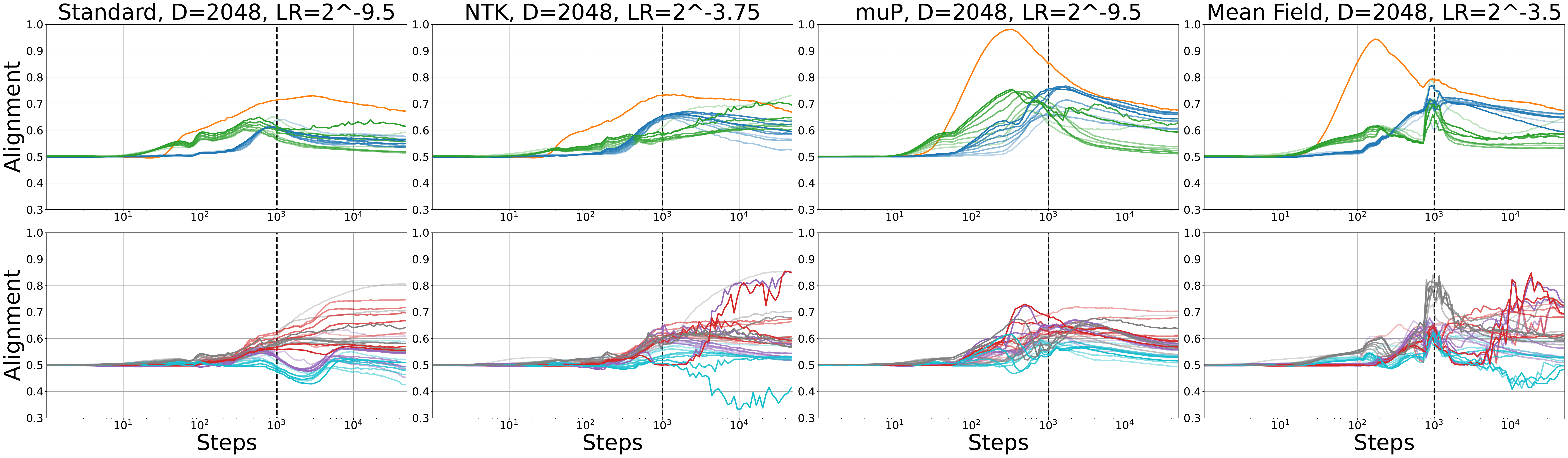}
       
        \figvspace
        \figvspace
       
        \includegraphics[width=\linewidth, trim={0, 0, 0, 0},clip]{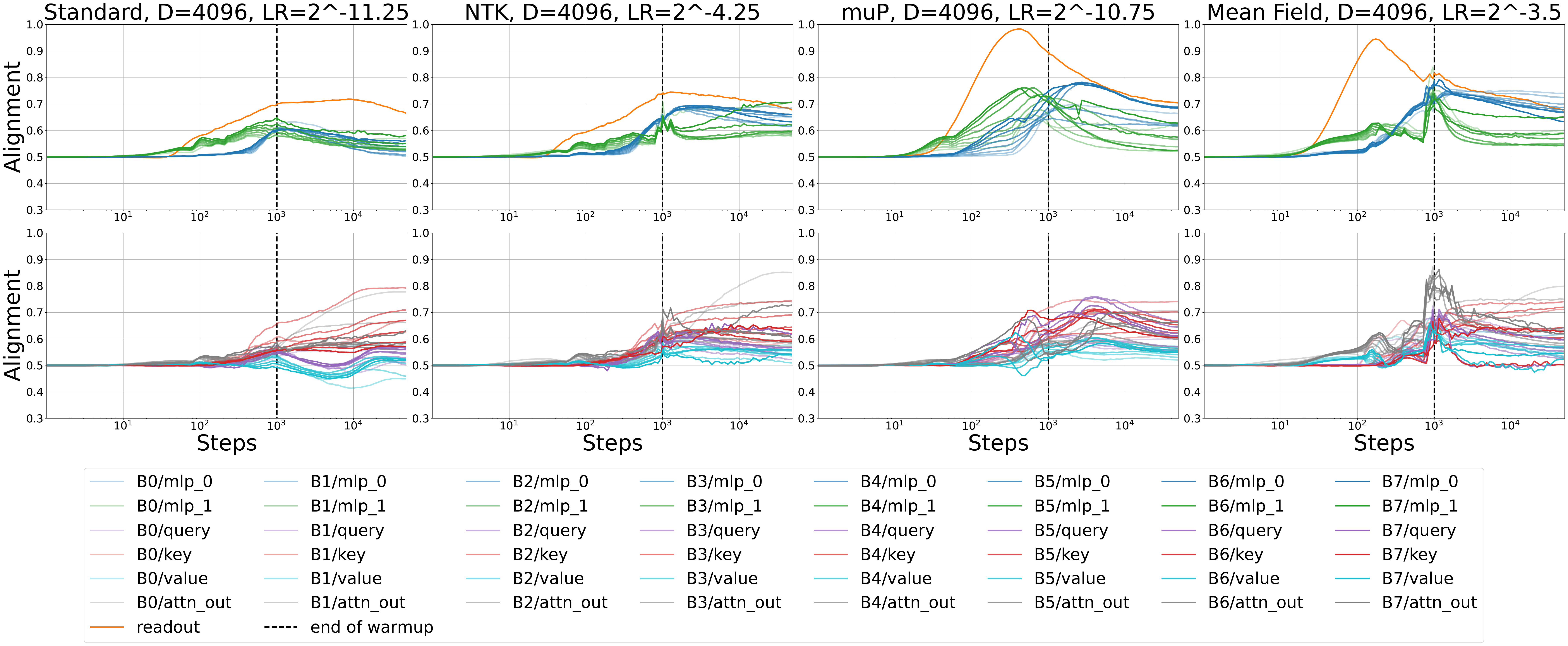}
        \caption{The log alignment ratio measured in all dense layers across training steps for Adam using a global learning rate at approximately the optimal learning rate for each setting. Top = $167M$ parameter model $(D=1024)$, middle = $535M$ parameter model $(D=2048)$, bottom = $1.9B$ parameter model $(D=4096)$. Note the log scale on the x-axis.}
        \label{fig:appendix_alignment_adam}
    \end{center}
\end{figure*}
\newpage

\begin{figure*}[ht]
    \begin{center}
        \textbf{Adafactor Alignment Experiments}\\
        \figvspace
        
        \includegraphics[width=\linewidth, trim={0, 0, 0, 0},clip]{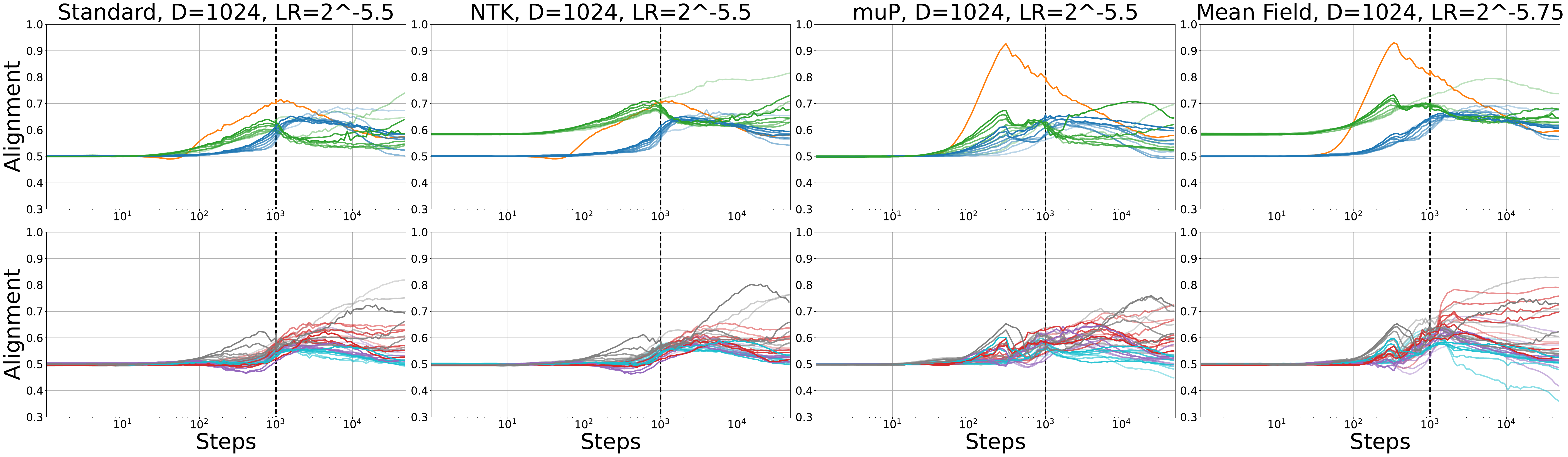}
       
        \figvspace
        \figvspace
       
        \includegraphics[width=\linewidth, trim={0, 0, 0, 0},clip]{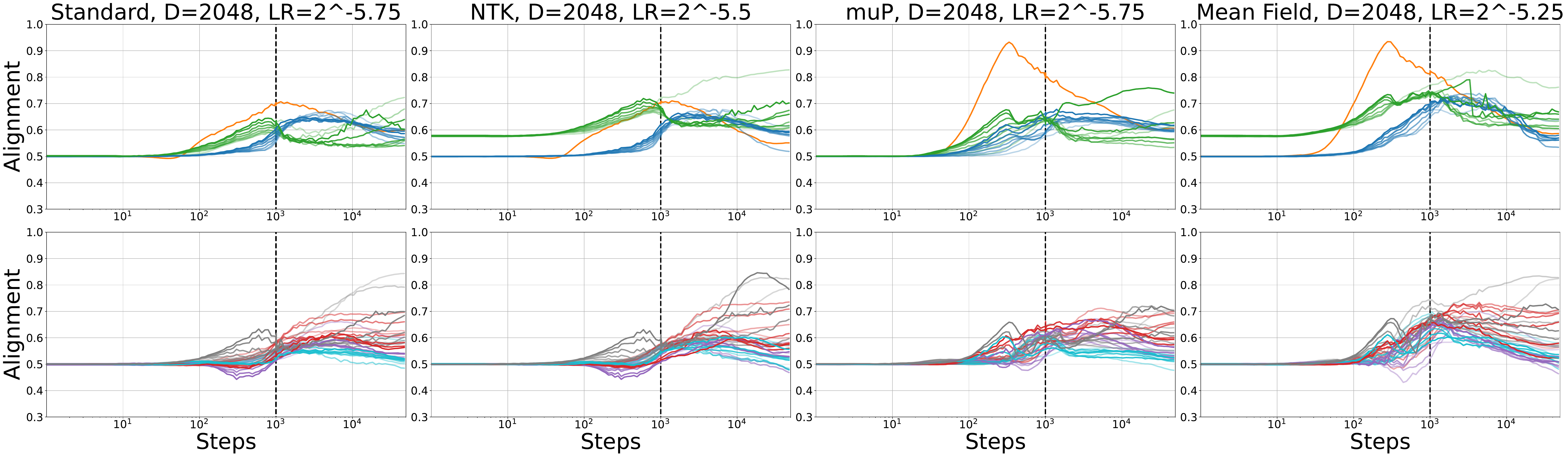}
       
        \figvspace
        \figvspace
       
        \includegraphics[width=\linewidth, trim={0, 0, 0, 0},clip]{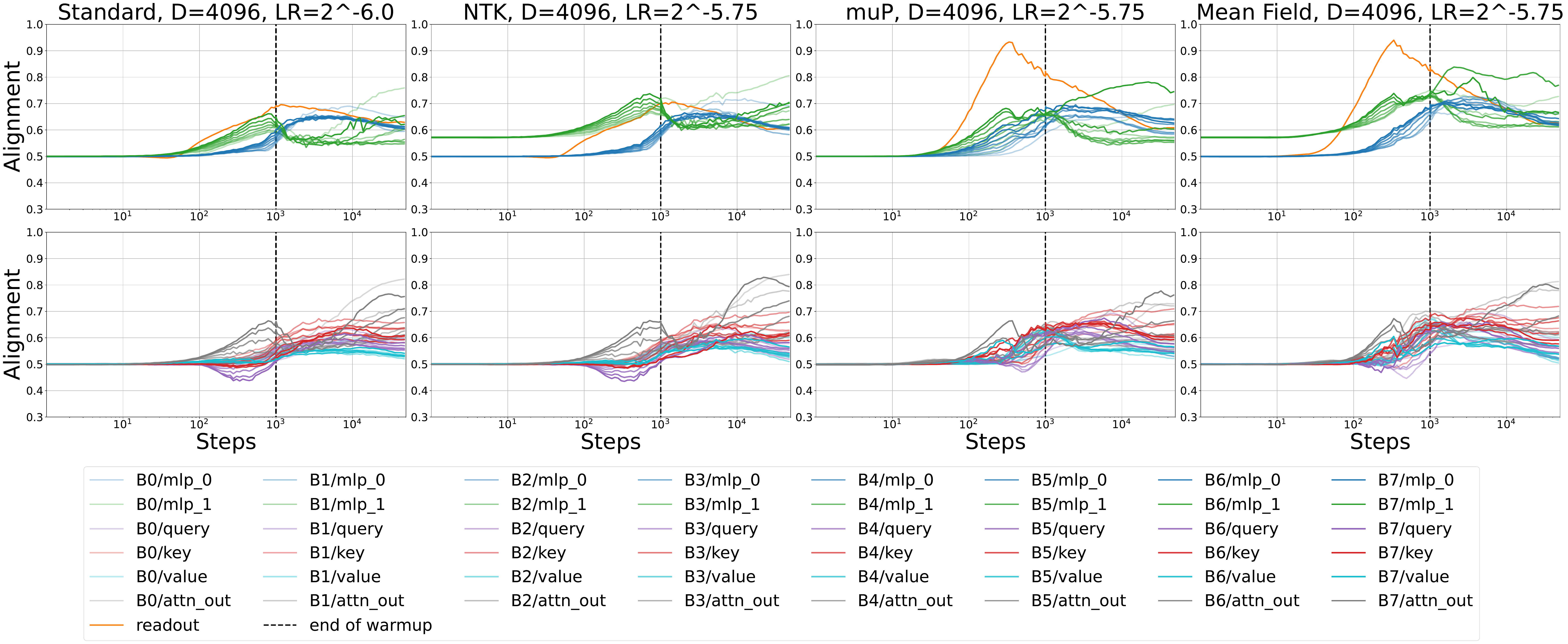}
        \caption{The log alignment ratio measured in all dense layers across training steps for Adafactor using a global learning rate at approximately the optimal learning rate for each setting. Top = $167M$ parameter model $(D=1024)$, middle = $535M$ parameter model $(D=2048)$, bottom = $1.9B$ parameter model $(D=4096)$. Note the log scale on the x-axis.}
        \label{fig:appendix_alignment_adafactor}
    \end{center}
\end{figure*}
\clearpage

\subsection{Additional Per-Layer Learning Rate Experiment Results}
\label{sec:app_per_layer_lr_results}
\begin{figure}[h]
\includegraphics[width=\linewidth]{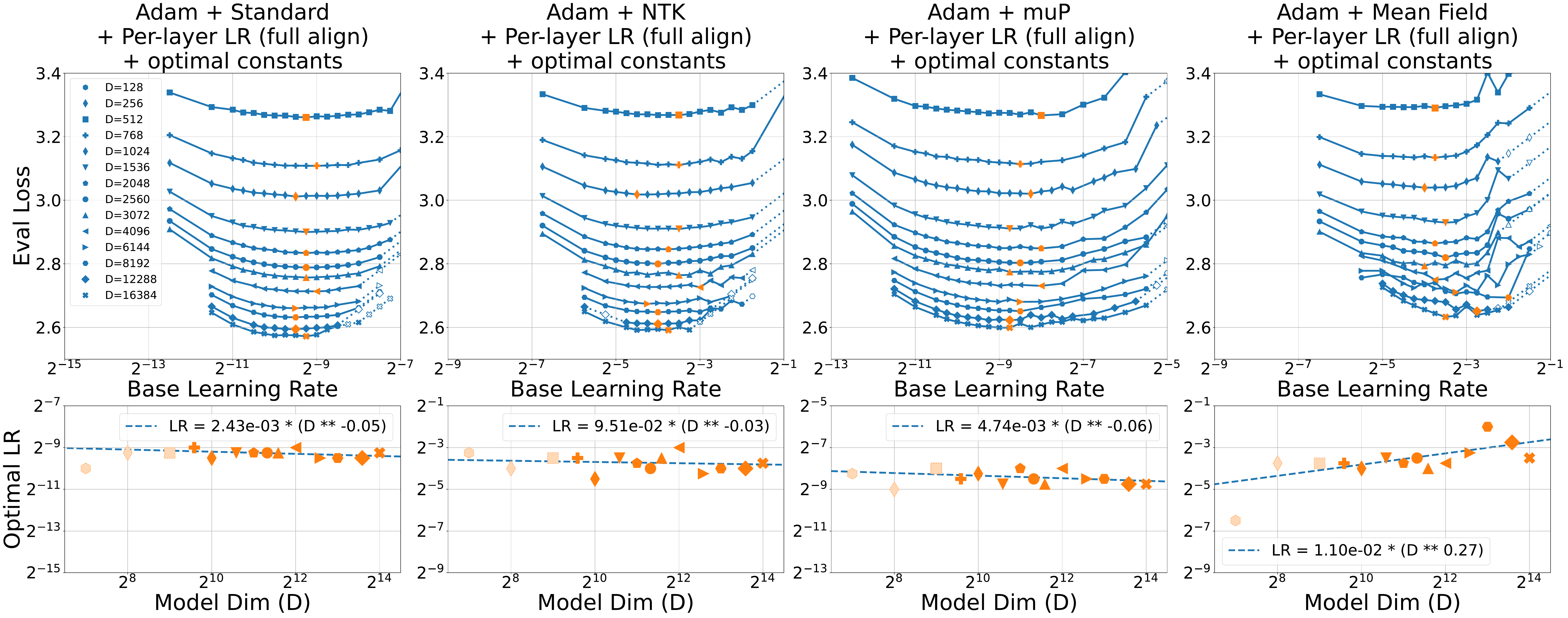}

\figvspace

\includegraphics[width=\linewidth]{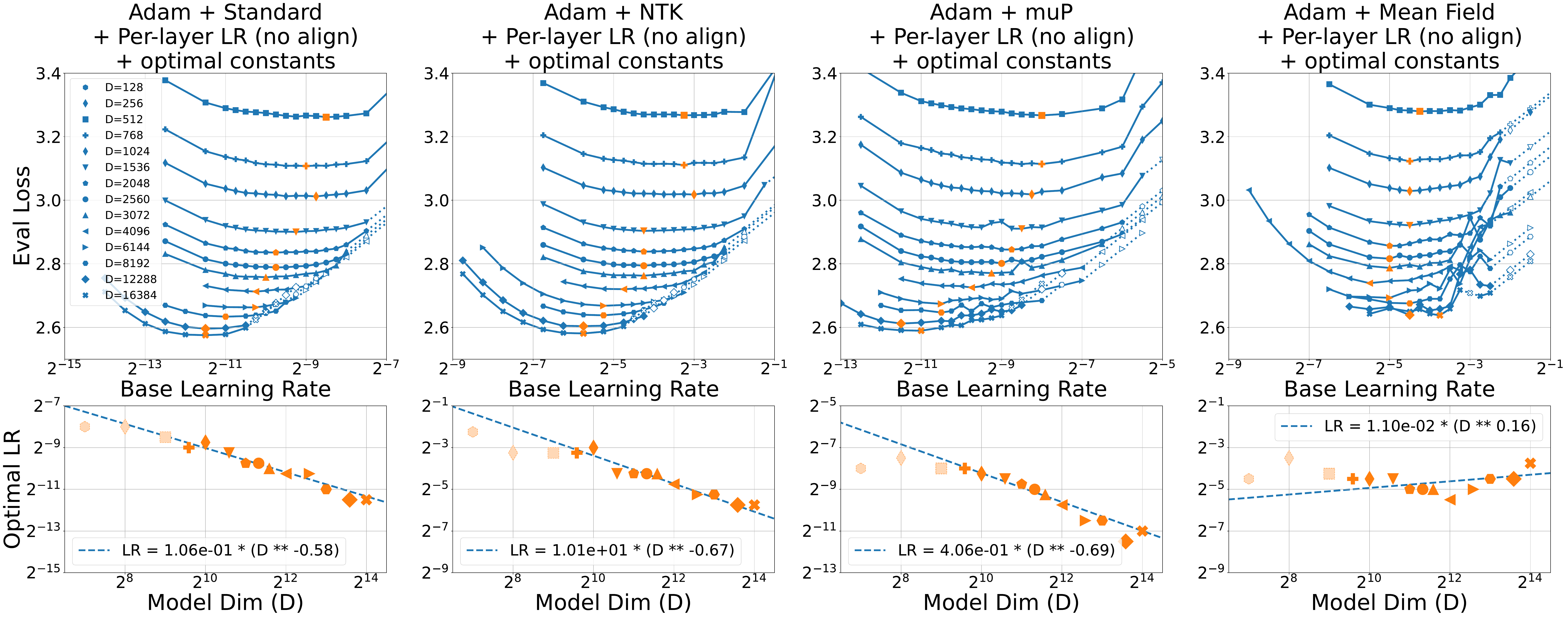}
\caption{\textbf{Despite slight improvements in the eval loss under no alignment assumptions for NTK, muP and MFP, the full alignment experiments show better scale-invariance of the optimal learning rate.} Learning rate sweeps and power laws fit to optimal learning rate vs model dim. Top = Adam + per-layer learning rates assuming full alignment + optimal constants. Bottom = Adam + per-layer learning rates assuming no alignment + optimal constants. Number of training steps = $50{,}000$.}
\label{fig:app_hparam_transfer_align_comparison}
\end{figure}

This section includes additional results for the per-layer learning rate experiments in \cref{sec:results_per_layer}. In \cref{fig:app_hparam_transfer_align_comparison}, we show the difference in the scale-dependence of the optimal learning rate for the full alignment vs no alignment settings for Adam. In \cref{tab:appendix_table} we report the eval losses for the six largest model sizes in all settings including per-layer epsilon settings. In \cref{fig:app_scaling_optimal_constants}, we show eval loss vs model size scaling curves for all optimizers for all settings that use optimal per-layer learning rate constant multipliers. In \cref{fig:app_scaling_ablation}, we include eval loss vs model size scaling curves for an ablation of global vs per-layer learning rates and default vs optimal constants. Learning rate sweeps for all settings are included in \sref{sec:app_lr_sweeps_sgd}, \sref{sec:app_lr_sweeps_adam} and \sref{sec:app_lr_sweeps_adam_ps}.

\vfill
\vfill
\clearpage

\begingroup
\renewcommand{\arraystretch}{1}
\begin{table*}[h!]
\centering
\caption{\textbf{Best eval losses for six largest model sizes.} For each optimizer $\times$ parameterization $\times$ model size $\times$ setting, we sweep the base LR and report the best eval loss. For per-layer epsilon experiments (rightmost two columns), we use base epsilon = 1e-12.}
\label{tab:appendix_table}
\adjustbox{scale=0.6}{
\begin{tabularx}{1.65\textwidth}{XXX | XXXXXXX}
\toprule[1.5\heavyrulewidth]
& & & Global LR \newline+ default & Global LR \newline+ optimal & Per-layer LR \newline+ full align \newline+ default & Per-layer LR \newline+ full align \newline+ optimal & Per-layer LR \newline+ no align \newline+ optimal & Per-layer LR \newline+ perf align \newline+ optimal \newline+ per-layer eps & Per-layer LR \newline+ no align \newline+ optimal \newline+ per-layer eps\\
\midrule[\heavyrulewidth]
\multirow{24}{*}{\Large SGD} & \multirow{6}{*}{\Large STP } & 1.1B & 3.464 & 3.256 & 3.810 & \textbf{3.122} & 3.252 &  &  \\ &  & 1.9B & 3.749 & \textbf{3.178} & 3.543 & 3.345 & 3.302 &  &  \\ &  & 4B & \textbf{3.220} & \textbf{3.220} & 3.628 & 3.264 & 3.294 &  &  \\ &  & 7B & 3.414 & 3.214 & 3.601 & 3.124 & \textbf{3.089} &  &  \\ &  & 15.3B & 3.172 & 3.074 & 3.386 & \textbf{3.037} & 3.377 &  &  \\ &  & 26.8B & 3.657 & \textbf{3.312} & 3.510 & 3.385 & 3.318 &  &  \\  \cmidrule{2-10}  & \multirow{6}{*}{\Large NTK } & 1.1B & 4.029 & \textbf{3.087} & 4.134 & 3.305 & 3.160 &  &  \\ &  & 1.9B & 4.029 & 3.385 & 4.029 & 3.374 & \textbf{3.325} &  &  \\ &  & 4B & 3.726 & 3.393 & 3.791 & \textbf{3.247} & 3.412 &  &  \\ &  & 7B & 3.794 & 3.134 & 3.704 & \textbf{3.076} & 3.087 &  &  \\ &  & 15.3B & 3.718 & 3.320 & 3.572 & \textbf{3.028} & 3.194 &  &  \\ &  & 26.8B & 3.732 & \textbf{3.210} & 3.627 & 3.313 & 3.324 &  &  \\  \cmidrule{2-10}  & \multirow{6}{*}{\Large muP } & 1.1B & 4.029 & 3.188 & 3.973 & 3.188 & \textbf{3.074} &  &  \\ &  & 1.9B & 3.883 & 3.759 & 3.883 & 3.759 & \textbf{3.050} &  &  \\ &  & 4B & 3.852 & 3.666 & 3.796 & 3.666 & \textbf{3.627} &  &  \\ &  & 7B & 3.464 & \textbf{3.098} & 3.532 & \textbf{3.098} & 3.828 &  &  \\ &  & 15.3B & 3.357 & 3.252 & 3.430 & 3.578 & \textbf{3.166} &  &  \\ &  & 26.8B & 4.224 & 3.810 & 4.184 & \textbf{3.809} & 4.222 &  &  \\  \cmidrule{2-10}  & \multirow{6}{*}{\Large MFP } & 1.1B & \textbf{3.805} & 4.010 & 4.255 & 4.082 & 4.095 &  &  \\ &  & 1.9B & 4.217 & 4.057 & 4.378 & 4.132 & \textbf{4.048} &  &  \\ &  & 4B & 4.131 & 3.795 & 3.939 & 3.946 & \textbf{3.782} &  &  \\ &  & 7B & 3.874 & 3.825 & 4.034 & 3.820 & \textbf{3.700} &  &  \\ &  & 15.3B & 3.968 & 3.910 & 4.131 & 3.945 & \textbf{3.742} &  &  \\ &  & 26.8B & 4.131 & 4.092 & 4.319 & 4.092 & \textbf{3.898} &  &  \\  \midrule[\heavyrulewidth] \multirow{24}{*}{\Large Adam} & \multirow{6}{*}{\Large STP } & 1.1B & 2.776 & 2.760 & 2.766 & \textbf{2.757} & 2.758 & \textbf{2.757} & 2.758 \\ &  & 1.9B & 2.734 & 2.715 & 2.717 & \textbf{2.713} & \textbf{2.713} & 2.714 & \textbf{2.713} \\ &  & 4B & 2.688 & 2.667 & 2.666 & \textbf{2.660} & 2.663 & 2.661 & 2.663 \\ &  & 7B & 2.665 & 2.641 & 2.636 & \textbf{2.632} & 2.634 & \textbf{2.632} & 2.633 \\ &  & 15.3B & 2.638 & 2.608 & 2.598 & 2.594 & 2.596 & \textbf{2.593} & 2.596 \\ &  & 26.8B & 2.625 & 2.590 & 2.576 & \textbf{2.572} & 2.575 & 2.573 & 2.576 \\  \cmidrule{2-10}  & \multirow{6}{*}{\Large NTK } & 1.1B & 2.782 & \textbf{2.761} & 2.789 & 2.764 & 2.763 & \textbf{2.761} & 2.763 \\ &  & 1.9B & 2.736 & 2.726 & 2.743 & 2.726 & 2.720 & \textbf{2.717} & 2.719 \\ &  & 4B & 2.672 & 2.674 & 2.695 & 2.674 & 2.668 & \textbf{2.665} & 2.666 \\ &  & 7B & 2.647 & 2.643 & 2.666 & 2.648 & 2.638 & \textbf{2.634} & 2.636 \\ &  & 15.3B & 2.605 & 2.605 & 2.634 & 2.611 & 2.604 & 2.600 & \textbf{2.597} \\ &  & 26.8B & 2.592 & 2.584 & 2.615 & 2.591 & 2.581 & \textbf{2.576} & 2.577 \\  \cmidrule{2-10}  & \multirow{6}{*}{\Large muP } & 1.1B & 2.822 & 2.773 & 2.802 & 2.774 & 2.771 & 2.773 & \textbf{2.770} \\ &  & 1.9B & 2.799 & 2.730 & 2.755 & 2.731 & \textbf{2.725} & 2.728 & 2.727 \\ &  & 4B & 2.756 & 2.682 & 2.714 & 2.680 & \textbf{2.674} & 2.680 & 2.679 \\ &  & 7B & 2.738 & 2.656 & 2.692 & 2.651 & 2.647 & 2.652 & \textbf{2.646} \\ &  & 15.3B & 2.729 & 2.628 & 2.664 & 2.623 & 2.612 & 2.623 & \textbf{2.611} \\ &  & 26.8B & 2.727 & 2.614 & 2.646 & 2.599 & \textbf{2.590} & 2.601 & \textbf{2.590} \\  \cmidrule{2-10}  & \multirow{6}{*}{\Large MFP } & 1.1B & 2.816 & 2.798 & 2.807 & 2.793 & 2.787 & \textbf{2.778} & 2.780 \\ &  & 1.9B & 2.770 & 2.756 & 2.756 & 2.750 & 2.739 & \textbf{2.736} & 2.738 \\ &  & 4B & 2.717 & 2.711 & 2.706 & 2.711 & 2.693 & 2.681 & \textbf{2.680} \\ &  & 7B & 2.699 & 2.693 & 2.686 & 2.694 & 2.675 & \textbf{2.645} & 2.649 \\ &  & 15.3B & 2.668 & 2.650 & 2.649 & 2.651 & 2.639 & 2.612 & \textbf{2.609} \\ &  & 26.8B & 2.638 & 2.649 & 2.630 & 2.633 & 2.638 & 2.590 & \textbf{2.583} \\  \midrule[\heavyrulewidth] \multirow{24}{*}{\makecell[l]{\Large Adam+\\\Large Param\\\Large Scaling}} & \multirow{6}{*}{\Large STP } & 1.1B & 2.778 & \textbf{2.760} & 2.784 & 2.779 & \textbf{2.760} & 2.815 & 2.765 \\ &  & 1.9B & 2.722 & 2.717 & 2.723 & 2.753 & 2.717 & 2.780 & \textbf{2.716} \\ &  & 4B & 2.668 & 2.668 & 2.677 & 2.730 & 2.668 & 2.737 & \textbf{2.667} \\ &  & 7B & 2.637 & 2.638 & 2.660 & 2.709 & 2.638 & 2.718 & \textbf{2.635} \\ &  & 15.3B & 2.600 & \textbf{2.599} & 2.630 & 2.687 & \textbf{2.599} & 2.691 & 2.601 \\ &  & 26.8B & 2.580 & 2.577 & 2.613 & 2.675 & 2.577 & 2.678 & \textbf{2.576} \\  \cmidrule{2-10}  & \multirow{6}{*}{\Large NTK } & 1.1B & 2.751 & 2.753 & 2.759 & 2.786 & 2.753 & 2.792 & \textbf{2.735} \\ &  & 1.9B & 2.699 & 2.698 & 2.722 & 2.752 & 2.698 & 2.760 & \textbf{2.690} \\ &  & 4B & 2.654 & 2.653 & 2.680 & 2.721 & 2.653 & 2.716 & \textbf{2.643} \\ &  & 7B & 2.624 & 2.626 & 2.658 & 2.703 & 2.626 & 2.696 & \textbf{2.613} \\ &  & 15.3B & 2.592 & 2.591 & 2.632 & 2.681 & 2.591 & 2.669 & \textbf{2.580} \\ &  & 26.8B & 2.570 & 2.566 & 2.623 & 2.667 & 2.566 & 2.654 & \textbf{2.554} \\  \cmidrule{2-10}  & \multirow{6}{*}{\Large muP } & 1.1B & 2.740 & \textbf{2.738} & 2.753 & 2.746 & \textbf{2.738} & 2.748 & \textbf{2.738} \\ &  & 1.9B & 2.698 & 2.694 & 2.727 & 2.705 & 2.694 & 2.711 & \textbf{2.691} \\ &  & 4B & 2.654 & \textbf{2.651} & 2.701 & 2.666 & \textbf{2.651} & 2.669 & 2.656 \\ &  & 7B & 2.627 & \textbf{2.623} & 2.682 & 2.643 & \textbf{2.623} & 2.649 & 2.625 \\ &  & 15.3B & \textbf{2.590} & 2.591 & 2.659 & 2.618 & 2.591 & 2.622 & 2.598 \\ &  & 26.8B & \textbf{2.574} & 2.575 & 2.655 & 2.606 & 2.575 & 2.611 & \textbf{2.574} \\  \cmidrule{2-10}  & \multirow{6}{*}{\Large MFP } & 1.1B & 2.773 & \textbf{2.770} & 2.775 & 2.832 & \textbf{2.770} & 2.847 & 2.787 \\ &  & 1.9B & \textbf{2.727} & 2.729 & 2.730 & 2.843 & 2.729 & 2.806 & 2.742 \\ &  & 4B & \textbf{2.675} & 2.701 & 2.698 & 2.804 & 2.701 & 2.776 & 2.697 \\ &  & 7B & 2.669 & 2.679 & 2.675 & 2.781 & 2.679 & 2.765 & \textbf{2.659} \\ &  & 15.3B & 2.641 & 2.648 & 2.656 & 2.765 & 2.648 & 2.740 & \textbf{2.622} \\ &  & 26.8B & 2.624 & 2.623 & 2.640 & 2.772 & 2.623 & 2.730 & \textbf{2.602} \\ 
\bottomrule[1.5\heavyrulewidth]
\end{tabularx}}
\end{table*}
\endgroup

\clearpage

\thispagestyle{plain}
\begin{SidewaysFigure}
\includegraphics[width=\linewidth, trim={0, 0, 0, 0},clip]{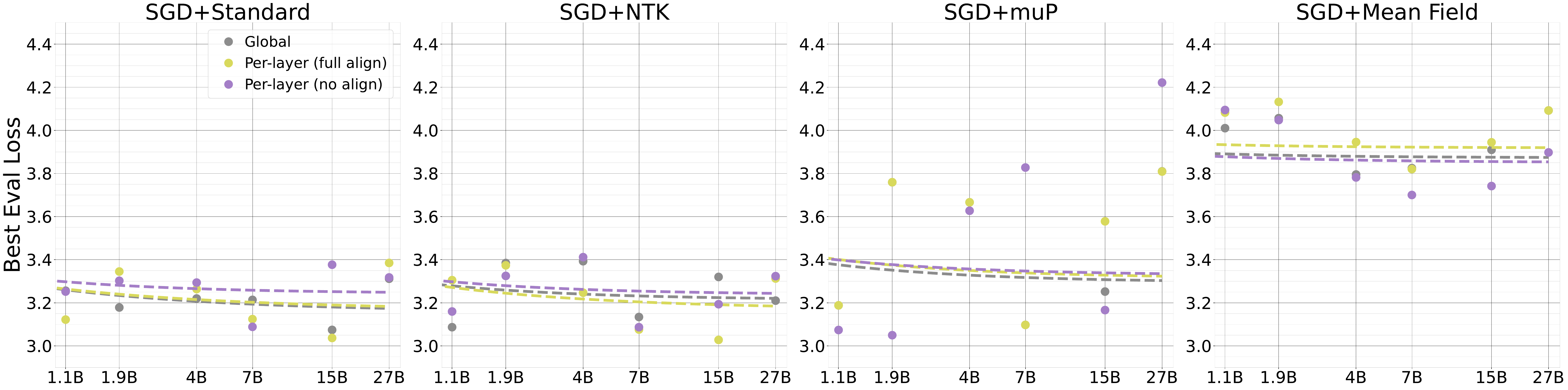}
\includegraphics[width=\linewidth, trim={0, 0, 0, 0},clip]{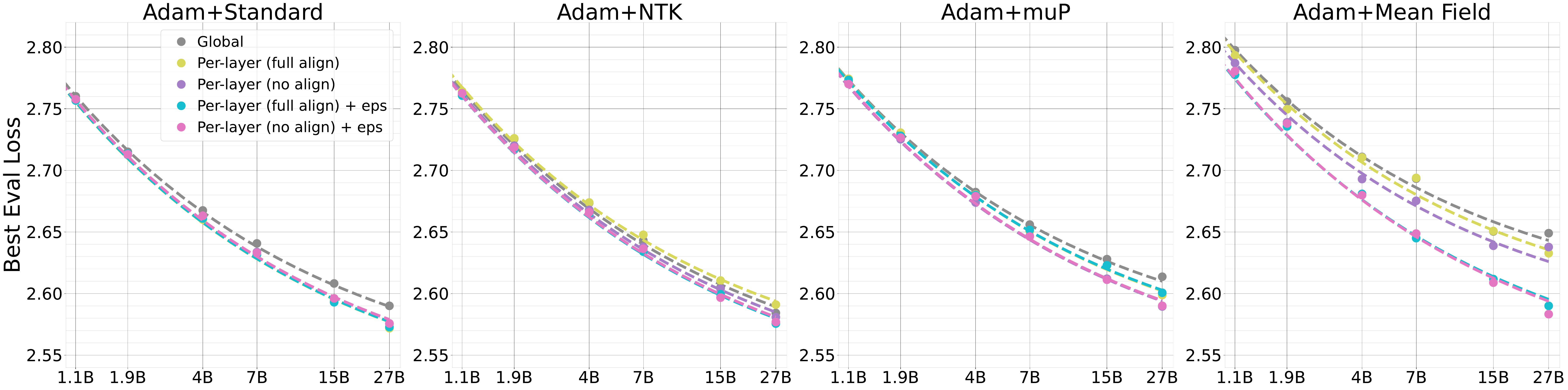}
\includegraphics[width=\linewidth, trim={0, 0, 0, 0},clip]{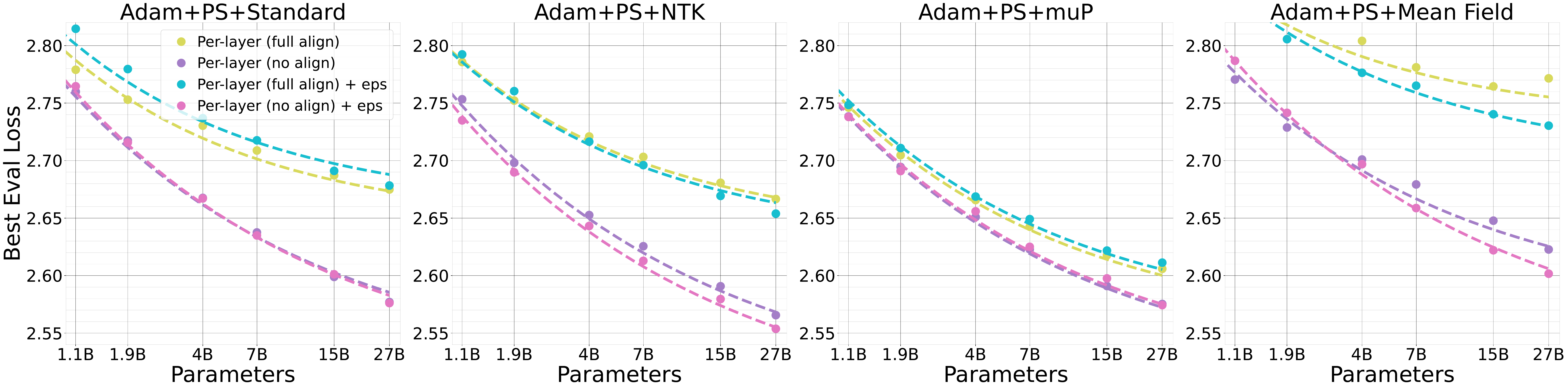}
\caption{Eval losses for the six largest model sizes for all settings with optimal constants. Rows = optimizers (SGD, Adam, Adam+parameter scaling), columns = parameterizations (standard, NTK, muP, Mean Field). Settings denoted "+eps" use per-layer epsilon with base epsilon = 1e-12. Note that Adam+parameter scaling global learning rate coincides with per-layer no alignment so there is no separate curve to show for global learning rates in the bottom row.}
\label{fig:app_scaling_optimal_constants}
\end{SidewaysFigure}
\clearpage

\thispagestyle{plain}
\begin{SidewaysFigure}
\includegraphics[width=\linewidth, trim={0, 0, 0, 0},clip]{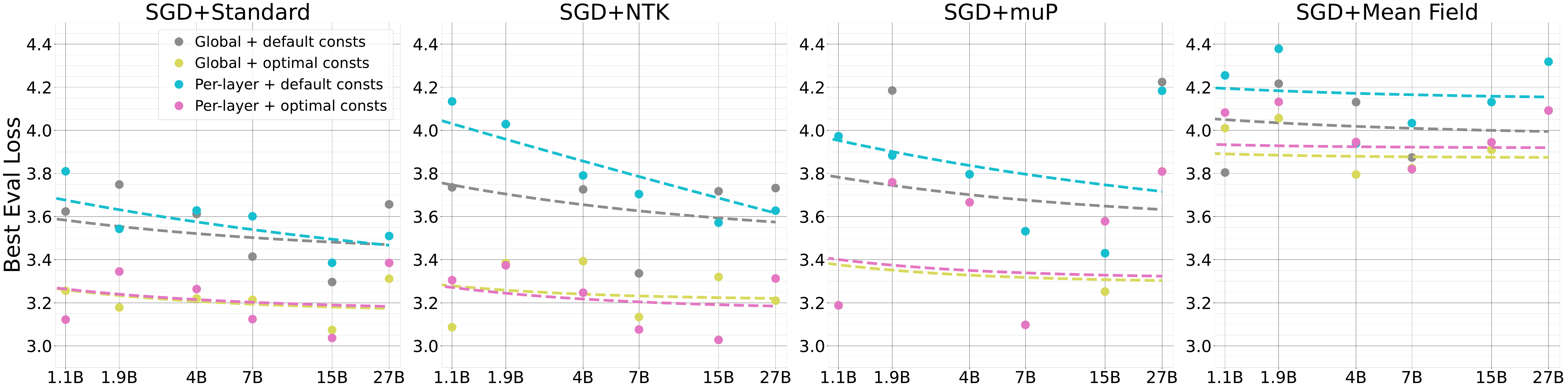}
\includegraphics[width=\linewidth, trim={0, 0, 0, 0},clip]{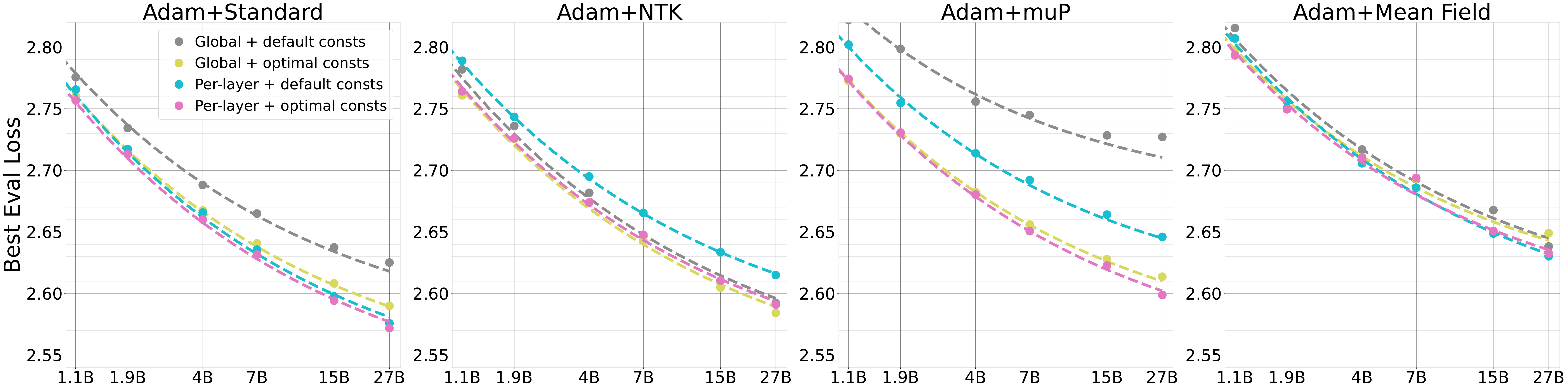}
\includegraphics[width=\linewidth, trim={0, 0, 0, 0},clip]{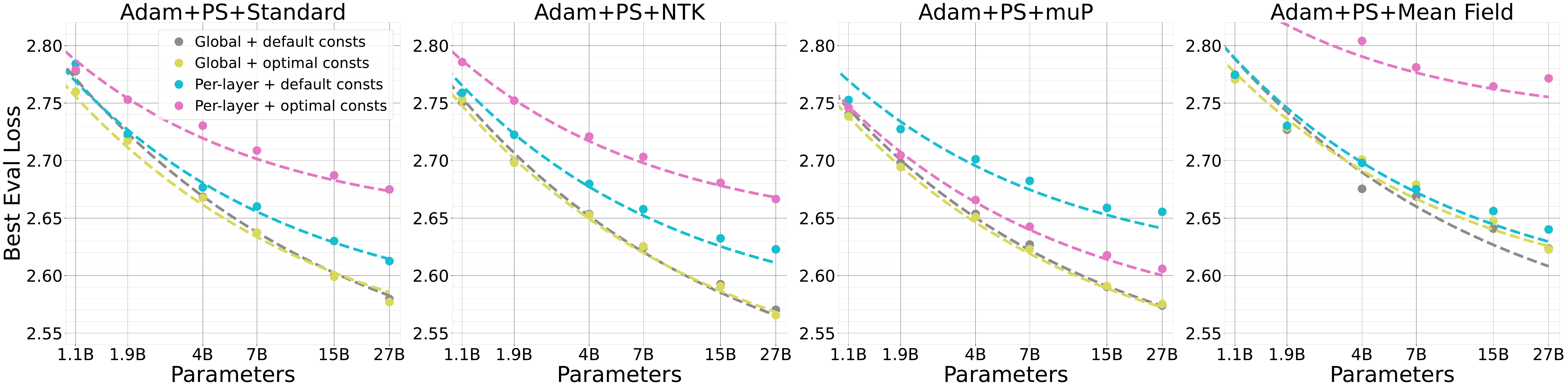}
\caption{Ablation showing eval losses for the six largest model sizes for all combinations of global or per-layer (full alignment) learning rates, and default or optimal constants. Rows = optimizers (SGD, Adam, Adam+parameter scaling), columns = parameterizations (standard, NTK, muP, Mean Field).}
\label{fig:app_scaling_ablation}
\end{SidewaysFigure}
\clearpage

\subsection{Additional Adam Epsilon Experiments}

In this section, we include additional results for epsilon experiments across all parameterizations.

\begin{figure*}[ht]
    \begin{center}
        \includegraphics[width=\linewidth, trim={0, 0, 0, 0},clip]{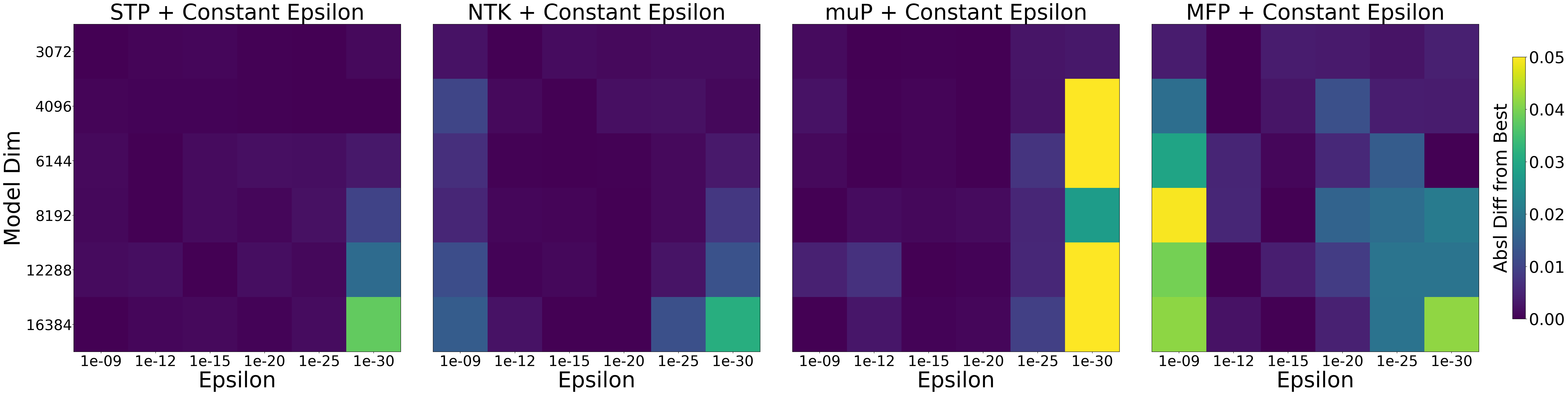}
       
        \figvspace
       
        \includegraphics[width=\linewidth, trim={0, 0, 0, 0},clip]{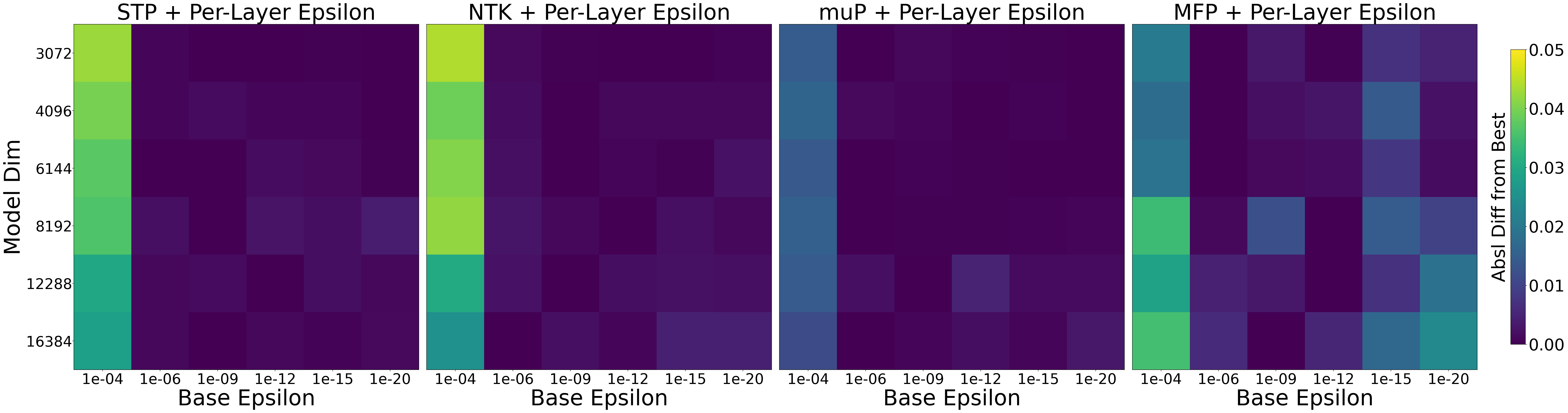}

        \caption{\textbf{All parameterizations are affected by epsilon but different parameterizations have different levels of sensitivity and different optimal values.} All experiments use Adam + per-layer learning rates assuming full alignment + optimal constants. Top row = constant epsilon, bottom row = per-layer epsilon. Number of training steps = $50{,}000$.}
        \label{fig:epsilon_appendix_heatmaps}
        \vspace{-24pt}
    \end{center}
\end{figure*}

In \cref{fig:epsilon_appendix_heatmaps}, we show heatmaps from tuning epsilon on all parameterizations. For both the constant epsilon and per-layer epsilon settings, we perform a learning rate sweep at each model dim for each value of epsilon or base epsilon. Using the best eval loss from each learning rate sweep, the heatmaps compare the eval loss to the other values of epsilon within that epsilon setting for that parameterization and model size. That is, each heatmap entry is colored based on the absolute difference to the best eval loss of the six entries in its row.

These results show that all parameterizations are affected by the choice of epsilon, but that different parameterizations have both different optimal values and different levels of sensitivity to epsilon. In all parameterizations, a constant epsilon of 1e-30 is too small and harms performance for the largest $26.8B$ parameter model ($D = 16{,}384$). Similarly, for per-layer epsilon, a base epsilon of 1e-4 is too large and harms performance. Compared to standard parameterization, muP tolerates large values of epsilon better (e.g. per-layer epsilon with base epsilon = 1e-4) and is harmed more by very small value epsilon (e.g. constant epsilon = 1e-30). Mean-field parameterization is most sensitive to epsilon and has the most narrow range of epsilon values with good performance, for both the constant and per-layer epsilon settings. NTK is overall quite similar to standard parameterization, with the exception that it NTK performance is harmed slightly by the default constant value of 1e-9.

\begin{figure*}[ht]
    \begin{center}
        \includegraphics[width=0.48\linewidth, trim={0, 0, 0, 0},clip]{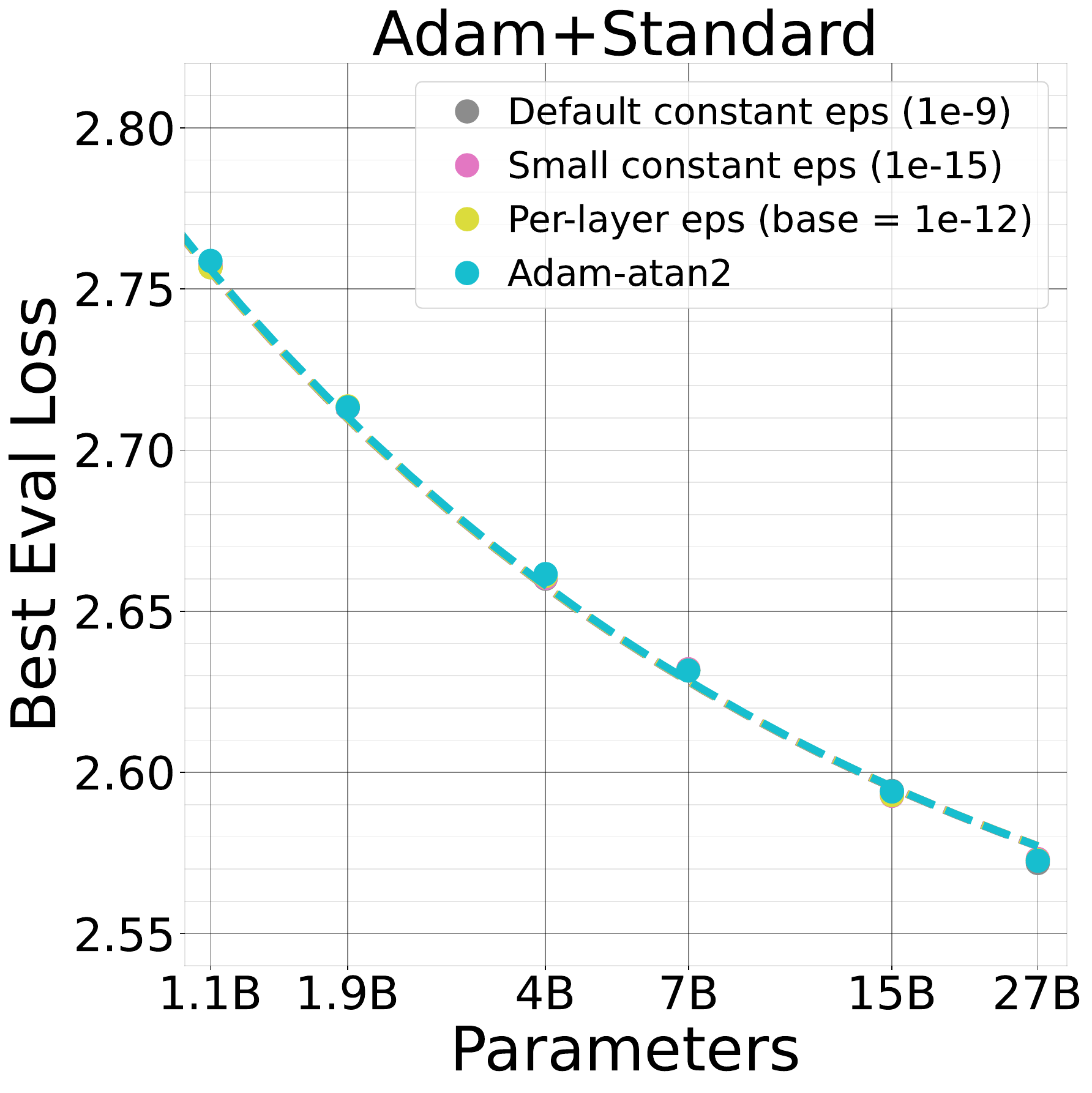}
        \hfill
        \includegraphics[width=0.48\linewidth, trim={0, 0, 0, 0},clip]{icml2024/figures/epsilon/ntk_epsilon_eval_loss_comparisons_zoom6.pdf}
       
        \figvspace
       
        \includegraphics[width=0.48\linewidth, trim={0, 0, 0, 0},clip]{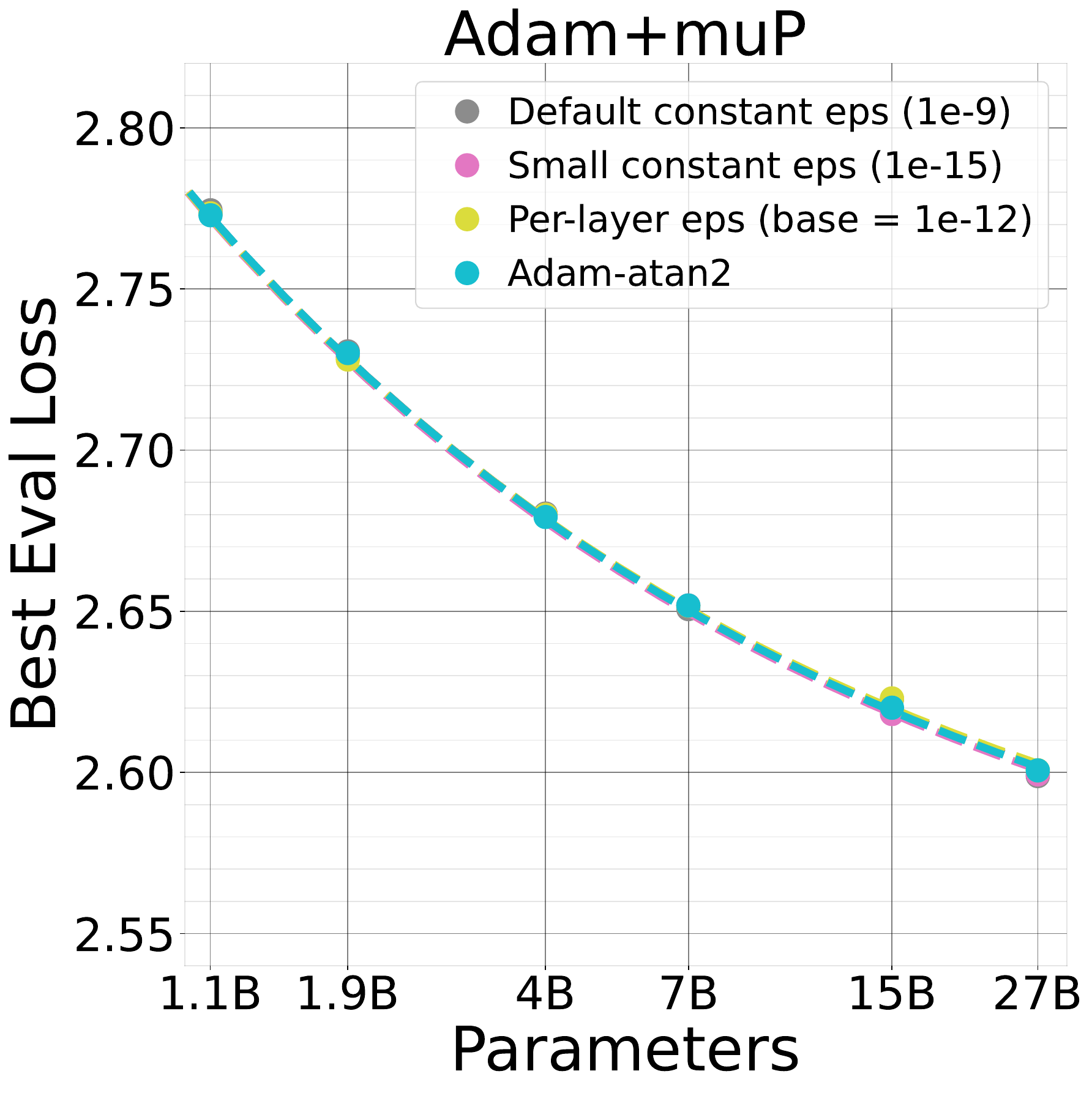}
        \hfill
        \includegraphics[width=0.48\linewidth, trim={0, 0, 0, 0},clip]{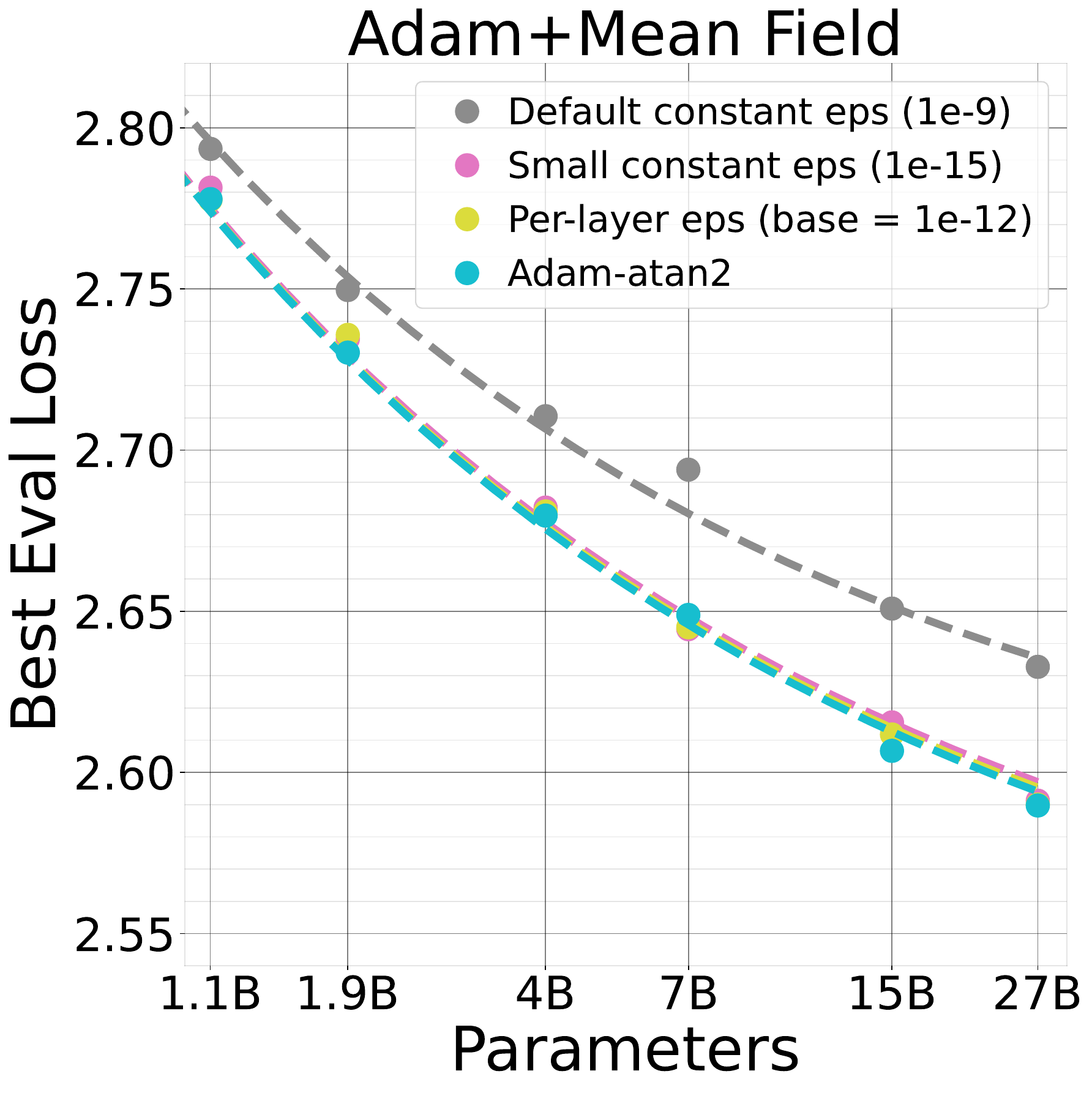}

        \caption{\textbf{All three epsilon mitigations similarly improve Adam performance on NTK and MFP, and do not change performance on STP and muP.} Experiments for all parameterizations comparing the three epsilon mitigations for Adam (small constant epsilon = 1e-15, per-layer epsilon with base epsilon = 1e-12, and Adam-atan2) to the baseline default constant epsilon of 1e-9.}
        \label{fig:epsilon_appendix_scaling_plots}
        \vspace{-24pt}
    \end{center}
\end{figure*}

In \cref{fig:epsilon_appendix_scaling_plots}, we show the eval loss vs model size for all three epsilon mitigations (constant epsilon = 1e-15, per-layer epsilon with base epsilon = 1e-12, and \emph{Adam-atan2}) compared against the baseline with the default constant epsilon of 1e-9 in all parameterizations. All three epsilon mitigations result in similar performance improvements, with different parameterizations having different sensitivity. Both standard and muP perform well with the default constant value of epsilon (1e-9) in our model sizes and are not affected by any of the epsilon mitigations. NTK performance improves slightly and MFP performance improves more substantially with all of the epsilon mitigations.

\clearpage

\subsection{Weight Decay Experiments}
\label{app:weight_decay}
In current practice, weight decay is typically used for training large Transformers and may improve training stability by providing a small amount of regularization \citep{brown2020gpt3}. For the majority of our experiments, we do not use weight decay in order to reduce the number of possible confounding factors and focus our investigation on the impact of the parameterization and optimizer choices.

As a cross-check to ensure our conclusions are likely to transfer to settings with weight decay, we perform a set of experiments for Adam using per-layer learning rates assuming full alignment with a small constant weight decay of 1e-4, using ``decoupled" or ``independent" weight decay as proposed in AdamW \citep{loshchilov2018decoupled}. In decoupled weight decay, the weight decay is not scaled by the base learning rate; our value of 1e-4 decoupled weight decay corresponds to the higher values around 1e-2 or 1e-1 typically used for weight decay that does scale by the base learning rate.

Across parameterizations, with weight decay we see an improvement in the eval loss but similar learning rate scaling compared to the no weight decay setting. This suggests that while weight decay plays a beneficial role, it does not significantly alter the scaling behavior or have major parameterization-specific interactions and therefore we expect that our conclusions should transfer to settings with a small amount of weight decay. Learning rate sweeps for the weight decay experiments are included in  \cref{fig:appendix_adam_weight_decay}.

\vspace{48pt}

\begin{figure*}[ht]
    \begin{center}
        \includegraphics[width=\linewidth, trim={0, 0, 0, 0},clip]{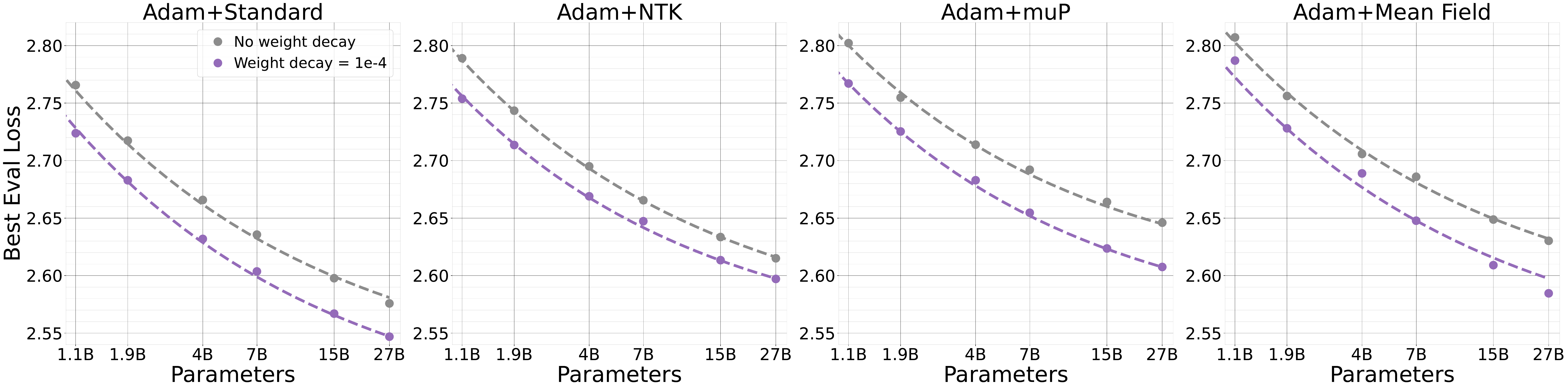}
       
        \caption{\centering{\textbf{Weight decay = 1e-4 (decoupled) improves the eval loss for all parameterizations and model sizes but overall scaling behavior is similar.} All experiments use Adam + per-layer learning rates assuming full alignment + default constants. Number of training steps = $50{,}000$.}}
        \label{fig:wd_lr_sweeps}
    \end{center}
\end{figure*}

\vfill

\clearpage
\thispagestyle{plain}
\begin{SidewaysFigure}
        \includegraphics[width=\linewidth]{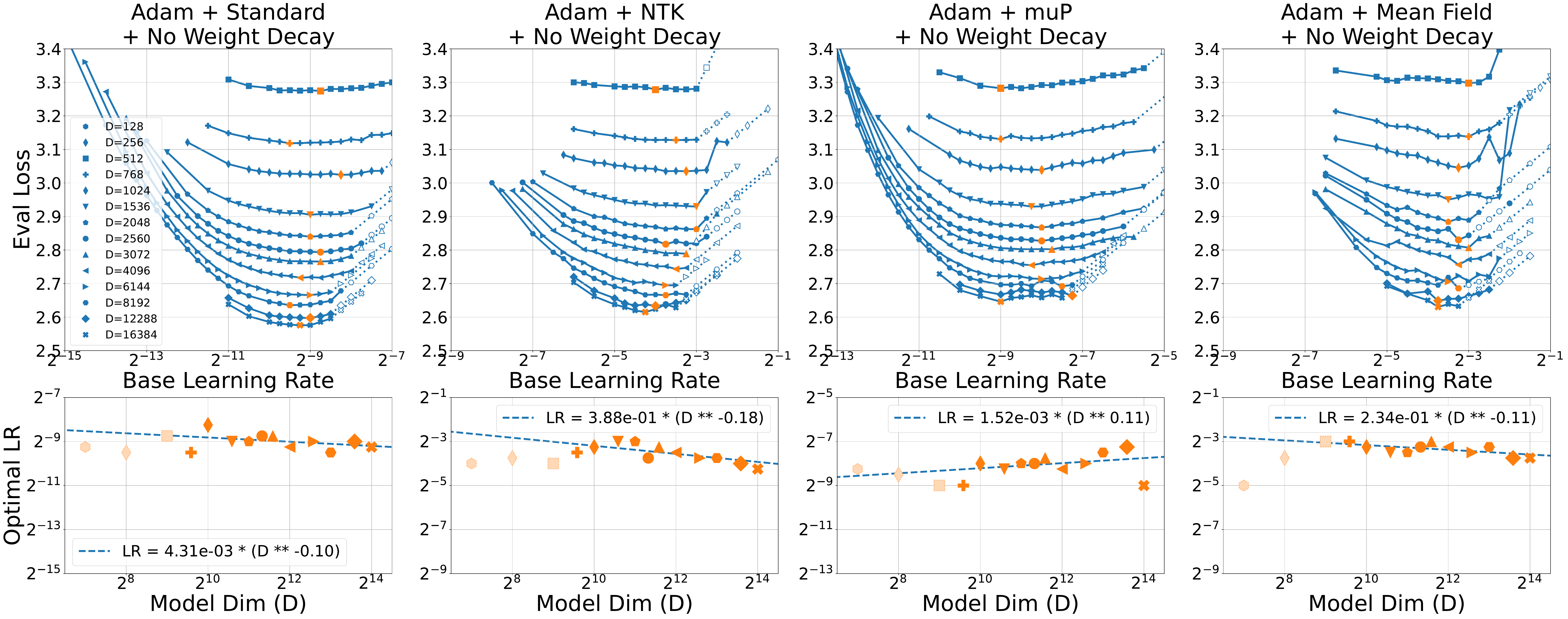}
        
        \figvspace
        
        \includegraphics[width=\linewidth]{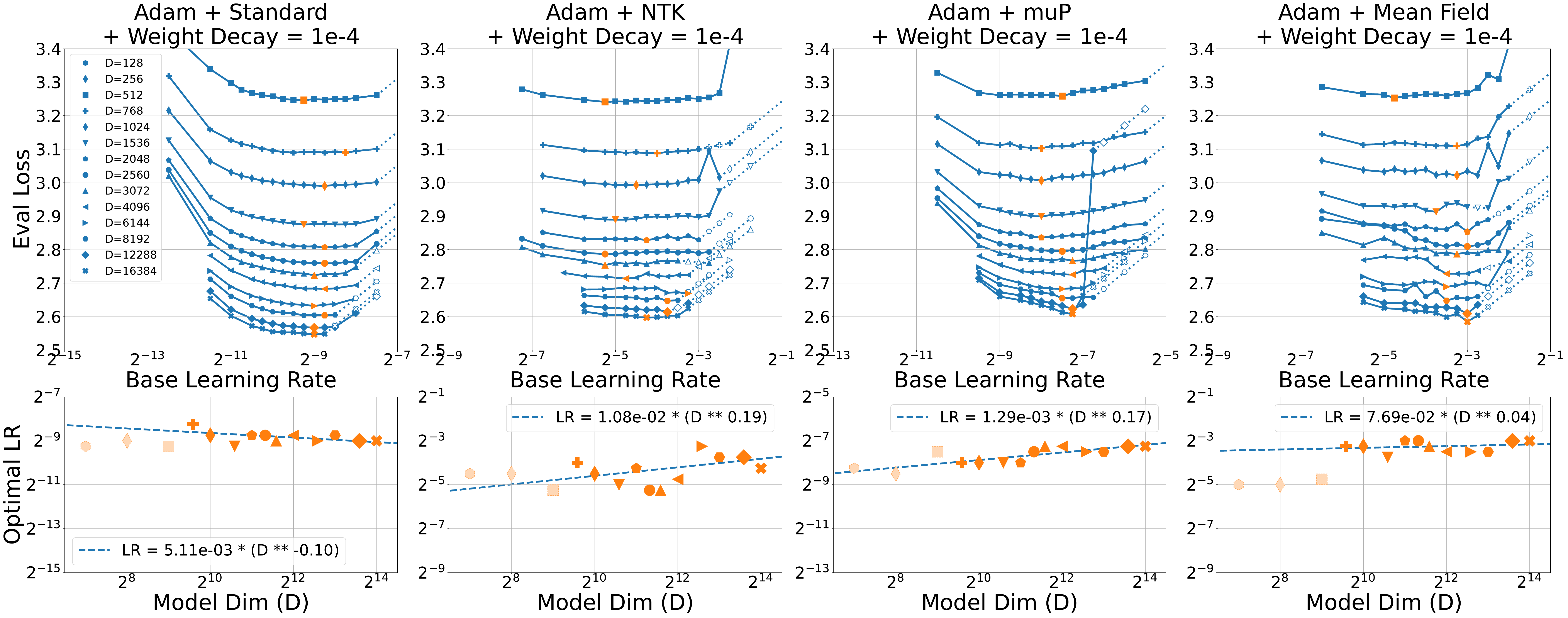}
        \caption{\textbf{Weight decay improves eval losses but learning rate scaling behavior is similar.} Top = Adam + per-layer learning rates assuming full alignment + default constants + no weight decay. Bottom = Adam + per-layer learning rates assuming full alignment + default constants + decoupled weight decay = 1e-4. Number of training steps = $50{,}000$.}
        \label{fig:appendix_adam_weight_decay}
\end{SidewaysFigure}
\clearpage

\subsection{Adafactor and Adam + Parameter Scaling experiments}
\label{app:adafactor_adam_ps}
As a cross-check that Adafactor and Adam + parameter scaling are in similar width-scaling regimes, we compare the two optimizers on all parameterizations in two settings: global learning rate + default constants and per-layer learning rate + full alignment + optimal constants. Due to the factored matrices in Adafactor, we encountered issues with tensor shape mismatches when using Adafactor with our implementation of FSDP, which the limited the model sizes we could use for Adafactor. Instead, we use Adam + parameter scaling for all our experiments in \cref{sec:results_per_layer}.

\begin{figure*}[ht]
    \begin{center}
        \includegraphics[width=\linewidth, trim={0, 0, 0, 0},clip]{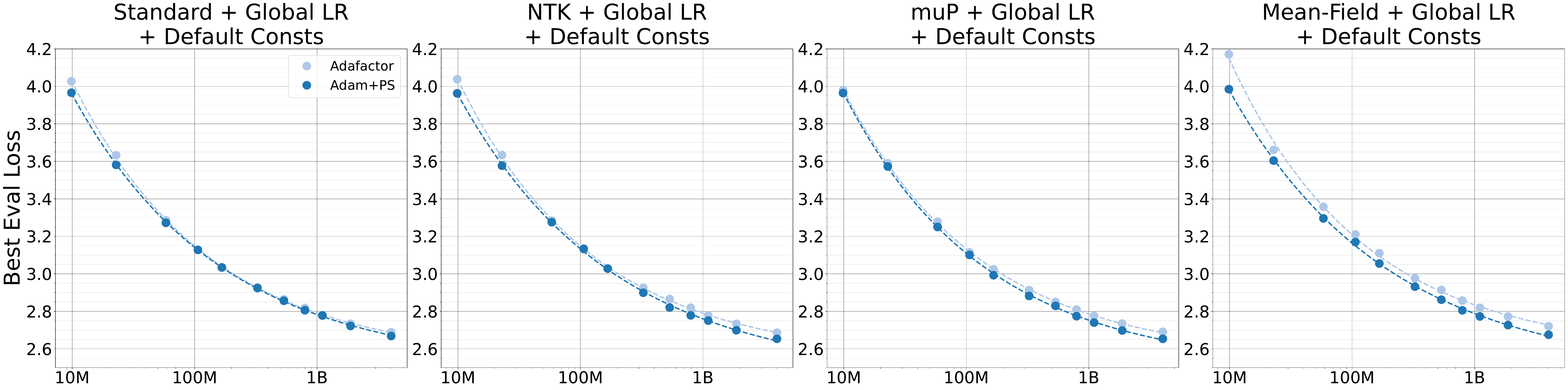}
       
        \figvspace

        \includegraphics[width=\linewidth, trim={0, 0, 0, 0},clip]{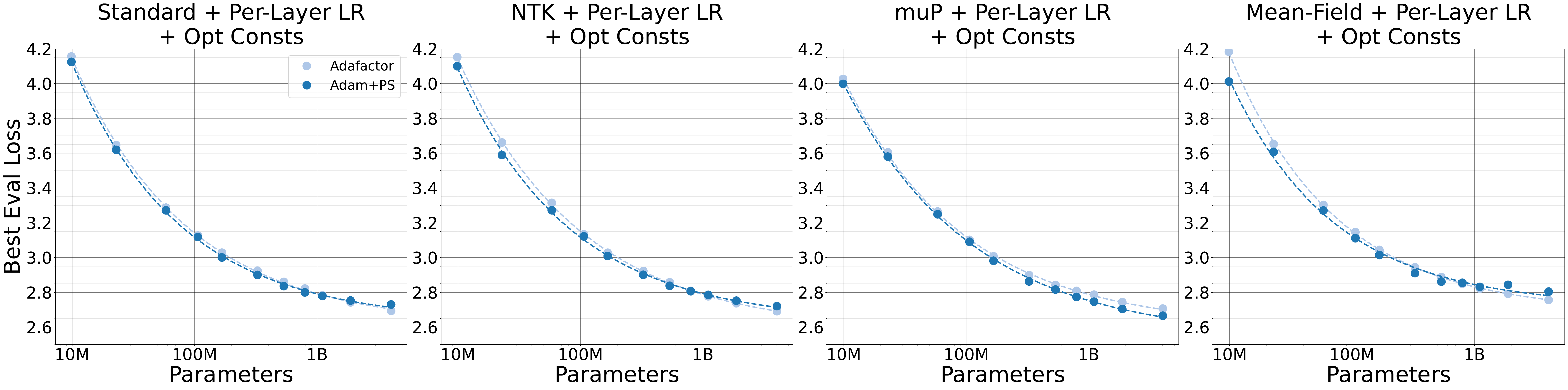}
        \caption{\centering{\textbf{Adafactor and Adam + parameter scaling are in the same width-scaling regime.} Figures show the best eval loss across a learning rate sweep at each model size for both optimizers. Top row = global learning rate + default constants, bottom row = per-layer learning rate assuming full alignment + optimal constants. There are minor performance differences between the optimizers but the overall scaling behavior is similar in all settings.}}
        \label{fig:adafactor_cross_check}
    \end{center}
\end{figure*}

See \cref{app:optim_details} for details on the optimizers and hyperparameters. The differences between Adam+parameter scaling and Adafactor are: the factored second moment estimate in Adafactor, different values of beta1 and beta2, update clipping in Adafactor, and the value of epsilon. We see in \cref{fig:adafactor_cross_check} that there are minor differences in performance but overall the optimizers show similar scaling behavior across model sizes up to $4B$ parameters, suggesting these two optimizers should be considered members of the same width-scaling regime.

\vspace{48pt}

\clearpage

\subsection{Fixed Step vs Compute Optimal experiments}
\label{app:fixed_vs_compute_opt}
Since the cost of compute is currently the most significant factor that limits the scale of large model training runs, the dominant paradigm for training large models in practice is the compute-optimal regime. The compute-optimal setting~\citep{kaplan2019notes} aims to maximize model performance under a fixed budget of FLOPS for training, where these FLOPS can be traded off between the number of parameters in the model and the number of training tokens the model is trained on. The Chinchilla paper~\citep{hoffmann2022training} finds empirically that the optimal tradeoff occurs when the number of parameters and number of tokens scale in proportion. When the batch size and context length are fixed, as in our setting, the number of training tokens is proportional to the number of training steps. Due to the $n \times n$ parameter matrices in dense hidden layers with width $n$, the number of parameters grow quadratically with respect to the width. Therefore, the Chinchilla results imply that the compute optimal number of steps grows \emph{quadratically} with respect to the model width.

This contradicts the fixed step assumption used in the theoretical derivations in both this paper and \citet{yang2021tensoriv,yang2023tensorivb}, which assume that the number of training steps $T$ is $O(1)$. Intuitively, this fixed step assumption is used so that the derivations can consider the contributions to the scaling exponents of a single step at a time: if we satisfactorily bound the contribution of each step to the scaling exponents, and then take only a constant number of steps, then the constant number of steps does not introduce any width-dependent scaling factors. The naive extension of this theory to a setting with $\Theta(n^2)$ instead of $O(1)$ training steps would give impractical bounds: in the worst-case analysis, each learning rate would need to be divided by $n^2$ to correct for the $n^2$ number of steps giving learning rates that are far too conservative to be useful.

We therefore take an empirical approach rather than a worst-case theoretical analysis to investigate the role of the training horizon. We perform a set of experiments using both fixed step and compute optimal training horizons in the global learning rate settings for SGD+momentum, Adam and Adafactor across all parameterizations using default constant learning rate multipliers. In each setting, we sweep both model width and learning rate, and then fit a power law with an irreducible loss term to determine the scaling exponent for the optimal learning rate. The measured learning rate exponents are reported in \cref{tab:compute_opt_exponents}. For all fixed step experiments, we train for $50{,}000$ steps. For the compute optimal setting, we compute the training horizon using the Chinchilla-optimal heuristic \citep{hoffmann2022training} with 20x multiplier, i.e. the number of training tokens is equal to 20 times the number of non-embedding parameters. Full results for the learning rate sweeps are included in \cref{fig:app_compute_opt_sgd}, \ref{fig:app_compute_opt_adam} and \ref{fig:app_compute_opt_adafactor}.

\begin{table}[htbp]
  \centering
  \caption{\textbf{Power law exponents fit to the optimal learning rate vs model dimension} for each optimizer $\times$ parameterization combination, measured for fixed step (50k) and compute optimal training horizon experiments.}
  \vspace{2pt}
  \begin{adjustbox}{scale=0.8}
    \begin{tabular}{rrrr}
        \toprule
            &           & \textbf{Fixed Step (50k)}           & \textbf{Compute Optimal}  \\
         \midrule
          \multirow{4}{*}{SGD} & STP        & -0.38         & -1.27  \\
          & NTK                             & 0.56          & 0.04   \\
          & muP                             & -0.17         & -0.85  \\
          & MFP                             & 0.31          & -0.41  \\  \midrule
          \multirow{4}{*}{Adam} & STP       & -0.95         & -1.18  \\
          & NTK                             & -0.66         & -0.67  \\
          & muP                             & -1.09         & -1.38  \\
          & MFP                             & -0.16         & -0.45  \\ \midrule
          \multirow{4}{*}{Adafactor} & STP  & -0.12         & -0.55  \\
          & NTK                             & -0.10         & -0.52  \\
          & muP                             & -0.15         & -0.66  \\
          & MFP                             & -0.09         & -0.57 \\
        \bottomrule
    \end{tabular}
    \end{adjustbox}
  \label{tab:compute_opt_exponents}
\end{table}

Our exponent measurements show that in every parameterization $\times$ optimizer setting, the learning rate exponent in the compute optimal setting is \emph{smaller} than in the fixed steps setting, indicating that the learning rate would need to decrease more aggressively as width grows than predictions from the fixed steps setting would imply. The median difference from the twelve optimizer $\times$ parameterization settings is $0.46$ and ranges from $0.01$ to $0.89$. We note this difference of $0.46$ is much less than the difference of $2$ that would come from the naive worst-case theoretical analysis.

This result has implications both for theoretical and empirical settings. First, it motivates theoretical work to consider the compute optimal setting instead of the fixed steps setting. Second, it implies that hyperparameter search should be careful not to assume that results from the fixed step setting will extrapolate to the compute optimal setting. In particular, given a fixed compute budget to spend on hyperparameter search before training a single large model with a compute optimal horizon, one possible approach to choosing the learning rate would be to train model sizes close to the final model size for shorter training horizons, and use this to extrapolate the best learning rate for the large model compute optimal run. A priori, this strategy might be advantageous because the shorter training horizons let you use larger models for the same hyperparameter search budget so the search occurs closer in size to the final model. However, our results suggest that this may not be a viable strategy: if we extrapolate a learning rate to larger models based on an exponent fitted in the fixed steps setting, we may significantly overestimate the learning rate that is optimal in the compute optimal setting. Instead, we recommend considering performing the hyperparameter search for the learning rate by training smaller models at their compute optimal training horizon and then extrapolating across model sizes to the compute optimal setting for the largest model.

\clearpage
\thispagestyle{plain}
\begin{SidewaysFigure}
\centering{\textbf{SGD global learning rate experiments: 50k steps and compute optimal}}\\
\includegraphics[width=0.98\linewidth]{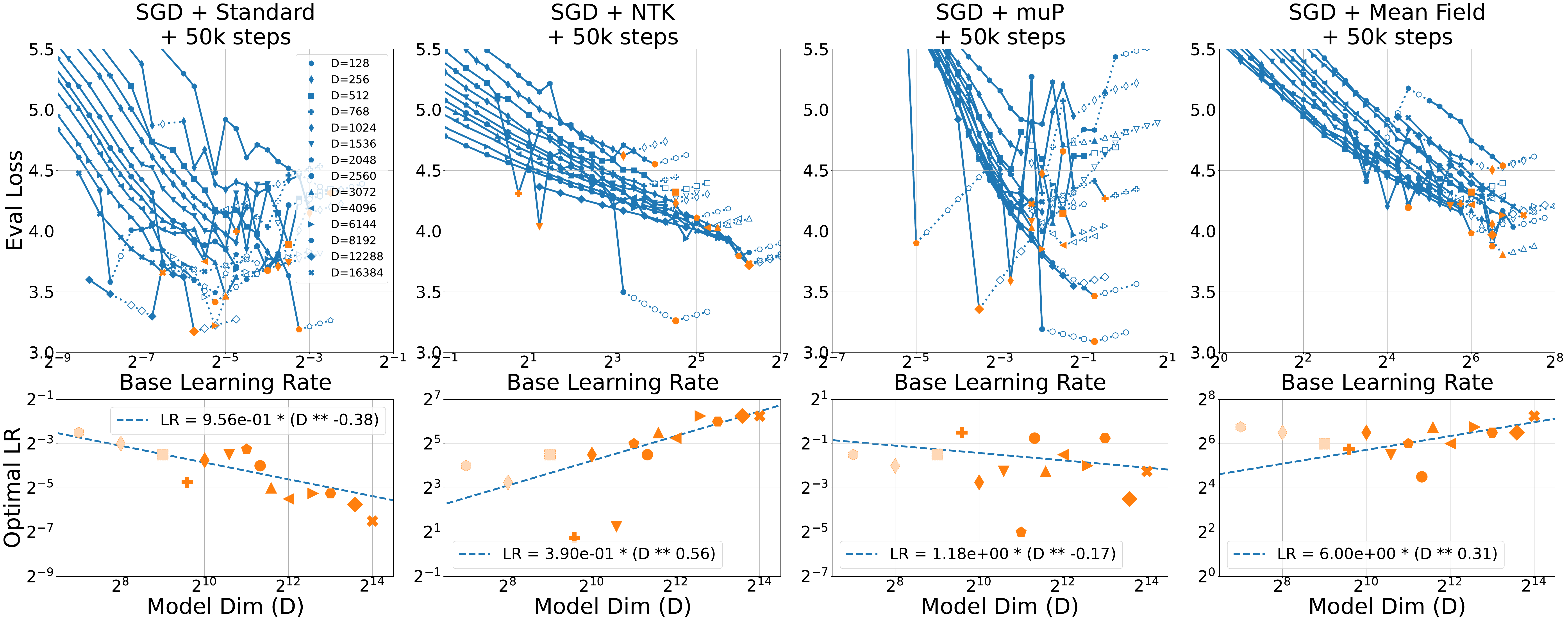}

\figvspace

\includegraphics[width=0.98\linewidth]{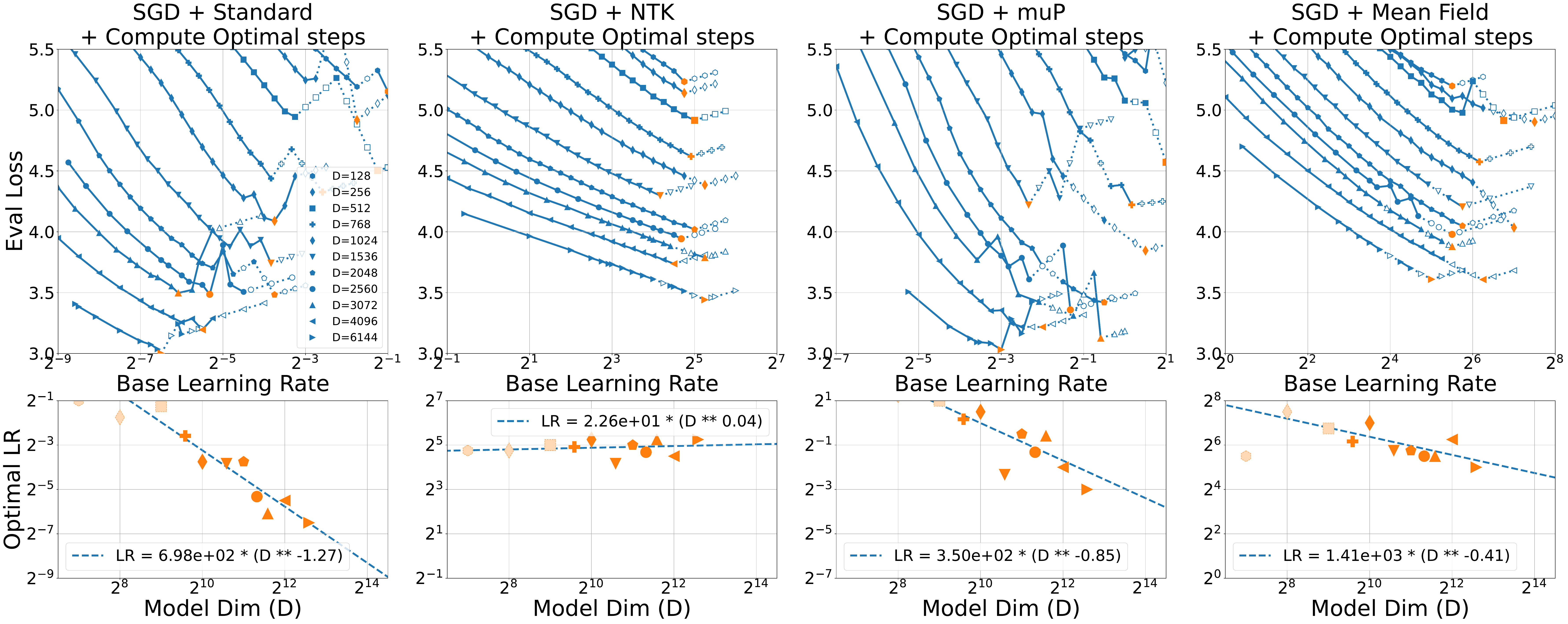}
\caption{SGD learning rate sweeps and power laws fit to optimal learning rate vs model dim, using global learning rate and default constants. Top = $50{,}000$ steps. Bottom = compute optimal (Chinchilla 20x) training steps.}
\label{fig:app_compute_opt_sgd}
\end{SidewaysFigure}
\clearpage

\thispagestyle{plain}
\begin{SidewaysFigure}
\centering{\textbf{Adam global learning rate experiments: 50k steps and compute optimal}}\\
\includegraphics[width=0.98\linewidth]{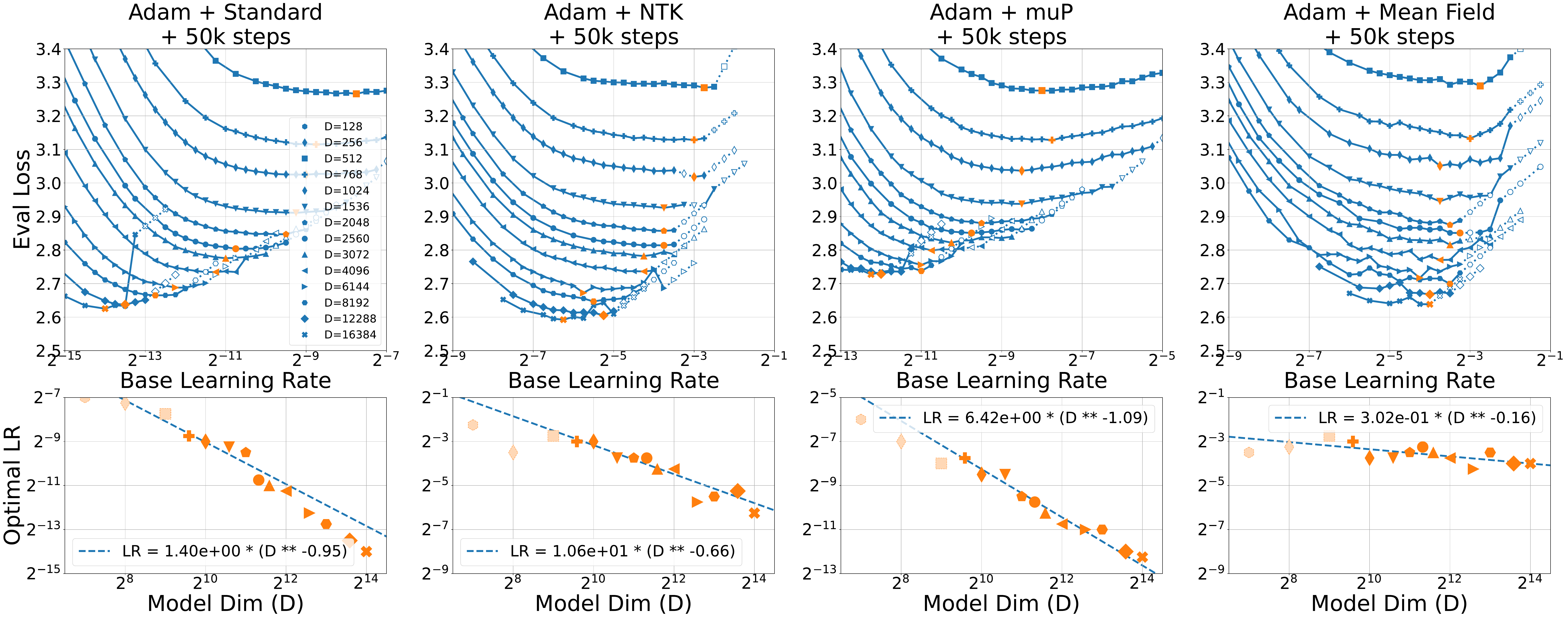}

\figvspace

\includegraphics[width=0.98\linewidth]{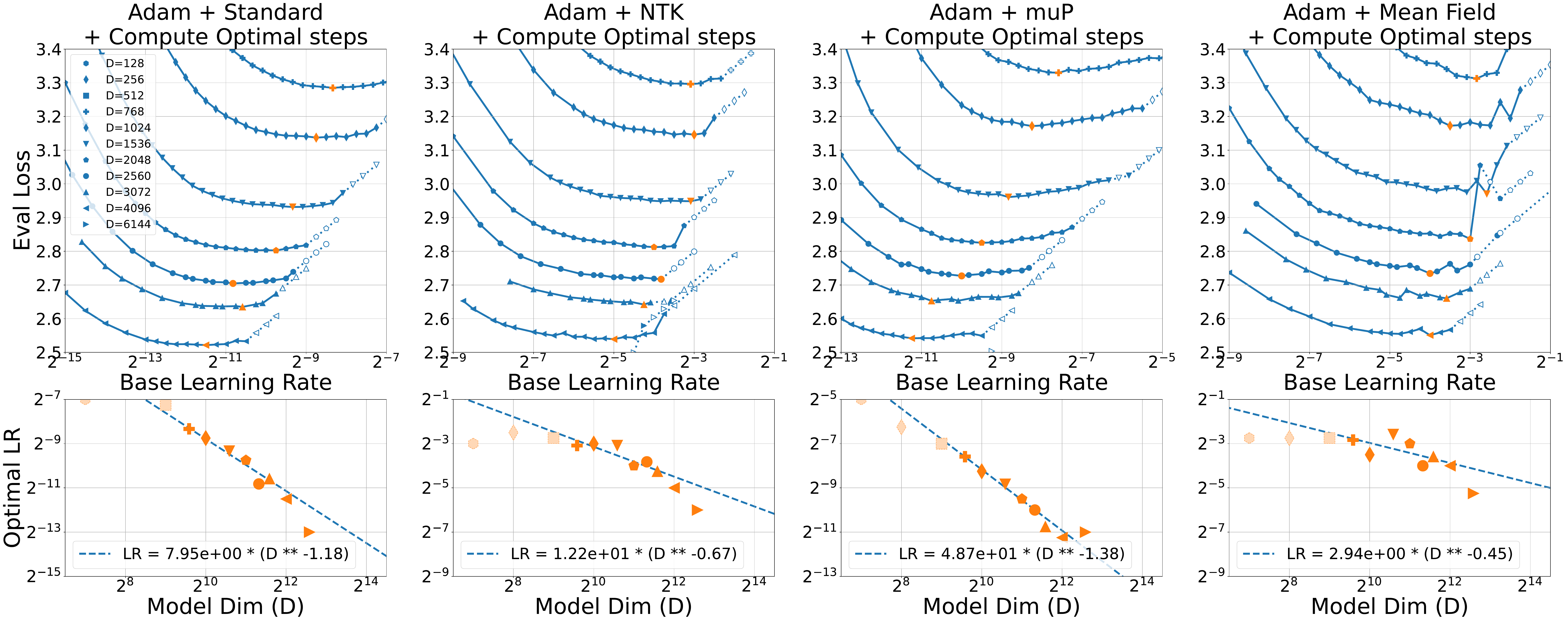}
\caption{Adam learning rate sweeps and power laws fit to optimal learning rate vs model dim, using global learning rate and default constants. Top = $50{,}000$ steps. Bottom = compute optimal (Chinchilla 20x) training steps.}
\label{fig:app_compute_opt_adam}
\end{SidewaysFigure}
\clearpage

\thispagestyle{plain}
\begin{SidewaysFigure}
\centering{\textbf{Adafactor global learning rate experiments: 50k steps and compute optimal}}\\
\includegraphics[width=0.98\linewidth]{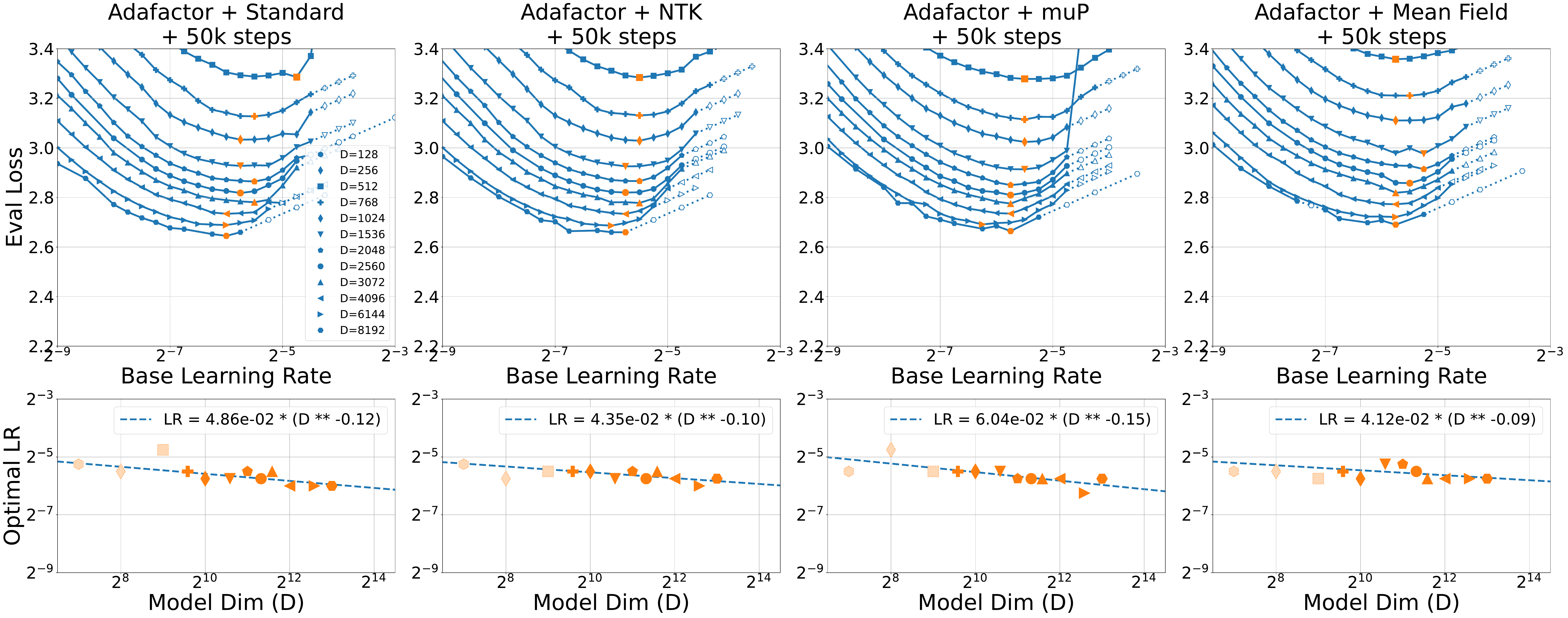}

\figvspace

\includegraphics[width=0.98\linewidth]{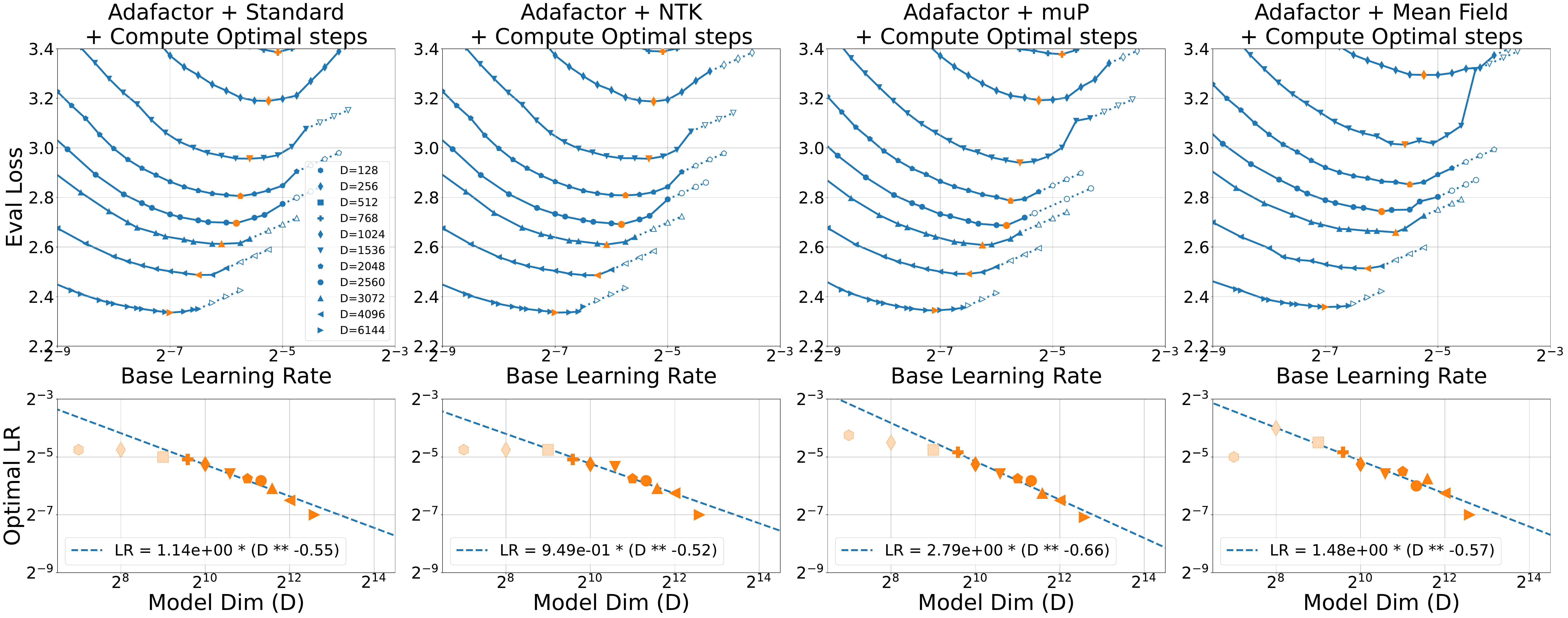}
\caption{Adafactor learning rate sweeps and power laws fit to optimal learning rate vs model dim, using global learning rate and default constants. Top = $50{,}000$ steps. Bottom = compute optimal (Chinchilla 20x) training steps.}
\label{fig:app_compute_opt_adafactor}
\end{SidewaysFigure}
\clearpage

\clearpage
\thispagestyle{plain}
\begin{SidewaysFigure}
\subsection{Learning Rate Sweeps for SGD, all settings}
\label{sec:app_lr_sweeps_sgd}
\vspace{12pt}
\includegraphics[width=0.98\linewidth]{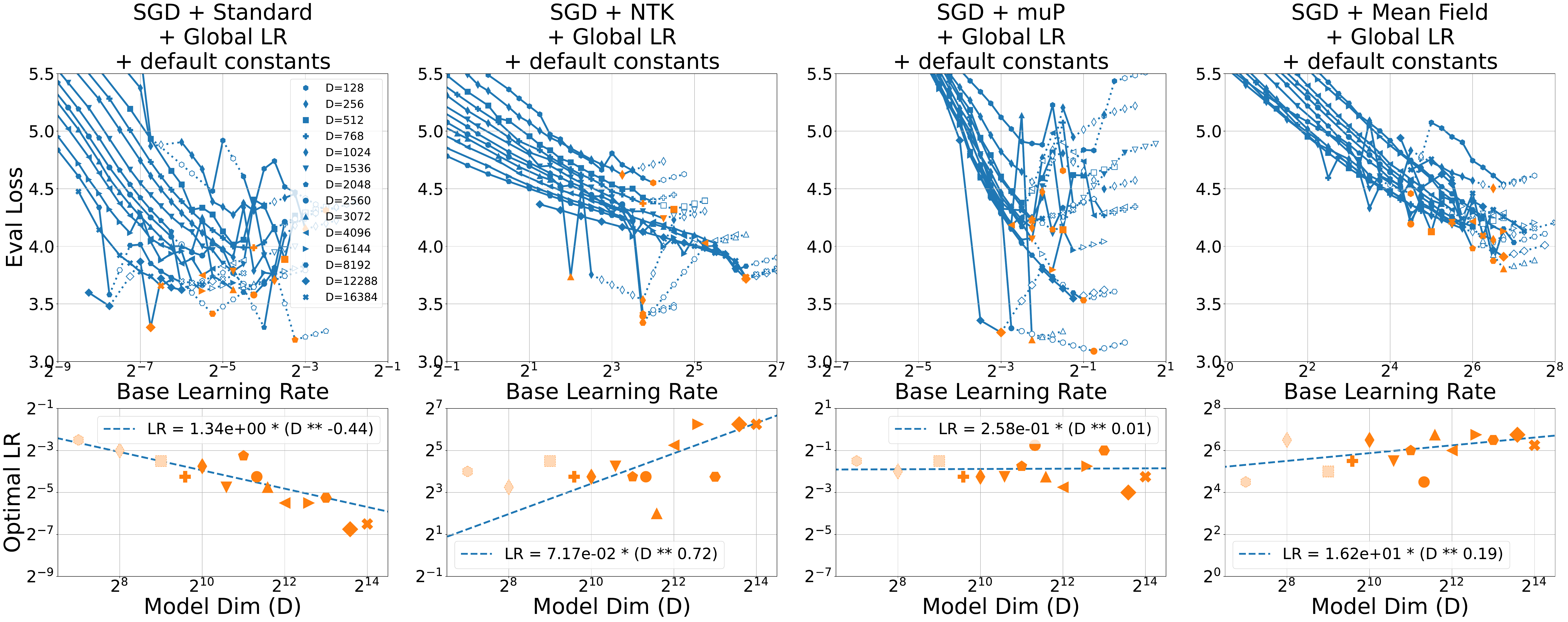}

\figvspace

\includegraphics[width=0.98\linewidth]{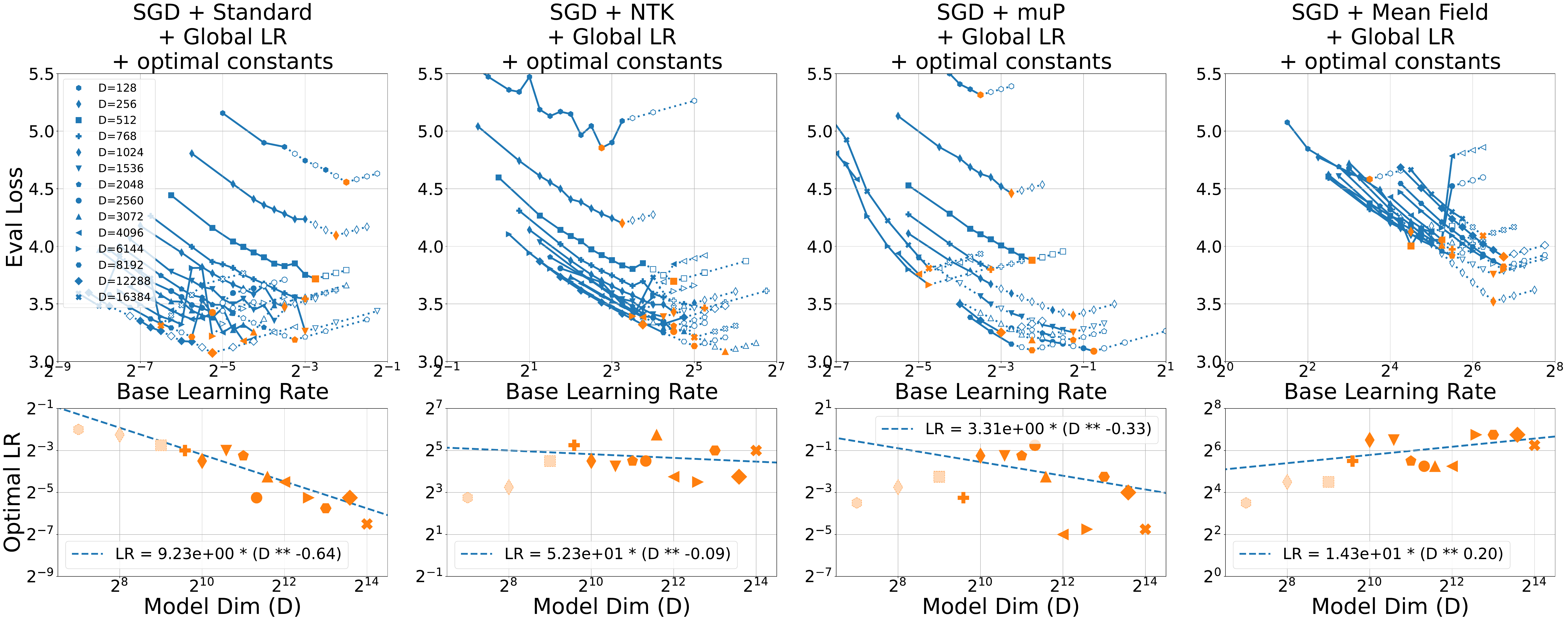}
\caption{Learning rate sweeps and power laws fit to optimal learning rate vs model dim. Top = SGD + global learning rate + default constants. Bottom = SGD + global learning rate + optimal constants. Number of training steps = $50{,}000$.}
\end{SidewaysFigure}
\clearpage

\thispagestyle{plain}
\begin{SidewaysFigure}
\includegraphics[width=0.98\linewidth]{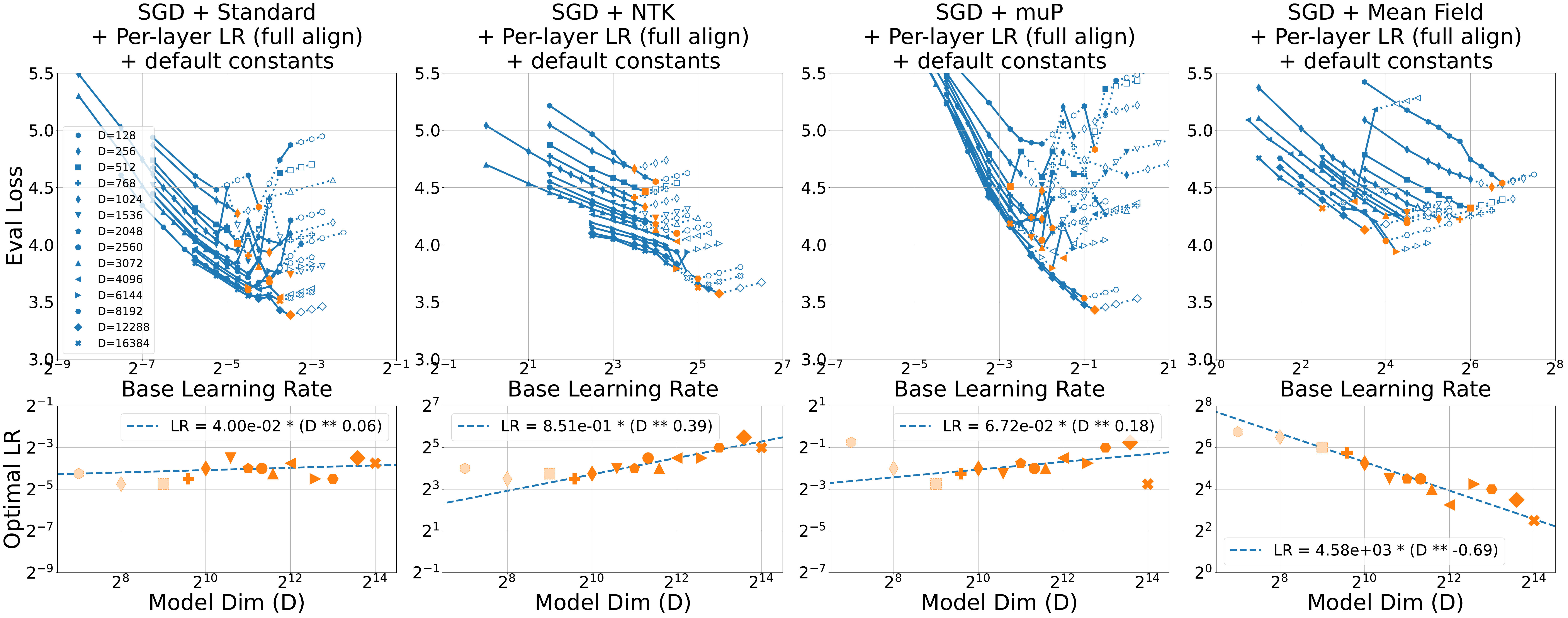}

\figvspace

\includegraphics[width=0.98\linewidth]{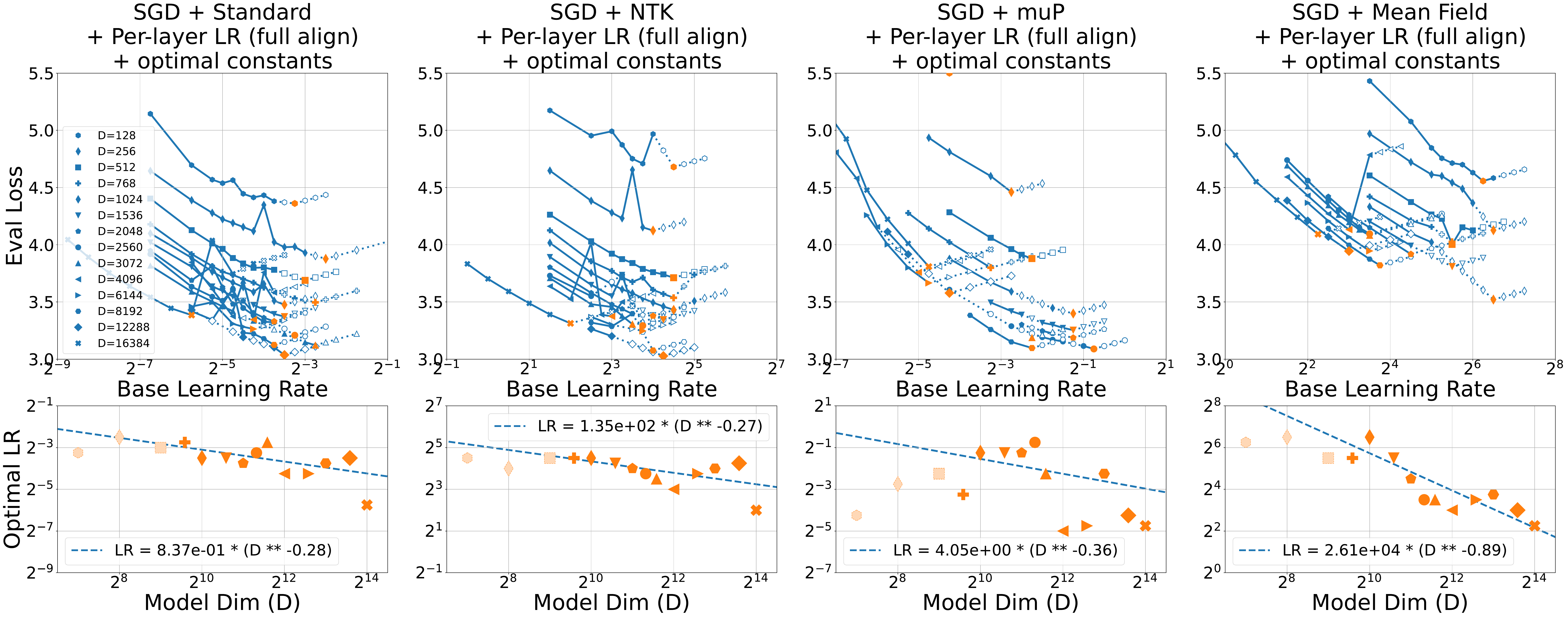}
\caption{Learning rate sweeps and power laws fit to optimal learning rate vs model dim. Top = SGD + per-layer learning rate assuming full alignment + default constants. Bottom = SGD + per-layer learning rate assuming full alignment + optimal constants. Number of training steps = $50{,}000$.}
\end{SidewaysFigure}
\clearpage

\thispagestyle{plain}
\begin{SidewaysFigure}
\includegraphics[width=0.98\linewidth]{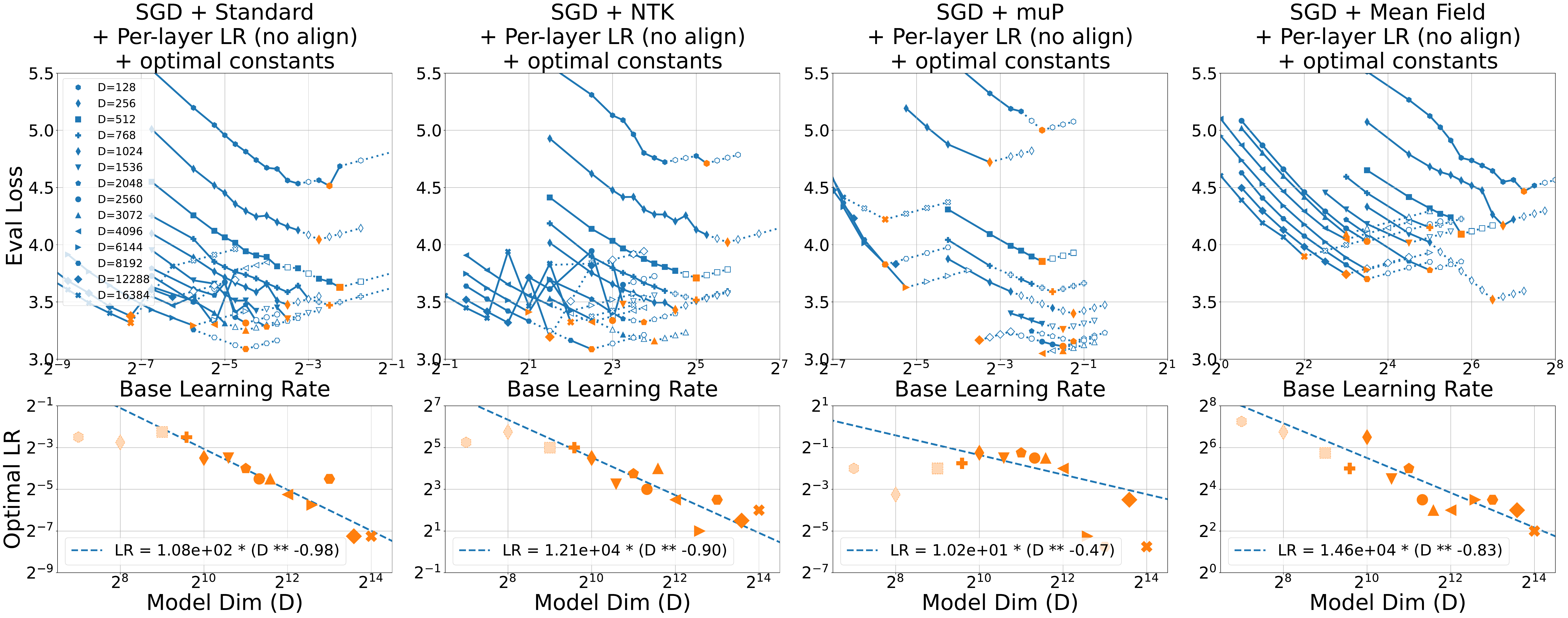}
\caption{Learning rate sweeps and power laws fit to optimal learning rate vs model dim. SGD + per-layer learning rate assuming no alignment + optimal constants. Number of training steps = $50{,}000$.}
\end{SidewaysFigure}
\clearpage

\clearpage
\thispagestyle{plain}
\begin{SidewaysFigure}
\subsection{Learning Rate Sweeps for Adam, all settings}
\label{sec:app_lr_sweeps_adam}
\vspace{12pt}
\includegraphics[width=0.98\linewidth]{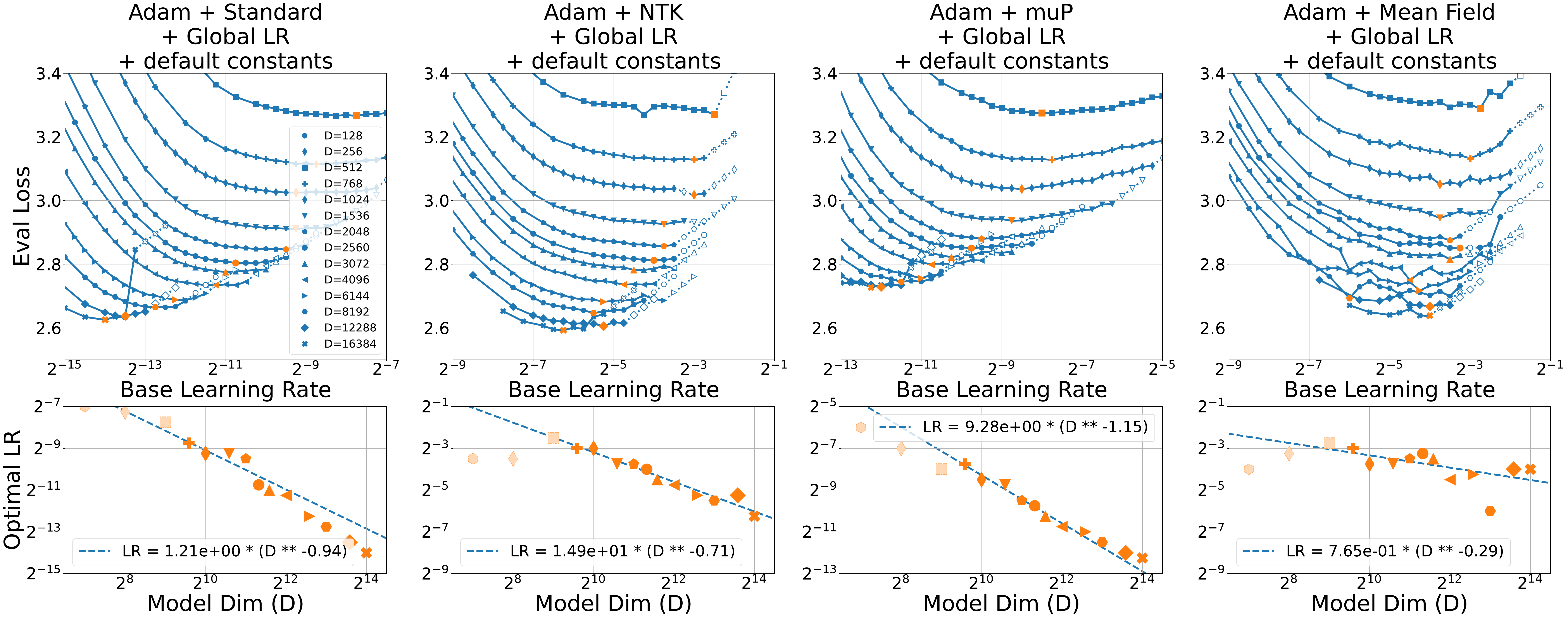}

\figvspace

\includegraphics[width=0.98\linewidth]{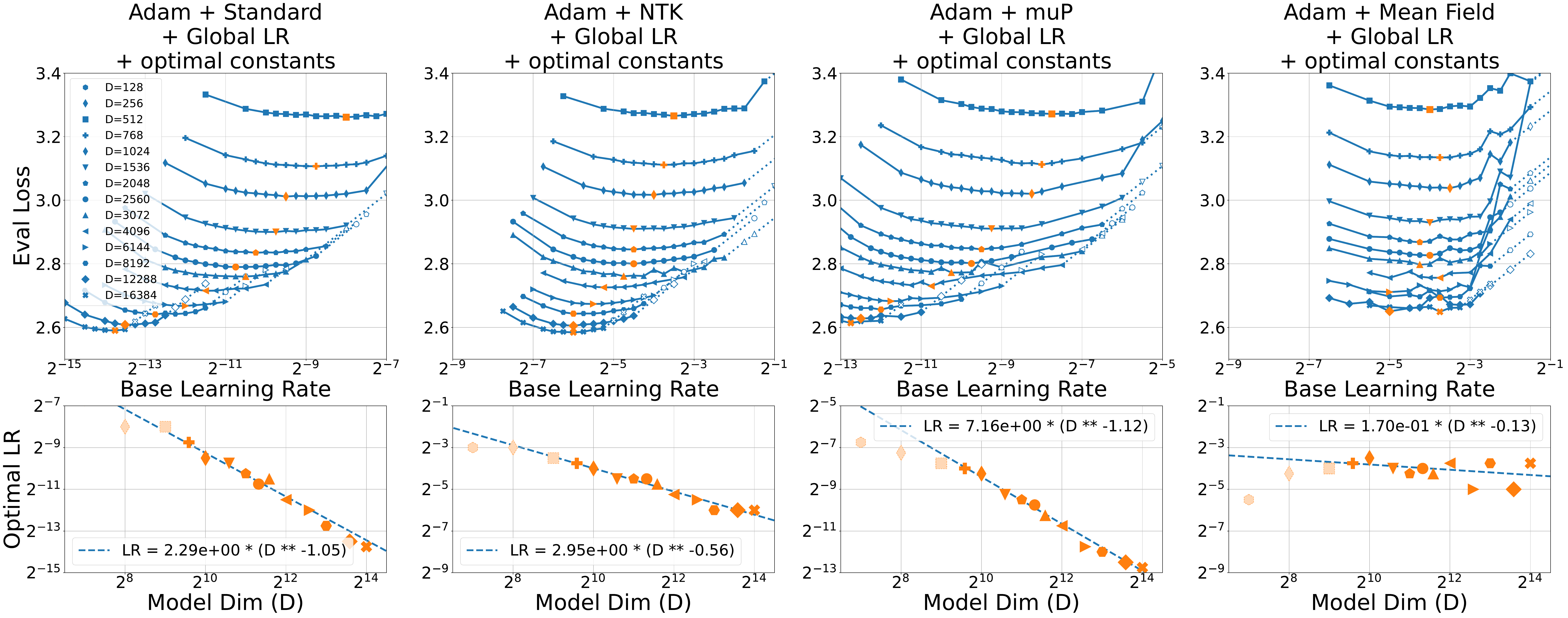}
\caption{Learning rate sweeps and power laws fit to optimal learning rate vs model dim. Top = Adam + global learning rate + default constants. Bottom = Adam + global learning rate + optimal constants. Number of training steps = $50{,}000$.}
\end{SidewaysFigure}
\clearpage

\thispagestyle{plain}
\begin{SidewaysFigure}
\includegraphics[width=\linewidth]{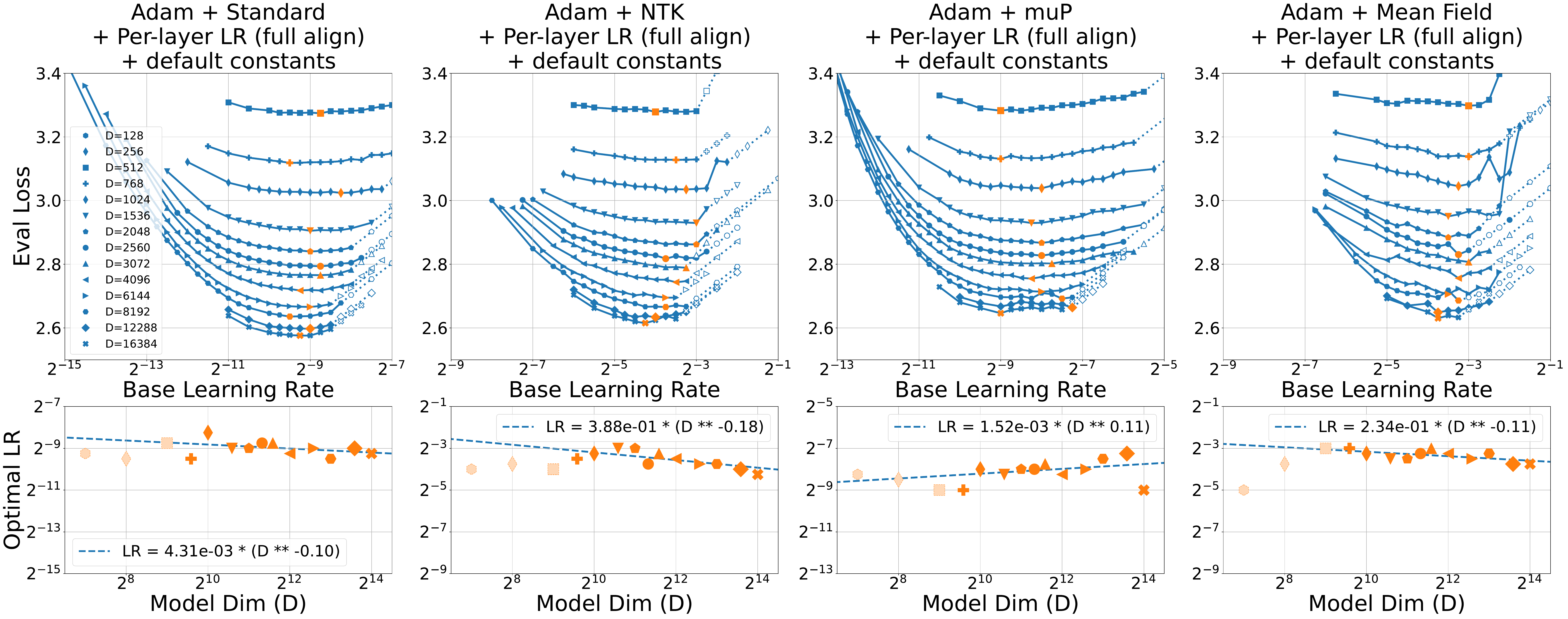}

\figvspace

\includegraphics[width=\linewidth]{icml2024/figures/lr_sweeps/appendix/adam/adam+50k_steps_per_module_lr_optimal_constants.pdf}
\caption{Learning rate sweeps and power laws fit to optimal learning rate vs model dim. Top = Adam + per-layer learning rates assuming full alignment + default constants. Bottom = Adam + per-layer learning rates assuming full alignment + optimal constants. Number of training steps = $50{,}000$.}
\end{SidewaysFigure}
\clearpage

\thispagestyle{plain}
\begin{SidewaysFigure}
\includegraphics[width=\linewidth]{icml2024/figures/lr_sweeps/appendix/adam/adam+50k_steps_per_module_lr_optimal_constants_no_align.pdf}

\figvspace

\includegraphics[width=\linewidth]{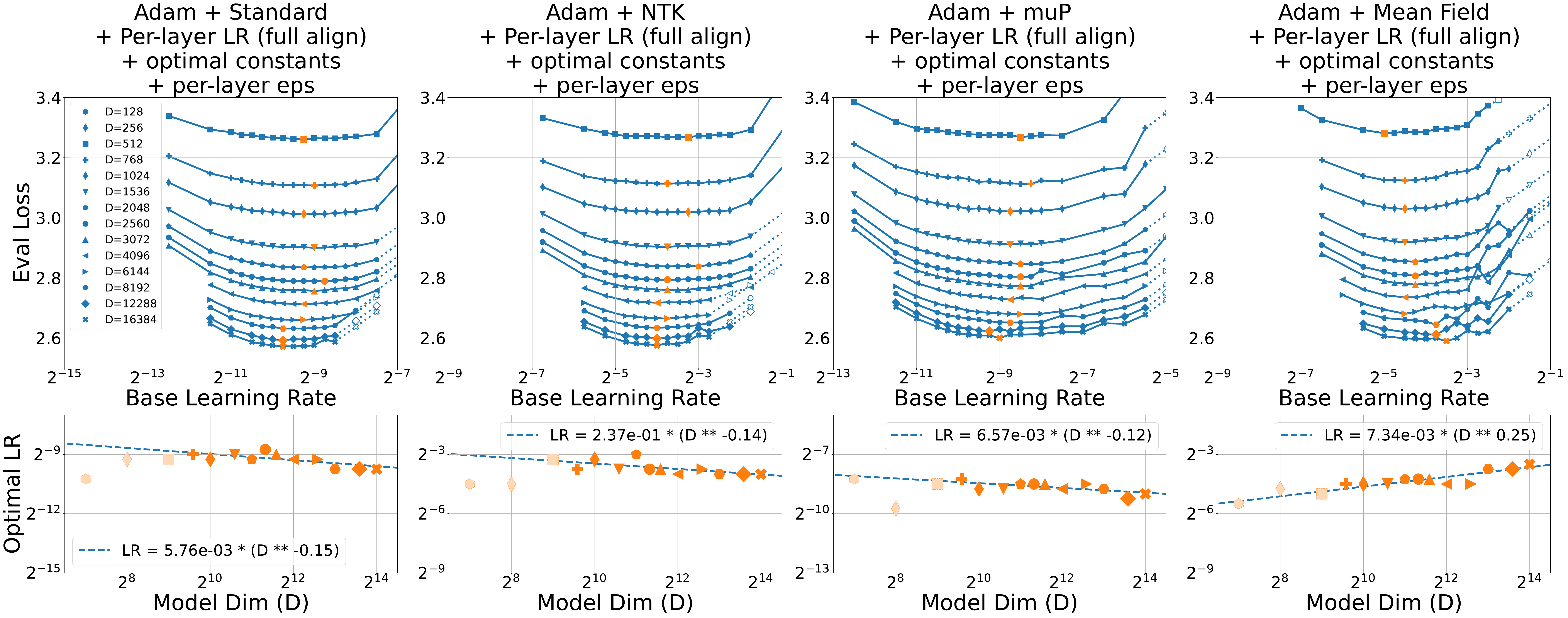}
\caption{Learning rate sweeps and power laws fit to optimal learning rate vs model dim. Top = Adam + per-layer learning rates assuming no alignment + optimal constants. Bottom = Adam + per-layer learning rates assuming full alignment + optimal constants + per-layer epsilon with base epsilon = 1e-12. Number of training steps = $50{,}000$.}
\end{SidewaysFigure}
\clearpage

\thispagestyle{plain}
\begin{SidewaysFigure}
\includegraphics[width=\linewidth]{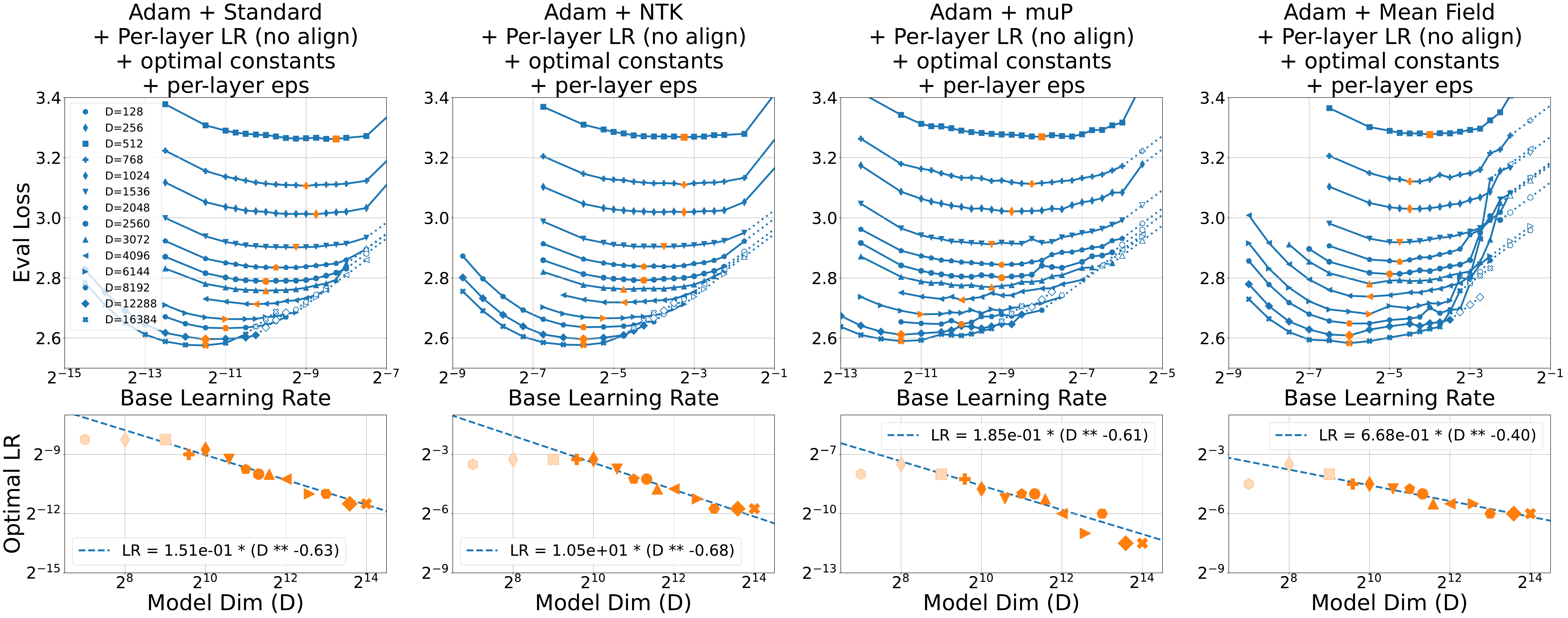}

\figvspace

\includegraphics[width=\linewidth]{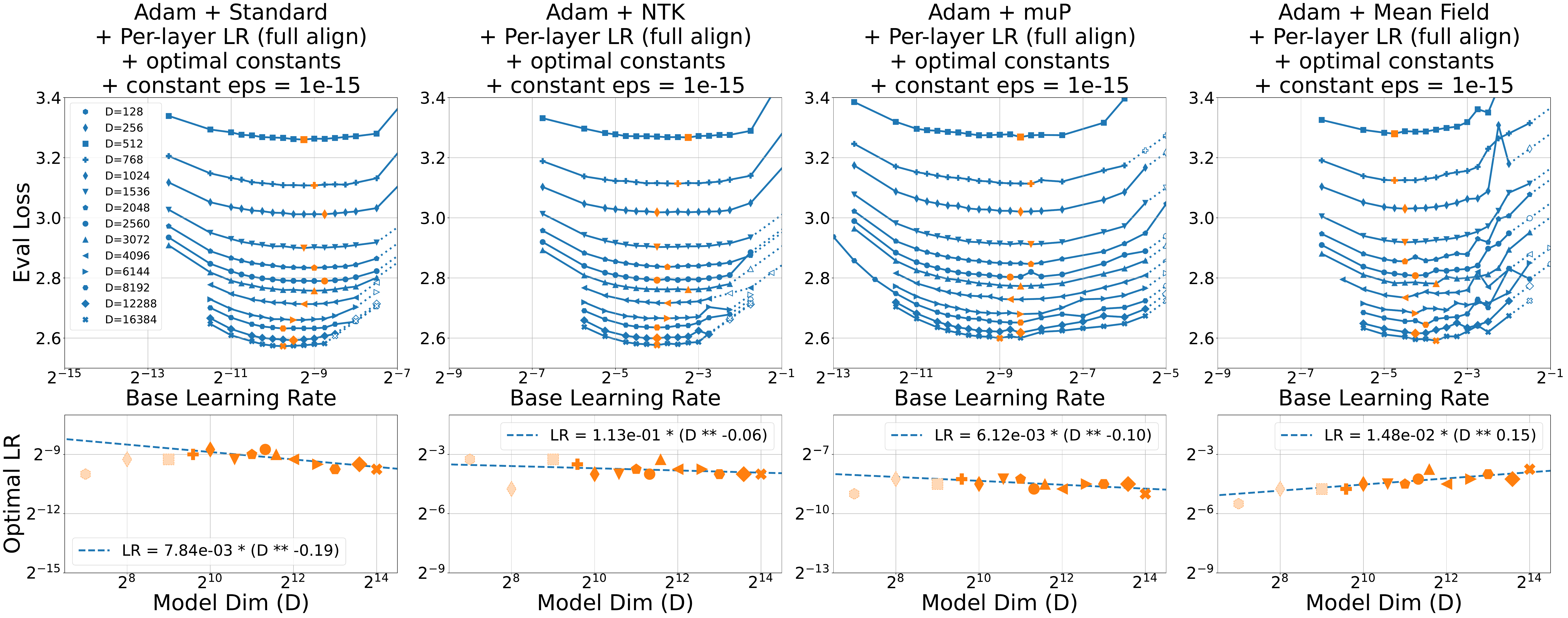}
\caption{Learning rate sweeps and power laws fit to optimal learning rate vs model dim. Top = Adam + per-layer learning rates assuming no alignment + optimal constants + per-layer epsilon with base epsilon = 1e-12. Bottom = Adam + per-layer learning rates assuming full alignment + optimal constants + constant epsilon = 1e-15. Number of training steps = $50{,}000$.}
\end{SidewaysFigure}
\clearpage

\thispagestyle{plain}
\begin{SidewaysFigure}
\includegraphics[width=\linewidth]{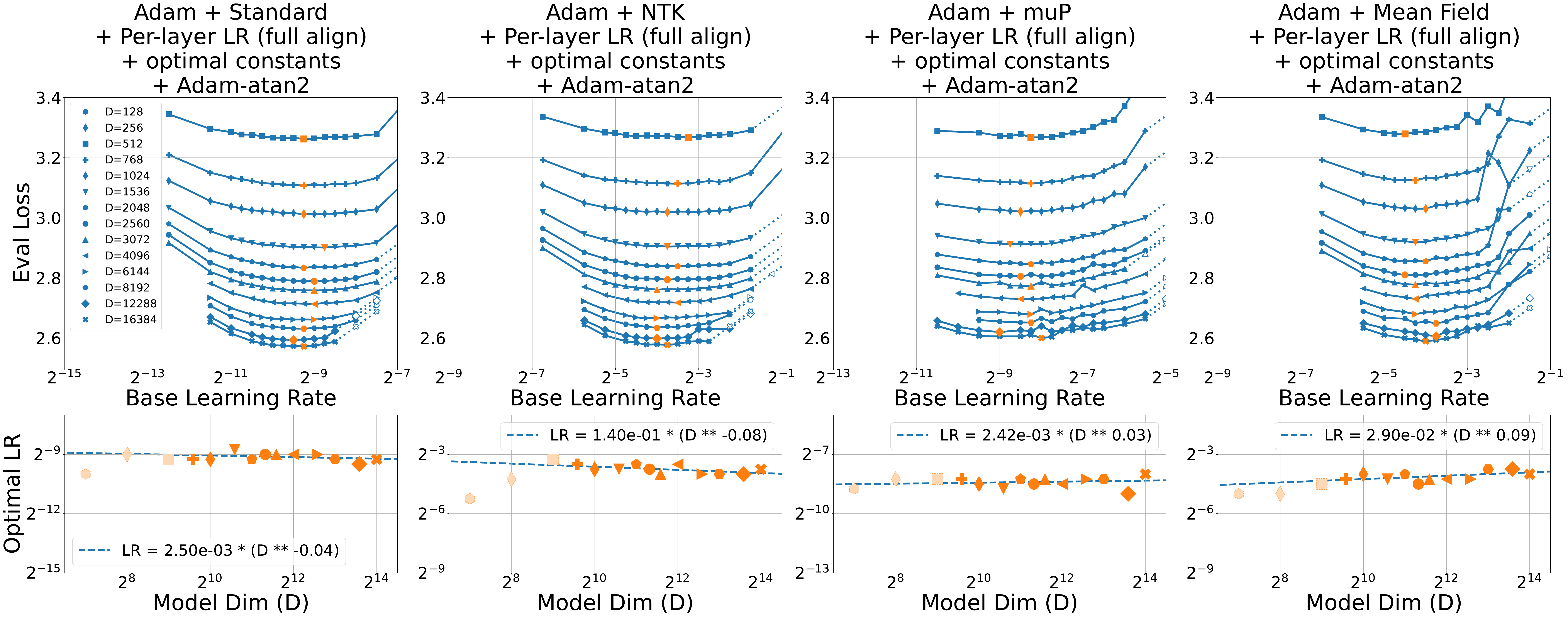}
\caption{Learning rate sweeps and power laws fit to optimal learning rate vs model dim. Adam-atan2 + per-layer learning rates assuming full alignment + optimal constants. Number of training steps = $50{,}000$.}
\end{SidewaysFigure}
\clearpage

\thispagestyle{plain}
\begin{SidewaysFigure}
\subsection{Learning Rate Sweeps for Adam + Parameter Scaling, all settings}
\label{sec:app_lr_sweeps_adam_ps}
\vspace{12pt}
\includegraphics[width=0.98\linewidth]{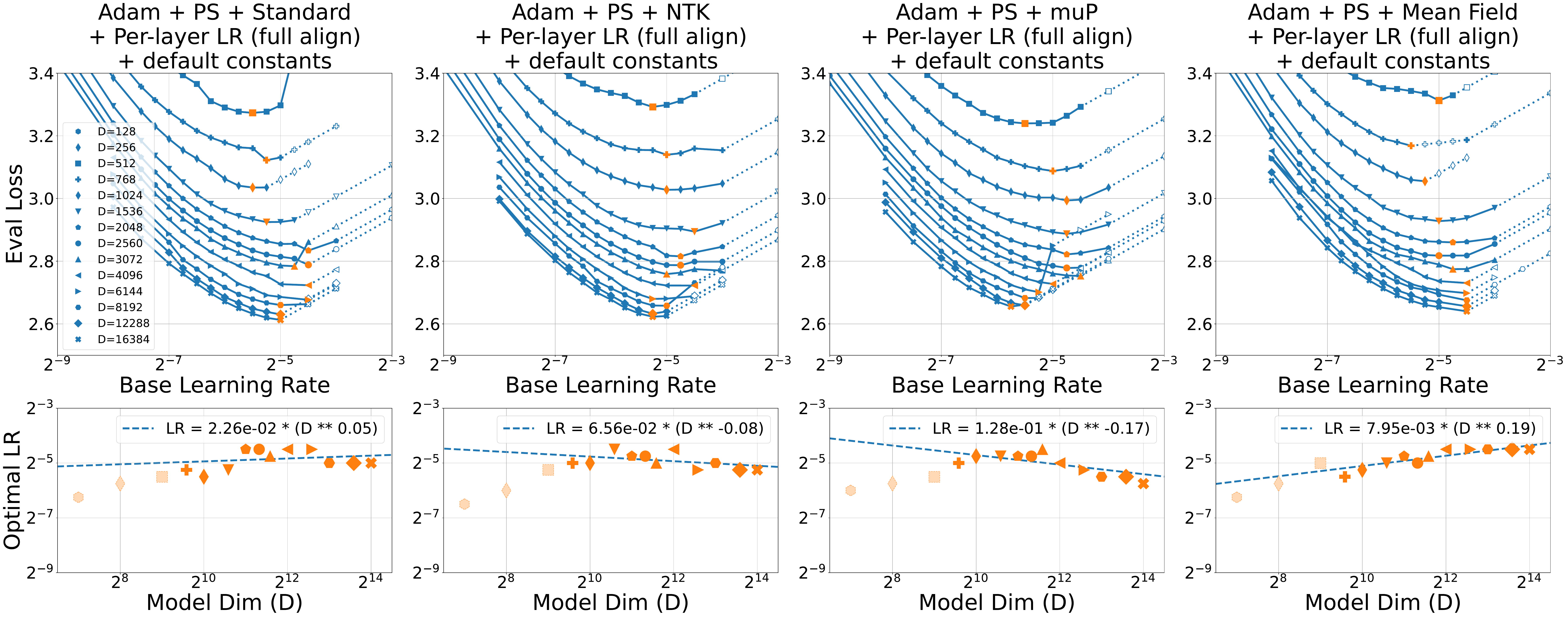}

\figvspace

\includegraphics[width=0.98\linewidth]{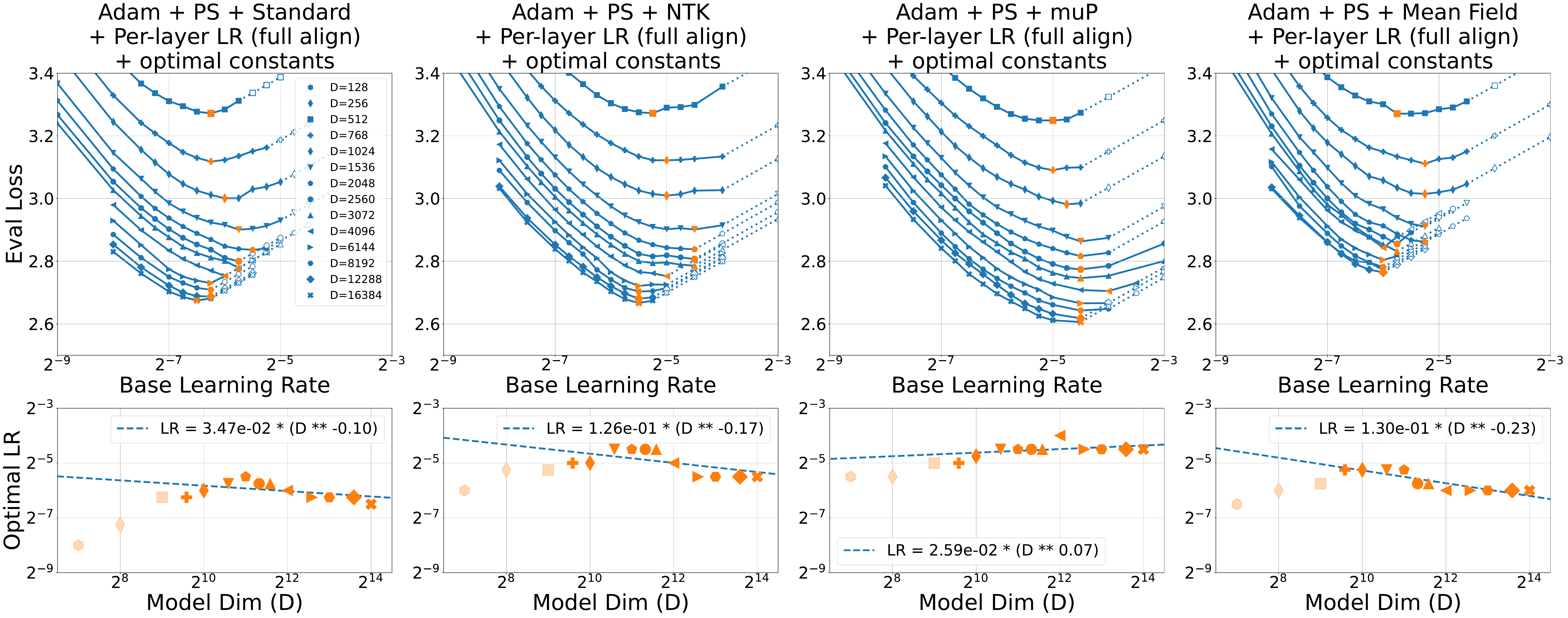}
\caption{Learning rate sweeps and power laws fit to optimal learning rate vs model dim. Top = Adam + parameter scaling + per-layer learning rates assuming full alignment + default constants. Bottom = Adam + parameter scaling + per-layer learning rates assuming full alignment + optimal constants. Number of training steps = $50{,}000$.}
\label{fig:lr_sweep_adam_ps_full_align}
\end{SidewaysFigure}
\clearpage

\thispagestyle{plain}
\begin{SidewaysFigure}
\includegraphics[width=0.98\linewidth]{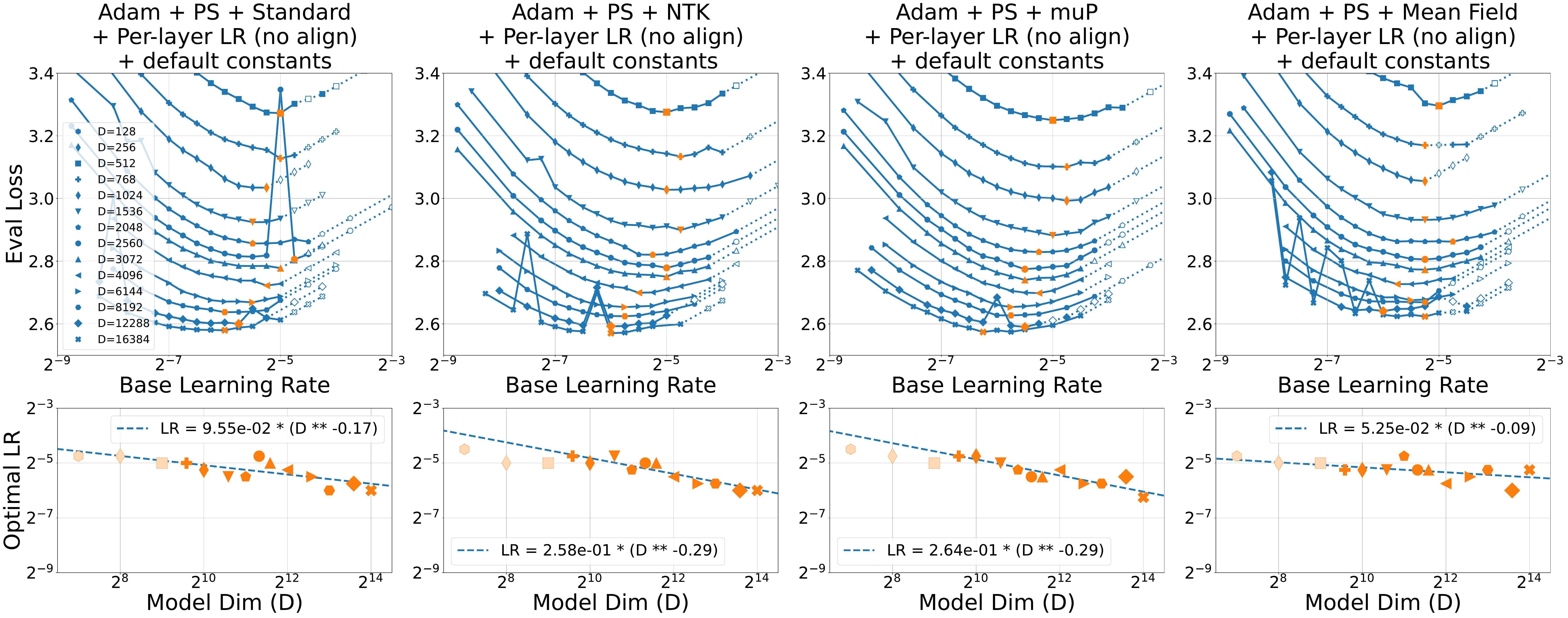}

\figvspace

\includegraphics[width=0.98\linewidth]{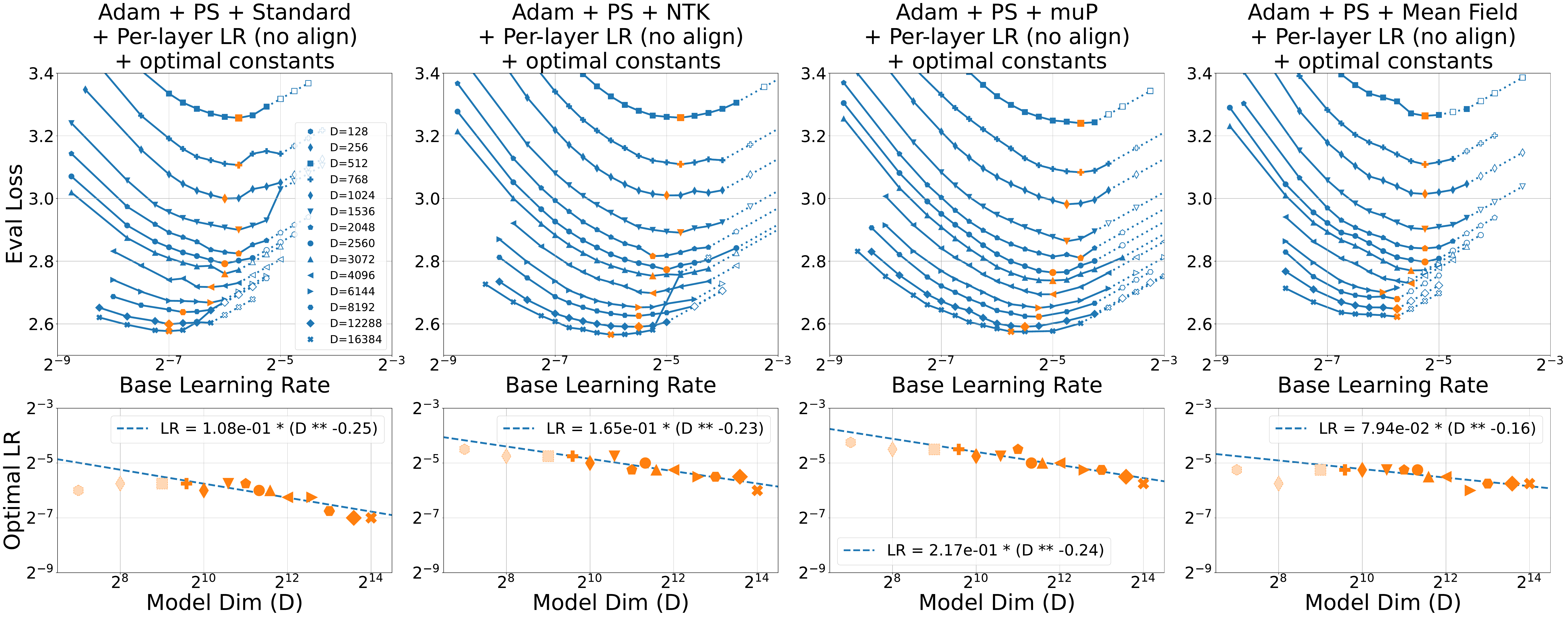}
\caption{Learning rate sweeps and power laws fit to optimal learning rate vs model dim. Top = Adam + parameter scaling + per-layer learning rates assuming no alignment (equivalent to global learning rate) + default constants. Bottom = Adam + parameter scaling + per-layer learning rates assuming no alignment (equivalent to global learning rate) + optimal constants. Number of training steps = $50{,}000$.}
\label{fig:lr_sweep_adam_ps_no_align}
\end{SidewaysFigure}
\clearpage

\thispagestyle{plain}
\begin{SidewaysFigure}
\includegraphics[width=0.98\linewidth]{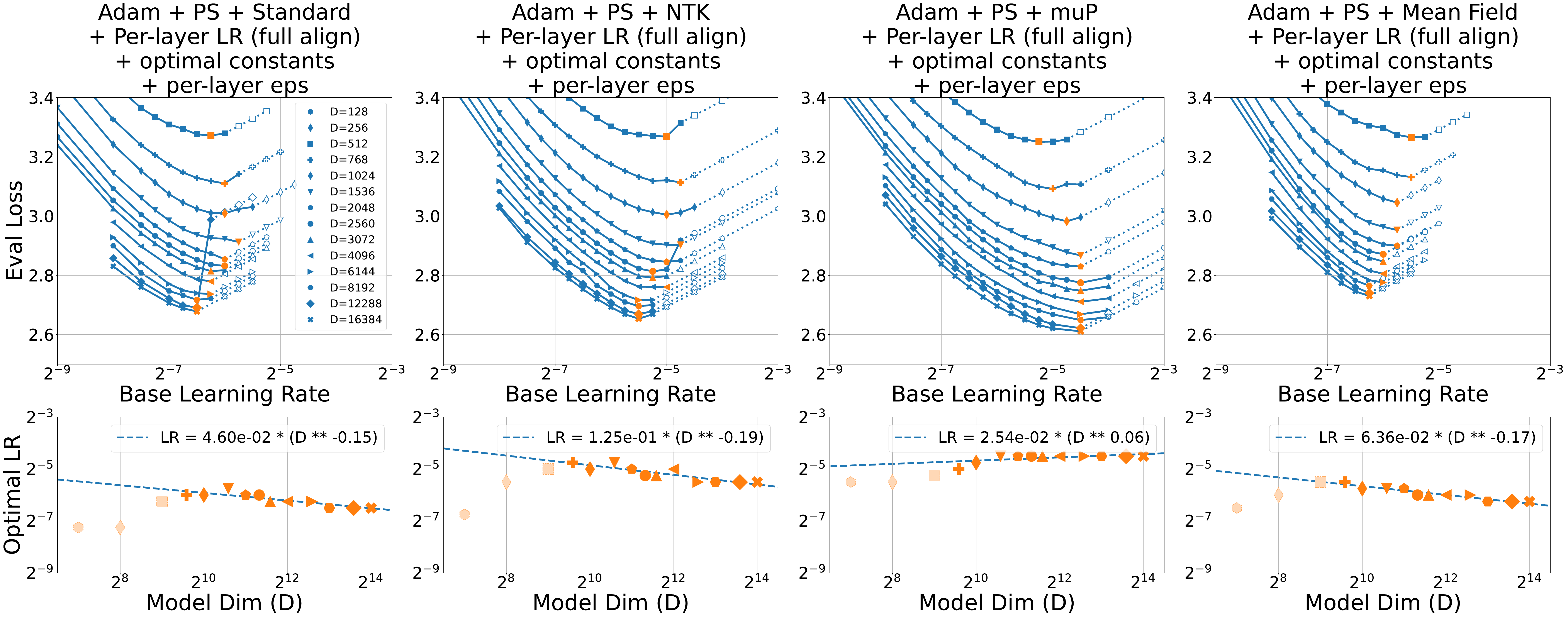}

\figvspace

\includegraphics[width=0.98\linewidth]{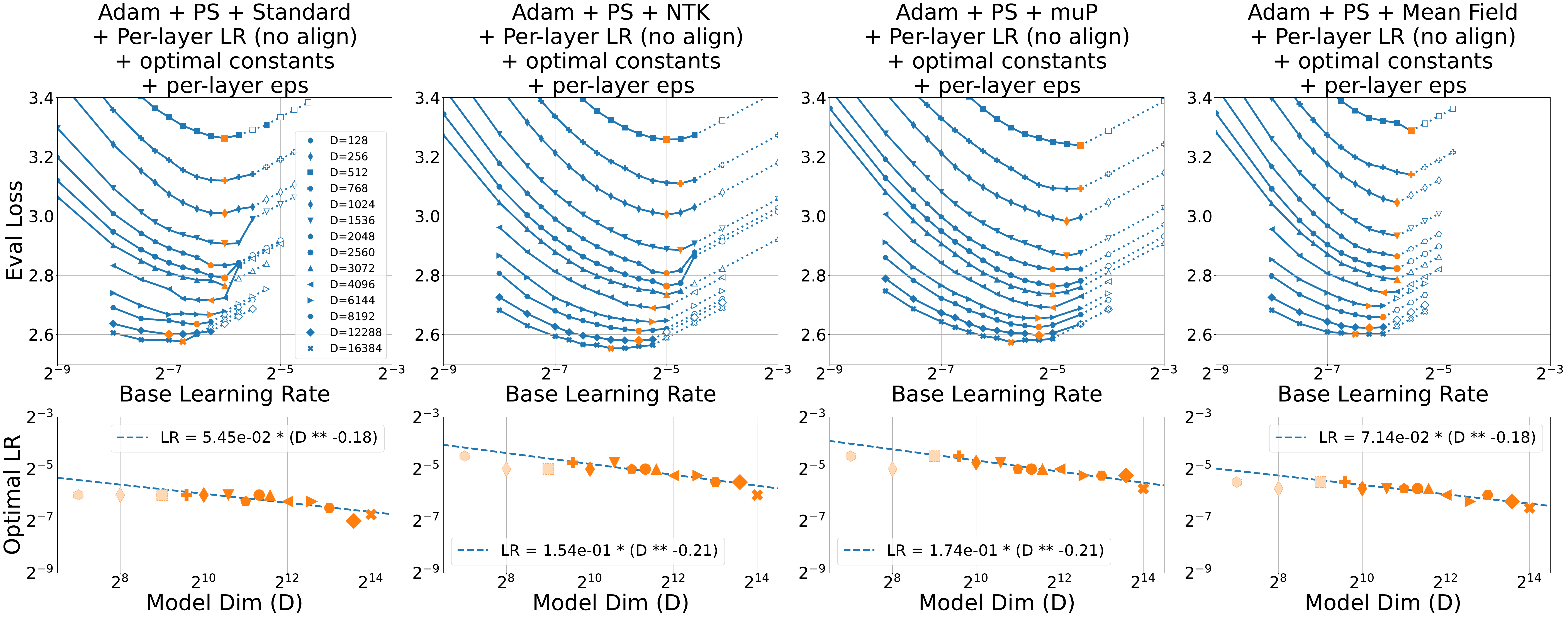}
\caption{Learning rate sweeps and power laws fit to optimal learning rate vs model dim. Top = Adam + parameter scaling + per-layer learning rates assuming full alignment + optimal constants + per-layer epsilon with base epsilon = 1e-12. Bottom = Adam + parameter scaling + per-layer learning rates assuming no alignment (equivalent to global learning rate) + optimal constants + per-layer epsilon with base epsilon = 1e-12. Number of training steps = $50{,}000$.}
\end{SidewaysFigure}
\clearpage

\end{document}